\documentclass{article}

\usepackage{PRIMEarxiv}

\usepackage[utf8]{inputenc} 
\usepackage[T1]{fontenc}    
\usepackage{url}            
\usepackage{booktabs}       
\usepackage{amsmath,amsfonts} 
\usepackage{nicefrac}       
\usepackage{microtype}      
\usepackage{fancyhdr}       
\usepackage{graphicx}       
\usepackage[section]{placeins} 
\usepackage{array}
\usepackage{multirow}
\usepackage{textcomp}
\usepackage{hyperref}       
\graphicspath{{figures/}}   

\hypersetup{
hidelinks,
pdftitle={MorphUNet: Alpha-Controlled Biometric Transport for Diffusion-Based Face Morphing Attacks},
pdfsubject={Face morphing attack generation and biometric vulnerability evaluation},
pdfauthor={Taimoor Rizwan, Sara Atito, Zhenhua Feng, Muhammad Awais, Josef Kittler},
pdfkeywords={face morphing attack, diffusion model, MorphUNet, biometric security, face recognition, morphing attack detection},
}

\newcommand{\morphunet}{MorphUNet}
\newcommand{\map}{MAP}
\newcommand{\mmpmr}{MMPMR}

\newcommand{\etal}{et al.}
\title{MorphUNet: Alpha-Controlled Biometric Transport for Diffusion-Based Face Morphing Attacks
\thanks{Corresponding author: Taimoor Rizwan.}
}

\author{
  Taimoor Rizwan, Sara Atito, Muhammad Awais, Josef Kittler \\
  University of Surrey \\
  Guildford, United Kingdom \\
  \And
  Zhenhua Feng \\
  Jiangnan University \\
  Wuxi, China \\
}

\begin{document}
\maketitle

\begin{abstract}
Face morphing attacks create synthetic facial images that can be verified against multiple identities, posing a direct threat to automated border control and identity verification systems.
This paper introduces \morphunet{}, a diffusion morphing framework that formulates two-parent generation as alpha-controlled biometric transport.
\morphunet{} decomposes each parent into CLIP appearance evidence and ArcFace identity evidence, aligns ArcFace descriptors into a CLIP-compatible token space, and preserves the two contributors as separate identity-aware token banks.
To the best of our knowledge, \morphunet{} is the first diffusion-based face morphing framework to use trainable parent-separated dual cross-attention inside the denoising U-Net.
This Biometric Transport Layer is a trainable U-Net attention block that carries parent-specific identity evidence through denoising by attending to the two parents separately before combining their biometric residuals with the morphing parameter $\alpha$.
DDIM-inverted latent interpolation provides a structurally coherent starting point for denoising, while the proposed weaker-parent-guided candidate selection favours morphs that maximise the lower of the two parent-similarity scores, reducing collapse toward only one contributor.
We evaluate \morphunet{} against three state-of-the-art morphing baselines, StableMorph, MIPGAN-II, and MorDIFF, on FEI and FRLL using six independent face recognition systems, and propose CFD-based unseen-identity stress testing for morphing attacks across gender pairing, ethnicity pairing, demographic shifts, and parent-similarity extremes.
\morphunet{} achieves the best Morphing Attack Potential (\map{}) when at least three of the six systems must be fooled by the same morph, reaching 0.919 on FEI and 0.886 on FRLL, and obtains the best FID on both datasets with 35.19 on FEI and 44.86 on FRLL.
In morphing attack detectability experiments, \morphunet{} also gives the highest APCER at the main 5\% BPCER operating point in the same-dataset detector setting, and remains highly difficult to detect under cross-dataset detector transfer with APCER values of 0.996 on FEI and 0.946 on FRLL.
The full evaluation analyses \map{}, MAD, per-system vulnerability, identity balance, image quality, top- and bottom-similarity stress tests, and CFD unseen-identity robustness.
\end{abstract}


\section{Introduction}
\label{sec:introduction}

Face recognition systems are widely used in identity verification pipelines, but they remain vulnerable to face morphing attacks in which a single submitted image is accepted as more than one person.
This threat is especially relevant in document issuance and border-control settings, where a morphed enrolment image can allow multiple individuals to authenticate against the same credential~\cite{ferrara2014magicpassport,robertson2019facemorphing}.
Unlike generic face synthesis, a morphing attack has two simultaneous objectives.
It must look like a plausible bona fide photograph, and, as a template, it must retain enough biometric evidence from both contributors so that any one of them can be  accepted by face recognition systems.
The second requirement makes morphing fundamentally different from ordinary identity-preserving image generation: the goal is not to reproduce one identity, but to form a controlled trajectory between two identities without allowing either one to disappear.

The development of morphing attacks reflects this tension between realism and biometric balance.
Traditional landmark-based methods directly align, warp, and blend parent images, making the contribution of each parent visible and interpretable, but often leaving local artefacts around facial boundaries and texture transitions~\cite{makrushin2017visapp,scherhag2019survey}.
GAN and latent-space methods improve visual realism by projecting the result onto a learned face manifold, yet latent optimisation can still converge to a realistic face that is biased toward one parent~\cite{zhang2021mipgan,damer2021regenmorph}.
Diffusion models offer a stronger generative prior and a denoising process that can be conditioned over many steps~\cite{ho2020ddpm,rombach2022ldm}.
However, simply interpolating inverted parent latents or averaging identity conditions does not solve the central biometric problem: the two identities must remain separately accessible long enough for the generator to preserve the weaker contributor.

\morphunet{} addresses this problem by treating morph generation as alpha-controlled biometric transport inside a latent diffusion model.
Here, $\alpha \in [0,1]$ is the user-specified mixing coordinate on the morph trajectory, with endpoint values favouring one parent and intermediate values seeking a controlled biometric compromise.
Each parent is decomposed into CLIP appearance evidence and ArcFace identity evidence, and Biometric Token Alignment maps the ArcFace descriptor into a CLIP-compatible token space so that biometric information can be used by the diffusion U-Net.
We use the term Biometric Transport Layer for a trainable U-Net attention block that carries parent-specific identity evidence into the denoising process while keeping the two parents separate until fusion.
To the best of our knowledge, \morphunet{} is the first diffusion-based face morphing framework to introduce trainable parent-separated dual cross-attention inside the denoising U-Net.
This is the central architectural novelty of the method.
Rather than averaging the two identities into a single condition, the Biometric Transport Layer keeps parent-specific token banks separate, attends to each parent independently, and only then mixes the resulting biometric residuals according to $\alpha$.
DDIM-inverted parent-latent interpolation provides a structurally coherent initialisation for denoising, but the primary identity mechanism is this delayed, alpha-controlled transport of parent-specific biometric evidence through the denoising trajectory.

This framing leads naturally to a broader evaluation question.
A visually convincing morph is not necessarily a strong attack, and a morph that fools one recognition model may fail against another.
We therefore evaluate not only visual quality, but also cross-system attack transferability, weaker-parent preservation, morphing attack detectability, and robustness under gender, ethnicity, parent-similarity, and unseen-identity stress conditions.
Together, these requirements motivate a method and evaluation protocol centred on two-parent identity balance rather than visual plausibility alone.

This paper makes the following contributions:
\begin{itemize}
	\item We formulate diffusion face morphing as alpha-controlled biometric transport, where $\alpha$ explicitly parametrises the intended movement from parent $A$ to parent $B$ in both the starting latent and the identity evidence injected during denoising.
	\item To the best of our knowledge, we introduce the first trainable parent-separated dual cross-attention layer for diffusion-based face morphing, preserving the two parents as separate identity-aware token banks and delaying fusion until alpha-controlled residual mixing inside the U-Net.
	\item We introduce Biometric Token Alignment, an ArcFace-to-CLIP token translation module that makes recognition-oriented identity descriptors compatible with the visual token space used by the diffusion backbone.
	\item We introduce a supervised alpha-parametrised morph-trajectory training strategy together with weaker-parent-guided inference-time candidate selection, providing a second mechanism for preserving both contributors by selecting morphs that keep both parent similarities high rather than improving only the easier parent.
	\item We evaluate \morphunet{} against three competing state-of-the-art morphing methods, StableMorph~\cite{kabbani2025stablemorph}, MIPGAN-II~\cite{zhang2021mipgan}, and MorDIFF~\cite{damer2023mordiff}, on FEI and FRLL with six independent face recognition systems using Morphing Attack Potential (\map{}) at $c=1$, $c=3$, and $c=6$, where $c$ is the number of systems that must accept the same morph as both parents; \morphunet{} achieves the highest $c=3$ score on both datasets.
	\item We propose a CFD-based unseen-identity stress-testing protocol for morphing attacks, covering gender pairing, ethnicity pairing, demographic shifts, and parent-similarity extremes outside the primary benchmark identities.
\end{itemize}

\section{Related Work}
\label{sec:related-work}

\subsection{Traditional Face Morphing Attacks}
\label{sec:related-traditional-morphing}

Face morphing attacks aim to create a single facial image that can be verified as more than one contributing identity.
Early morphing pipelines were primarily image-level methods: facial landmarks were detected, parent faces were warped into a shared geometry, and texture or colour information was blended to produce the final image~\cite{ferrara2014magicpassport,makrushin2017visapp}.
These methods are simple, interpretable, and can preserve visible traits from both contributors.
They also established the operational threat model for passport and identity-document enrolment, where one submitted image can create an identity link for multiple people~\cite{ferrara2014magicpassport,robertson2019facemorphing}.
However, landmark-driven blending is sensitive to landmark quality, pose, expression, illumination, hair, face boundary, and background differences.
When parent images are not closely matched, direct warping and texture fusion can produce ghosting, local boundary artefacts, or unnatural transitions around the eyes, mouth, jawline, and hairline~\cite{makrushin2017visapp,scherhag2019survey}.

\subsection{Deep-Learning and Latent-Space Morphing}
\label{sec:related-deep-morphing}

Deep-learning-based morphing methods reduce the dependence on direct pixel blending by creating morphs in learned representation spaces.
GAN-based methods are a major example of this direction because they impose a learned face prior and map the output back onto a realistic facial manifold~\cite{karras2019stylegan,karras2020stylegan2}.
The MIPGAN family introduced MIPGAN-I and the later MIPGAN-II variant, both of which optimise a GAN latent code using identity and perceptual objectives so that the generated face remains close to both parents while preserving visual realism~\cite{zhang2021mipgan}.
In our experiments, the GAN-based baseline is MIPGAN-II, while MIPGAN-I is discussed as its predecessor in the morphing literature.
Other re-generation and StyleGAN-based pipelines similarly exploit learned generative spaces to reduce low-level blending artefacts and produce more realistic attacks~\cite{damer2021regenmorph,venkatesh2020ganmorphs}.

Autoencoder and variational-autoencoder formulations provide another route to representation-level morphing by explicitly encoding an input image into a compact latent representation and decoding it back to the image domain~\cite{kingma2014vae}.
In principle, such representations make interpolation straightforward, since two parent embeddings can be mixed before decoding.
In practice, early autoencoder-style morphing is limited by reconstruction fidelity and by the semantic structure of the learned latent space.
This limitation is one reason why later diffusion-autoencoder methods became attractive: they retain an explicit representation while improving image reconstruction and sample quality~\cite{preechakul2022diffae}.

The main limitation of latent optimisation is that the realism prior does not automatically enforce biometric balance.
A generated face can look plausible while being much closer to one parent than the other.
This is especially problematic for morphing attacks, where the objective is not an average-looking face but a face that retains sufficient evidence from both identities across different recognition systems.
Latent optimisation is also dependent on inversion quality and can become computationally expensive when each pair requires iterative fitting.
\morphunet{} is motivated by this limitation: it does not rely on a single optimised latent code to carry both identities, but preserves the parents as separate conditioning sources during denoising.

\subsection{Diffusion Models for Face Morphing}
\label{sec:related-diffusion}

Diffusion models have become attractive for face morphing because they combine high image fidelity with flexible conditioning mechanisms~\cite{ho2020ddpm,song2020ddim,rombach2022ldm}.
Unlike GANs, diffusion models generate images through an iterative denoising process, allowing structure, appearance, and identity evidence to influence the sample over many steps.
Latent diffusion further reduces computational cost by denoising in an autoencoder latent space while retaining high-quality synthesis and cross-attention conditioning~\cite{rombach2022ldm}.

Existing diffusion morphing methods commonly use one of two strategies.
The first is latent interpolation: parent images are inverted into a diffusion latent space and an intermediate latent is formed by linear or spherical interpolation.
This can preserve coarse pose, illumination, and layout, but it fixes the parent mixture before denoising and does not teach the denoising network how to correct the path toward a biometric morph.
The second strategy uses identity-conditioned synthesis, where recognition embeddings, template embeddings, or visual encoders guide the denoising process.
When the two parents are averaged or fused before they reach the diffusion U-Net, whether in pixel space, latent space, or an embedding condition, the model is no longer forced to keep both identities separately accessible during generation.
MorDIFF demonstrated that diffusion autoencoders can produce strong morphing attacks by exploiting high-fidelity reconstruction and latent-space interpolation~\cite{damer2023mordiff}.
LADIMO later explored face morph generation through biometric template inversion with latent diffusion, showing that recognition embeddings can be inverted and sampled to create morph variants~\cite{grimmer2024ladimo}.
StableMorph uses Stable Diffusion for high-quality face morph generation and emphasises visual realism as a central requirement for practical morphing attacks~\cite{kabbani2025stablemorph}.

These works show the promise of diffusion-based morphing, but also highlight a central challenge: the denoising model must preserve two biometric identities without prematurely averaging them.
\morphunet{} addresses this by treating the interpolated DDIM latent only as a structural initialisation, while identity preservation is learned as feature-level, parent-specific biometric transport inside the denoising U-Net.
In this way, the diffusion model is supervised to follow an alpha-parametrised morph trajectory rather than relying only on a precomputed average of the two parents.

\subsection{Identity-Preserving Diffusion Conditioning}
\label{sec:related-identity-conditioning}

Identity-preserving generation is closely related to morphing because both tasks require a pretrained generator to respect facial identity constraints rather than only produce a plausible face.
Face recognition networks such as FaceNet~\cite{schroff2015facenet}, ArcFace~\cite{deng2019arcface}, MagFace~\cite{meng2021magface}, and AdaFace~\cite{kim2022adaface} learn discriminative embeddings that are strong for verification, but these embeddings are not naturally compatible with the token spaces used by text- or image-conditioned diffusion models.
CLIP features provide rich visual-semantic conditioning for diffusion, but CLIP is not trained specifically as a biometric verifier~\cite{radford2021clip}.
This mismatch motivates methods that translate identity or image evidence into a diffusion-compatible conditioning space.

Personalised and identity-preserving diffusion methods have shown that additional image or identity prompts can be injected into frozen diffusion backbones using lightweight trainable conditioning modules and decoupled cross-attention~\cite{ye2023ipadapter,wang2024instantid}.
These works inspire \morphunet{} at the design level: they show that a strong frozen generator can be steered by adding trainable visual or identity conditioning paths, rather than retraining the whole diffusion model.
However, their usual goal is single-reference identity preservation, not biometric balance between two contributors.
\morphunet{} therefore does not adapt a single-reference adapter directly.
Instead, it introduces two trainable, decoupled cross-attention paths with parent-specific identity-aware token banks.
Each bank combines CLIP appearance with ArcFace identity evidence aligned into the CLIP token space, so the denoising U-Net can attend to parent $A$ and parent $B$ separately.
The Biometric Transport Layers then fuse the two parent residuals only after parent-specific attention, using $\alpha$ to control the mixture under supervised morph-trajectory training.
This delayed alpha-controlled fusion is central to the method: the two identities remain separable long enough for the denoising network to extract parent-specific evidence, and only then are they mixed along the intended morph trajectory.

\subsection{Literature Gap and Method Positioning}
\label{sec:related-positioning}

The related literature leaves a specific gap.
Traditional morphing methods preserve visible parent traits through image-level blending but are vulnerable to local artefacts and do not operate through learned biometric transport.
GAN, autoencoder, and latent-optimisation methods improve realism, but they often represent both parents through a single optimised code or shared latent path, which can bias the output toward one contributor.
Existing diffusion morphing methods improve sample quality and make latent interpolation or identity-conditioned synthesis possible, but they do not explicitly maintain two trainable parent-specific attention paths inside the denoising U-Net.
They therefore tend to rely on a parent mixture that is computed before or outside the denoising dynamics, rather than learning how parent-specific evidence should be transported and corrected across the denoising trajectory.
Another related family is image-specific adaptation, where a diffusion model, text embedding, or low-rank adapter is optimised for a particular subject, image, or image pair~\cite{gal2023textualinversion,ruiz2023dreambooth,hu2021lora,zhang2023diffmorpher}.
Such methods can capture the appearance of specific inputs, but they require per-concept, per-subject, or per-pair optimisation rather than learning a reusable morphing trajectory model.
Single-reference identity-preserving diffusion adapters are also not sufficient for this setting, because a morphing attack requires two identities to remain simultaneously and separately available before they are fused.

\morphunet{} is positioned at this gap.
Its novelty is not simply the use of a diffusion backbone or an identity embedding.
Instead, \morphunet{} proposes a supervised two-parent biometric transport mechanism that learns reusable morphing behaviour without per-image fine-tuning or per-pair adapter weight updates: Biometric Token Alignment translates ArcFace identity evidence into CLIP-compatible parent token banks; the Biometric Transport Layer applies trainable, decoupled cross-attention to the two parents separately; and $\alpha$ controls both the interpolated starting latent and the delayed fusion of parent-specific biometric residuals.
The alpha-parametrised training strategy and weaker-parent-guided candidate selection further make identity balance an explicit part of the generation process, not only a post-hoc evaluation metric.
This positions \morphunet{} as a diffusion morphing framework designed around two-parent identity preservation, cross-system attack transferability, and stress testing under gender, ethnicity, unseen-identity, and parent-similarity shifts.

\section{Architecture and Methodology}
\label{sec:method}

\subsection{Overview}
\label{sec:method-overview}
\begin{figure}[htbp]
	\centering
	\includegraphics[
		width=0.98\textwidth,
		trim={70 60 70 150},
		clip
	]{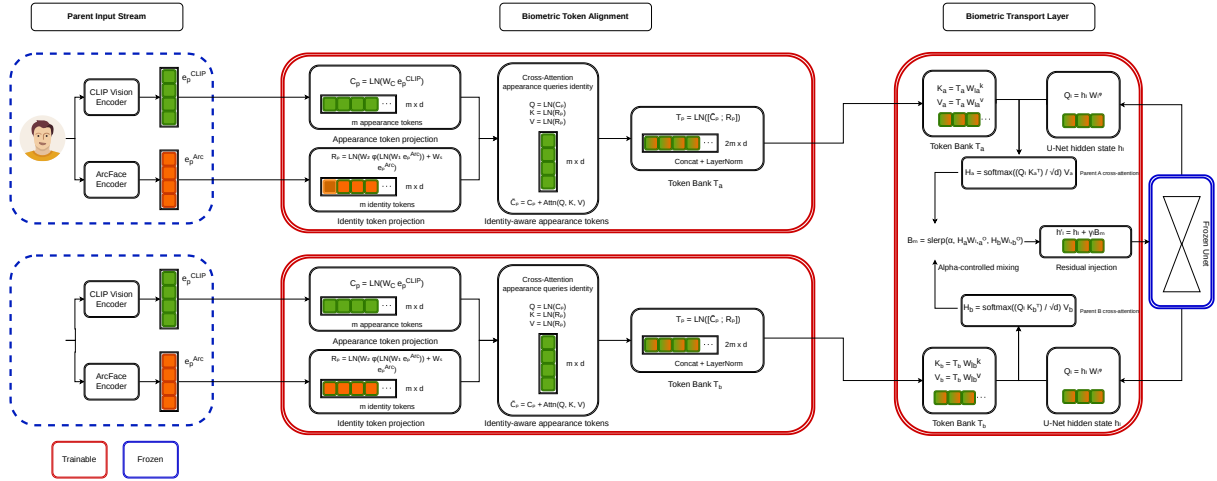}
	\caption{Overview of the \morphunet{} architecture: Two parent images are encoded using frozen CLIP and ArcFace encoders to obtain appearance and biometric identity descriptors. The Biometric Token Alignment module converts ArcFace identity evidence into diffusion-compatible CLIP-space tokens, producing separate parent token banks for parent A and parent B. These token banks are injected into the frozen diffusion U-Net through Biometric Transport Layers, where the U-Net hidden state attends to each parent separately before alpha-controlled residual mixing.}
	\label{fig:morphunet-architecture}
\end{figure}
Given two parent face images $x_A$ and $x_B$, the goal is to generate a morph image $x_M$ that is visually realistic and can be verified as both identities by face recognition systems.
An effective morph must satisfy three requirements simultaneously: it should remain visually plausible, preserve identity evidence from both contributing parents, and transfer across different face recognition systems.
\morphunet{} treats this as a two-parent biometric transport problem inside a latent diffusion model rather than as image blending or unconditional face synthesis.

The method decomposes each parent into two complementary channels.
A CLIP vision encoder~\cite{radford2021clip} provides an appearance descriptor that is well matched to the diffusion model's visual token space, while ArcFace~\cite{deng2019arcface} provides an identity descriptor aligned with face verification.
Both channels are necessary for morph generation.
Appearance descriptors preserve the global visual trajectory between the parents, including pose, texture, illumination, and other image-level cues that cannot be recovered from identity features alone.
Identity descriptors are nevertheless essential because the purpose of a morph is biometric: the generated image must retain enough recognisable identity evidence from both contributors to be accepted by face recognition systems.
\morphunet{} then learns Biometric Token Alignment, an ArcFace-to-CLIP mapping that translates identity evidence into the appearance-token space.
This produces a parent-specific token bank in which appearance tokens have already attended to ArcFace identity evidence.

Generation is controlled by two coupled mechanisms.
First, each parent image is DDIM-inverted into the latent space of the diffusion model, and the parent latents are blended to obtain a structured starting point for denoising.
Second, Biometric Transport Layers inject both parent token banks into the denoising U-Net and mix the resulting biometric residuals with the same interpolation parameter $\alpha$.
Thus, $\alpha$ controls both the starting latent and the identity evidence transported during denoising.

This formulation is designed to generate morphs that are not merely visually intermediate, but identity-balanced under face-recognition features.
In this view, morph generation is a supervised trajectory-learning problem between $x_A$ and $x_B$: the model must learn how appearance and identity change along the path between two parents, and the alpha-parametrised training examples provide intermediate supervision for that path.

\subsection{Preliminaries}
\label{sec:method-preliminaries}

\subsubsection{Latent Diffusion Denoising}
\label{sec:prelim-diffusion}

\morphunet{} follows the standard latent-diffusion formulation used by Stable-Diffusion-style models~\cite{rombach2022ldm}.
Let $x \in \mathbb{R}^{H \times W \times 3}$ denote an image, and let $\mathcal{E}$ and $\mathcal{D}$ denote the frozen VAE encoder and decoder.
The image is mapped to a spatial latent
\begin{equation}
	z_0 = \mathcal{E}(x).
\end{equation}
The decoder maps a clean latent back to the image domain, $\hat{x}=\mathcal{D}(z_0)$.
Diffusion is performed in the latent space rather than directly in pixel space, so the U-Net learns to denoise compact image latents.
Many diffusion papers denote the signal-retention schedule by $\alpha_t$ and its cumulative product by $\bar{\alpha}_t$.
For clarity in this paper, we reserve $\alpha$ for the morph-control parameter introduced in Section~\ref{sec:method-overview}.
Accordingly, the diffusion preliminaries use $\beta_t$ for the variance schedule, $a_t=1-\beta_t$ for the per-step signal-retention coefficient, and $\bar{a}_t=\prod_{s=1}^{t}a_s$ for the cumulative signal-retention coefficient.
Thus, $a_t$ and $\bar{a}_t$ are not additional schedules; they are shorthand quantities derived from $\beta_t$ and are used because the closed-form noising equation is written in terms of retained signal.
For a timestep $t \in \{1,\ldots,T\}$, the forward noising process is
\begin{equation}
	z_t =
	\sqrt{\bar{a}_t} z_0 +
	\sqrt{1-\bar{a}_t}\epsilon,
	\qquad
	\epsilon \sim \mathcal{N}(0,I),
\end{equation}
where $z_t$ is the noisy latent and $\epsilon$ is the Gaussian noise added to $z_0$.
At small $t$, $z_t$ remains close to the clean latent; at large $t$, the latent contains more noise and less visible image structure.
The reverse process is parameterised by a denoising U-Net $\epsilon_{\theta}$, which predicts the injected noise from the noisy latent, the timestep, and a conditioning representation $\mathcal{C}$:
\begin{equation}
	\hat{\epsilon} = \epsilon_{\theta}(z_t,t,\mathcal{C}).
\end{equation}
In text-to-image latent diffusion, $\mathcal{C}$ is commonly a text-token sequence.
In image- or identity-conditioned variants, $\mathcal{C}$ can instead contain visual or identity tokens injected through cross-attention.
Training minimises the standard noise-prediction objective~\cite{ho2020ddpm}:
\begin{equation}
	\mathcal{L}_{\mathrm{simple}} =
	\mathbb{E}_{z_0,t,\epsilon,\mathcal{C}}
	\left[
		\lVert \epsilon-\epsilon_{\theta}(z_t,t,\mathcal{C}) \rVert_2^2
	\right].
\end{equation}
Here, $\mathbb{E}$ denotes an average loss, not the VAE encoder $\mathcal{E}$.
The average is taken over clean training latents $z_0$, Gaussian noise samples $\epsilon$, conditioning inputs $\mathcal{C}$, and discrete timesteps $t$ sampled during training.
Thus, $t$ is not a continuous physical time variable; it is an index into the diffusion noise schedule that tells the U-Net how noisy the current latent is.
In \morphunet{}, $\mathcal{C}$ is not a text prompt.
It is the two-parent conditioning representation produced by Biometric Token Alignment (Section~\ref{sec:biometric-token-alignment}) and injected through the Biometric Transport Layers (Section~\ref{sec:biometric-transport-layer}).
This keeps the image prior fixed while making the trainable path responsible for transporting the two parent identities through denoising.

\subsubsection{DDIM Parent Inversion}
\label{sec:prelim-ddim-inversion}

DDIM sampling~\cite{song2020ddim} gives a deterministic reverse update from $z_t$ to $z_{t-1}$ when its sampler stochasticity parameter is set to zero.
This stochasticity parameter controls whether extra random noise is injected during sampling; it is separate from the diffusion timestep index $t$.
The U-Net prediction gives an estimate of the clean latent at timestep $t$:
\begin{equation}
	\hat{z}_0 =
	\frac{z_t-\sqrt{1-\bar{a}_t}\,
	\epsilon_{\theta}(z_t,t,\mathcal{C})}
	{\sqrt{\bar{a}_t}},
\end{equation}
where $\hat{z}_0$ is the model-implied denoised latent.
The deterministic reverse DDIM step is then
\begin{equation}
	z_{t-1} =
	\sqrt{\bar{a}_{t-1}}\hat{z}_0 +
	\sqrt{1-\bar{a}_{t-1}}\,
		\epsilon_{\theta}(z_t,t,\mathcal{C}).
\end{equation}
Repeating this reverse step from $T$ down to 0 converts a noisy terminal latent into a clean latent that can be decoded into an image.
Using inverted latents as generation anchors is a common practice in deterministic diffusion sampling and latent-diffusion pipelines~\cite{song2020ddim,rombach2022ldm}, and it is also used in diffusion-based face morphing and template-inversion settings to retain parent-specific structure~\cite{preechakul2022diffae,damer2023mordiff,grimmer2024ladimo}.
DDIM inversion applies the corresponding deterministic update in the forward direction.
Starting from a real image latent $z_0$, inversion repeatedly moves from $z_t$ to a noisier $z_{t+1}$ and produces a terminal latent $z_T$ whose reverse DDIM trajectory reconstructs the input under the same model and conditioning:
\begin{equation}
	z_{t+1} =
	\sqrt{\bar{a}_{t+1}}\hat{z}_0 +
	\sqrt{1-\bar{a}_{t+1}}\,
	\epsilon_{\theta}(z_t,t,\mathcal{C}).
\end{equation}

For parent images $x_A$ and $x_B$, \morphunet{} obtains DDIM inversion latents $z_{A,T}$ and $z_{B,T}$.
These parent-derived endpoints are not images themselves; they are noisy terminal latents anchored to the two real parents.
They preserve pose, illumination, and coarse facial structure more faithfully than random sampling.
They are therefore a justified starting point for morph generation, but they are not the main identity mechanism of \morphunet{}.
The model searches between the two real parent trajectories, while Biometric Token Alignment (Section~\ref{sec:biometric-token-alignment}) constructs the parent token banks and the Biometric Transport Layers (Section~\ref{sec:biometric-transport-layer}) control how identity evidence is injected during denoising.

\subsection{Data and Alpha Construction}
\label{sec:data-alpha-construction}

\subsubsection{Training Data Construction}
\label{sec:training-data}

\morphunet{} requires supervision that exposes the model to the path between two identities, not only to the two endpoints.
We therefore propose a triplet-based morph-trajectory training construction.
Each example is represented as $(x_A, x_B, x_{AB})$, where $x_A$ and $x_B$ are the two contributing parent images and $x_{AB}$ is a target morph image used to supervise an intermediate point on the trajectory between them.
Unlike fixed image-level blending, latent interpolation, per-pair latent optimisation, or image-specific diffusion adaptation~\cite{ferrara2014magicpassport,makrushin2017visapp,zhang2021mipgan,damer2023mordiff,grimmer2024ladimo,gal2023textualinversion,ruiz2023dreambooth,hu2021lora,zhang2023diffmorpher}, this construction trains the denoising pathway itself to model how identity evidence should move between parents.
The learned model can then be reused at inference without per-image fine-tuning or per-pair adapter updates.
The main training source is FRLL~\cite{debruine2017frll}, whose passport-style, identity-labelled images provide stable parent pairs for supervised morphing.
FEI~\cite{thomaz2010fei} is used as a complementary controlled dataset in the study and as a separate evaluation domain.
This design is deliberate: by training primarily on FRLL and evaluating on both FRLL and FEI, we can measure not only within-domain synthesis behaviour but also transfer to a distinct controlled face benchmark.
The relatively stable capture conditions help the model focus on the intended transition between two identities rather than on nuisance changes in pose, illumination, or background.

For each triplet, \morphunet{} extracts an appearance descriptor for each parent using the CLIP vision encoder and an identity descriptor for each parent using ArcFace.
The target morph image is used as the supervised denoising target and is also embedded with ArcFace to estimate whether the target is closer to one parent.
All images are resized and centre-cropped to $512 \times 512$ before being encoded by the diffusion VAE.

The resulting corpus is therefore not a generic image-generation dataset: each example contains two parent identities, one supervised morph target, parent appearance descriptors, parent identity descriptors, and an alpha value.
This structure is central to the proposed method because it supplies the information needed to learn an identity-aware trajectory between two parents.
The benchmark sets in Section~\ref{sec:datasets} are reported separately so that evaluation measures attack transferability and identity balance rather than only reconstruction of supervised targets.

The target morph $x_{AB}$ is used only during training to define the denoising target.
At inference time, \morphunet{} does not require a target morph; it only requires the two parent images.
This separation is important because the final evaluated morphs are generated from parent pairs and then selected using parent-wise similarity, rather than copied from the training targets.

\subsubsection{Alpha Supervision and Dynamic Alpha Sampling}
\label{sec:alpha-supervision}

The triplet construction in Section~\ref{sec:training-data} defines what the model is supervised to generate: a morph target explained by two parents.
Alpha supervision defines where that target lies on the parent-to-parent trajectory.
In \morphunet{}, $\alpha$ is part of the supervised denoising condition, so the transport layers learn how parent evidence should change as the morph coordinate changes rather than treating $\alpha$ only as a post-hoc interpolation ratio.

The training data contains nominal alpha labels describing the target morph position between the two parents.
Most balanced morph targets are centred on $\alpha=0.5$, corresponding to the intended midpoint between parent $A$ and parent $B$.
However, a target morph labelled as $0.5$ is not always equally similar to both parents in recognition-feature space.
If all such examples are treated as exactly centred, the model can learn a biased midpoint that reproduces whichever parent is already dominant in the target.

\morphunet{} therefore constructs a dynamic training alpha from the target morph's ArcFace similarities to both parents.
Let $s_A^{AB}$ and $s_B^{AB}$ be the ArcFace similarities between the target morph $x_{AB}$ and parents $x_A$ and $x_B$.
For nominal midpoint examples, the training alpha is assigned to one of two near-midpoint values according to which parent is stronger in the target:
\begin{equation}
	\alpha_{\mathrm{train}} =
	\begin{cases}
		0.4, & s_A^{AB} \geq s_B^{AB}, \\
		0.6, & s_A^{AB} < s_B^{AB}.
	\end{cases}
\end{equation}
This gives the model explicit near-midpoint supervision with parent-dominance information, rather than forcing all balanced morphs into a single rigid midpoint label.

Endpoint examples are also softened during training.
Samples labelled at $\alpha=0$ can be presented as either $0.0$ or $0.1$, and samples labelled at $\alpha=1$ can be presented as either $1.0$ or $0.9$.
This exposes the model to small movements away from exact endpoints and improves continuity in the alpha-controlled generation space.
Together, dynamic midpoint assignment and endpoint softening make $\alpha$ a learned morph-trajectory coordinate rather than a post-hoc blending knob.

\subsection{Architecture}
\label{sec:morphunet-architecture}

\morphunet{} is organised around one architectural principle: the pretrained diffusion model should keep its image prior, while all morph-specific learning should occur in a reusable two-parent conditioning path.
We propose this architecture to make two-parent identity preservation a trainable part of denoising rather than a precomputed blend or per-image adaptation.
The subsections below define the components that implement this principle and distinguish them from single-embedding conditioning, pre-fused parent prompts, and per-image adaptation.

\subsubsection{Backbone and Trainable Components}
\label{sec:backbone-trainable-components}

\morphunet{} is built on a latent diffusion backbone consisting of a VAE encoder/decoder and a U-Net denoiser.
The base VAE and U-Net are kept frozen, while morph-specific learning is restricted to the parent-conditioning path:
\begin{itemize}
	\item {the Biometric Token Alignment module that maps ArcFace identity evidence into CLIP-compatible parent tokens; }
	\item the Biometric Transport Layers inserted into the U-Net attention path.
\end{itemize}

Freezing the backbone is important for two reasons.
First, it preserves the image prior learned by the diffusion model, reducing the risk that the model simply memorises morph artifacts from the training set.
Second, it focuses learning on the harder morphing question: how to represent two parent identities and transport them through an existing denoising trajectory.
The novelty is therefore not full generator fine-tuning, but a reusable morph-specific conditioning path that modifies what the U-Net attends to rather than relearning the image generator or fitting new weights for each parent pair.

\subsubsection{Identity--Appearance Feature Decomposition}
\label{sec:feature-decomposition}

\morphunet{} represents each parent with two complementary descriptors.
For parent $p \in \{A,B\}$, the CLIP vision embedding $e^{\mathrm{CLIP}}_p$ captures image-level appearance, while the ArcFace embedding $e^{\mathrm{Arc}}_p$ captures identity evidence used by face verification systems.
This separation is deliberate.
We use it as the input factorisation for our proposed two-parent transport path, so appearance and biometric evidence can be aligned before being injected into the U-Net.
Appearance features retain the global visual trajectory between parents, including texture, illumination, and coarse facial layout.
Identity features provide the biometric constraint that a valid morph must preserve evidence for both contributors.
Unlike single-vector conditioning or early parent averaging, this decomposition prevents appearance and biometric identity from being collapsed into one ambiguous representation before the denoising model can use them.

A single descriptor is insufficient for this task.
Using only appearance risks producing a visually plausible intermediate face that is weak under face-recognition features.
Using only identity discards image-level cues that make the morph visually coherent.
\morphunet{} therefore keeps the two descriptors separate at extraction time and aligns them only inside a learned token representation.

\subsubsection{Biometric Token Alignment}
\label{sec:biometric-token-alignment}

The parent representation used by the U-Net is produced by our proposed Biometric Token Alignment: an ArcFace-aligned CLIP token bank.
The motivation is twofold.
First, the diffusion backbone is conditioned through a CLIP-compatible visual token space, so identity evidence must be expressed in a form the denoising network can use.
Second, CLIP appearance embeddings alone are not designed to preserve the fine-grained identity evidence required by face verification systems.
ArcFace supplies that missing identity information, while the alignment module translates it into the token space expected by the diffusion model.
This proposed component is novel in the morphing pipeline because it makes biometric recognition evidence usable as diffusion tokens before the two-parent transport step, rather than passing a raw identity vector or a pre-averaged parent embedding to the U-Net.

For parent $p$, the CLIP teacher embedding is projected into $m$ appearance tokens:
\begin{equation}
	C_p = \mathrm{LN}(W_C e^{\mathrm{CLIP}}_p),
\end{equation}
where $C_p \in \mathbb{R}^{m \times d}$.
The ArcFace embedding is projected into $m$ identity tokens using a skip-connected projection:
\begin{equation}
	R_p = \mathrm{LN}\left(
		W_2 \, \phi(\mathrm{LN}(W_1 e^{\mathrm{Arc}}_p))
		+ W_s e^{\mathrm{Arc}}_p
	\right),
\end{equation}
where $R_p \in \mathbb{R}^{m \times d}$, $\phi$ is a GELU activation, and $W_s$ is an identity-preserving skip projection.
The skip path is used because the ArcFace vector is already a compact verification embedding; the projection should adapt it to the token dimension without destroying its identity geometry.
The projector is pretrained before the full morphing model so that these identity tokens already have a diffusion-compatible geometry; the training details are given in Section~\ref{sec:alignment-pretraining}.

During morph generation, the alignment step lets the appearance tokens query the identity tokens:
\begin{equation}
	\widetilde{C}_p =
	C_p +
	\mathrm{Attn}
	\left(
		Q=\mathrm{LN}(C_p),
		K=\mathrm{LN}(R_p),
		V=\mathrm{LN}(R_p)
	\right).
\end{equation}
The final parent token bank is:
\begin{equation}
	T_p = \mathrm{LN}\left([\widetilde{C}_p ; R_p]\right),
\end{equation}
where $T_p \in \mathbb{R}^{2m \times d}$ contains identity-aligned appearance tokens and identity tokens.
This is the point where ArcFace evidence is brought into the CLIP-compatible token space used by the diffusion model.
The result is not a simple concatenation of two embeddings.
It is a parent-specific token bank in which appearance tokens have already attended to identity evidence, allowing the downstream U-Net attention layers to receive identity-aware visual tokens.
The downstream transport layer therefore receives two structured parent token banks, not two unrelated global descriptors.

\subsubsection{Biometric Transport Layer}
\label{sec:biometric-transport-layer}

The diffusion U-Net receives two parent token banks, $T_A$ and $T_B$.
The role of the Biometric Transport Layer is to inject these two parent representations into the denoising process without collapsing them into a single average identity.
We propose this Biometric Transport Layer as the main architectural contribution of \morphunet{}: it turns two-parent conditioning into a trainable, parent-separated attention operation rather than a pre-fused identity prompt.
Its key difference from ordinary cross-attention is that parent $A$ and parent $B$ use separate key/value and output projections before alpha-controlled fusion.
At layer $\ell$, the current U-Net hidden state $h_{\ell}$ forms a query and attends to each parent token bank separately:
\begin{equation}
	\begin{aligned}
	Q_{\ell} &= h_{\ell} W_{\ell}^{Q},\\
	K_p &= T_p W_{\ell,p}^{K},\\
	V_p &= T_p W_{\ell,p}^{V},
	\qquad p \in \{A,B\}.
	\end{aligned}
\end{equation}
The parent-specific cross-attention update is then:
\begin{equation}
	H_p =
	\mathrm{softmax}
	\left(
		\frac{Q_{\ell}K_p^{\top}}{\sqrt{d}}
	\right)V_p,
	\quad p \in \{A,B\}.
\end{equation}
Here, $Q_{\ell}$ is obtained from the U-Net hidden state at layer $\ell$, while $(K_p,V_p)$ are parent-specific key/value projections from token bank $T_p$.
The projections are decoupled across the two parents, so $W_{\ell,A}^{K}, W_{\ell,A}^{V}$ and $W_{\ell,B}^{K}, W_{\ell,B}^{V}$ can learn parent-specific transport paths rather than sharing one averaged conditioning route.
This separate-attention design is important because both identities must remain individually accessible before the morphing ratio is applied.
If the two parent token banks are averaged before attention, the weaker identity can disappear before the denoising network has used it.
By computing $H_A$ and $H_B$ separately, the model first estimates how each parent should influence the current hidden state, and only then mixes those influences.

The two parent updates are projected as biometric residuals:
\begin{equation}
	B_A = H_A W_{\ell,A}^{O},
	\quad
	B_B = H_B W_{\ell,B}^{O},
\end{equation}
and then mixed with the same interpolation value used for latent blending:
\begin{equation}
	B_M = \mathrm{slerp}(\alpha, B_A, B_B).
\end{equation}
The mixed residual is injected back into the U-Net hidden state:
\begin{equation}
	h_{\ell}^{\prime} = h_{\ell} + \gamma_{\ell} B_M,
\end{equation}
where $\gamma_{\ell}$ is the learned or configured transport scale for the layer.
We refer to this complete unit as a Biometric Transport Layer because it transports parent-conditioned identity and appearance evidence into the denoising hidden state.
The transport is residual rather than replacing the U-Net state.
This is a deliberate constraint: the frozen diffusion model remains responsible for natural image formation, while the learned transport path controls how much evidence from each parent is present at each denoising layer.
The layer therefore acts as a controlled biometric steering mechanism rather than a full generator.

The role of this decoupled conditioning design is analysed further in the supplementary material, where the full \morphunet{} transport path is compared against simplified variants such as CLIP-only conditioning, ArcFace-only conditioning, self-attention fusion, ArcFace-query conditioning, and raw interpolated attention injection.
These ablations are not presented as separate competing morphing systems, but as diagnostic variants that test whether attack strength comes from the full two-parent transport mechanism or from a simpler conditioning shortcut.

\subsubsection{Coupled Alpha Control}
\label{sec:coupled-alpha-control}

We propose coupled alpha control, where \morphunet{} uses $\alpha$ as a shared coordinate for both latent structure and transported parent evidence.
First, $\alpha$ controls the interpolation of parent inversion latents, affecting coarse pose, structure, and low-frequency image layout.
Second, the same $\alpha$ controls the mixture of Biometric Transport Layer residuals, affecting how parent evidence is injected during denoising.
This coupling is important because prior interpolation-based morphing can move the starting latent without giving the U-Net a learned mechanism for adjusting biometric evidence during denoising.
For a value $\alpha$, generation is therefore governed by:
\begin{equation}
	\left(
		z_{M,T}(\alpha),
		B_M(\alpha)
	\right),
\end{equation}
where $z_{M,T}(\alpha)$ is the interpolated terminal starting latent and $B_M(\alpha)$ is the alpha-controlled biometric residual.
The coupling avoids a mismatch between where generation starts and which parent evidence is injected.
Latent interpolation alone would move image structure between the two parents but would not guarantee balanced identity evidence during denoising.
Transport-layer interpolation alone would inject parent evidence into a denoising process that may not begin from a parent-consistent latent.
Using the same $\alpha$ for both makes the morphing coordinate consistent across structure and identity, which is necessary for a coherent trajectory between parent $A$ and parent $B$.

\subsection{Training}
\label{sec:morphunet-training}

\subsubsection{Biometric Token Alignment Pretraining}
\label{sec:alignment-pretraining}

The Biometric Token Alignment projector is trained before the full \morphunet{} denoising model.
We train this stage on CelebA~\cite{liu2015celeba}, which contains 202,599 face images from 10,177 identities.
For each training image, the ArcFace embedding is used as the input identity representation, while the CLIP visual encoder and frozen teacher projection produce the target token bank with the same token dimensionality used by the diffusion conditioning pathway.
The projector then learns to predict these CLIP-derived teacher tokens from the ArcFace embedding alone.
This stage is needed because the full morphing model should receive identity tokens that are already compatible with the diffusion token interface, rather than learning that translation from scratch while also learning two-parent morphing.

The pretraining objective combines global token alignment, token-wise alignment, reconstruction, and batch-level contrastive matching:
\begin{equation}
	\mathcal{L}_{\mathrm{align}} =
	\lambda_{\mathrm{cos}}\mathcal{L}_{\mathrm{cos}} +
	\lambda_{\mathrm{tok}}\mathcal{L}_{\mathrm{tokcos}} +
	\lambda_{\mathrm{mse}}\mathcal{L}_{\mathrm{mse}} +
	\lambda_{\mathrm{nce}}\mathcal{L}_{\mathrm{nce}}.
\end{equation}
The global cosine term aligns flattened predicted and teacher token banks, the token-wise cosine term preserves per-token direction, the MSE term stabilises the token scale, and the contrastive term encourages each predicted token bank to match its own teacher representation more strongly than other samples in the batch.
In our implementation, the default weights are $\lambda_{\mathrm{cos}}=1.0$, $\lambda_{\mathrm{tok}}=0.5$, $\lambda_{\mathrm{mse}}=0.25$, and $\lambda_{\mathrm{nce}}=0.1$.

\subsubsection{MorphUNet Training Objective}
\label{sec:training-objective}

After alignment pretraining, \morphunet{} is trained as an alpha-parametrised denoising model for supervised morph targets.
For each triplet $(x_A,x_B,x_{AB})$, the target morph $x_{AB}$ is encoded into a latent $z_{AB}$ using the frozen VAE.
Gaussian noise $\epsilon$ is sampled and added according to the diffusion noise schedule to obtain $z_t$.
The denoising network receives the noisy target latent, timestep, parent token banks, and alpha value, and is trained to predict the injected noise:
\begin{equation}
	\mathcal{L}_{\mathrm{MorphUNet}} =
	\mathbb{E}_{z_{AB}, \epsilon, t}
	\left[
		\left\|
		\epsilon -
		\epsilon_{\theta}(z_t, t, T_A, T_B, \alpha)
		\right\|_2^2
	\right].
\end{equation}

This objective keeps the diffusion target unchanged but changes the conditioning problem: the model must explain the target morph using two parent token banks and an alpha value.
The dynamic alpha construction in Section~\ref{sec:alpha-supervision} provides near-midpoint and endpoint supervision, encouraging a controllable identity trade-off rather than a single fixed midpoint.
Thus, the training objective teaches the transport layers how parent evidence should influence denoising along the morphing trajectory.

\subsubsection{Training and Inference Procedure}
\label{sec:training-procedure}

Training and inference use the same parent-conditioning pathway, but differ in the latent being denoised.
During training, the target morph image is encoded by the frozen VAE, noise is sampled, and a random timestep is selected.
The trainable \morphunet{} path receives the noisy target latent, timestep, parent token banks, and alpha value, then predicts the added noise.
Optimisation updates only the Biometric Token Alignment module and the attention layers inside the Biometric Transport path; the VAE and base U-Net remain frozen.

The trained \morphunet{} model is used without further fine-tuning for FEI and FRLL final morph generation, as well as for the gender/similarity stress-test subsets.
The CFD robustness study is reported separately as an unseen-domain stress test.

At inference, no target morph is available.
Both parent images are preprocessed, represented by CLIP and ArcFace, converted into parent token banks, and DDIM-inverted into the latent space.
For a generation value $\alpha$, the terminal inversion latents are blended using spherical interpolation:
\begin{equation}
	z_{M,T} = \mathrm{slerp}(\alpha, z_{A,T}, z_{B,T}).
\end{equation}
The blended terminal latent $z_{M,T}$ is denoised under the same dual-parent conditioning path used during training.
Thus, training teaches how parent token banks and alpha should explain a supervised morph target, while inference applies the learned transport path to parent-derived latents.
The final benchmark uses 50 inversion and denoising steps.
When multiple generated candidates are available for a parent pair, the final morph is selected using parent-wise verification scores rather than visual quality alone.
The selection favours candidates that improve the weaker parent while avoiding collapse toward the stronger parent, which is consistent with the identity-balance objective used throughout the evaluation.

\section{Evaluation Setup}
\label{sec:experimental-setup}

\subsection{Datasets}
\label{sec:datasets}

FEI~\cite{thomaz2010fei} and FRLL~\cite{debruine2017frll} are the main cross-FRS benchmarks.
They provide complementary controlled face-image domains: FEI contains 200 identities, while FRLL provides passport-style imagery relevant to document and identity-verification settings.
CFD~\cite{ma2015chicago} is reserved for \morphunet{}-only robustness analysis under demographic pairing, domain shift, and parent-pair difficulty.
We use biometric terminology according to the evaluation task.
For face-recognition verification, the genuine parent images are the enrolled reference images or templates, and each generated morph is treated as a probe compared against both parent references.
For morphing attack detection, the same genuine images are the bona fide, or non-attack, class, while generated morphs are attack samples.
The dataset composition is summarised in Table~\ref{tab:dataset-composition}; additional FEI and FRLL stress-test subsets are formed by gender pairing and top/bottom parent similarity buckets.

\begin{table}[htbp]
	\centering
	\scriptsize
	\caption{Dataset composition used in the evaluation. Genuine references are used as parent references in FRS verification and as bona fide samples in MAD; morph attack probes are generated parent-pair images.}
	\label{tab:dataset-composition}
	\resizebox{\linewidth}{!}{%
	\begin{tabular}{lccc}
		\toprule
		Dataset & Genuine references & Morph attack probes & Role \\
		\midrule
		FEI & 200 & 676 per method & Main cross-FRS benchmark \\
		FRLL & 91 & 1096 per method & Main cross-FRS benchmark \\
		CFD & Category-dependent & 28 sets, usually 100 each & Robustness stress test \\
		\bottomrule
	\end{tabular}
	}
\end{table}

\subsection{Attack Evaluation}
\label{sec:attack-evaluation-protocol}

The attack evaluation has two main axes.
Morphing Attack Potential (\map{}) measures face-recognition-system (FRS) attack success by quantifying how many independent face recognition systems are fooled by the same morph probe, while MAD measures whether a detector can identify the generated image as a morphing attack.
Per-FRS vulnerability, demographic and similarity stress tests, and qualitative inspection provide supporting analyses that explain \emph{why} a method succeeds or fails.
The comparative benchmark uses three state-of-the-art baselines spanning the main current morphing families: StableMorph~\cite{kabbani2025stablemorph}, MIPGAN-II from the MIPGAN family~\cite{zhang2021mipgan}, and MorDIFF~\cite{damer2023mordiff}.
We mention MIPGAN-I as the earlier member of the MIPGAN family, but use the name MIPGAN-II throughout the results because MIPGAN-II is the evaluated GAN baseline.

\subsubsection{MAP and Per-FRS Vulnerability}
\label{sec:frs-systems}

The main \map{} evaluation measures whether each generated morph probe is accepted as both contributing parents across multiple independent face recognition systems.
The final six-FRS protocol uses ArcFace~\cite{deng2019arcface}, VGG-Face~\cite{parkhi2015vggface}, FaceNet and FaceNet512~\cite{schroff2015facenet}, MagFace~\cite{meng2021magface}, and AdaFace~\cite{kim2022adaface}.
Using several matchers is important because a morph that succeeds on one embedding model may not transfer to another.
We therefore report \map{} at multiple criteria rather than relying on a single matcher or a single operating point.
Let $N$ be the number of morph probes and $J=6$ be the number of FRS matchers in the protocol.

For morph probe $i$ and matcher $j$, let $a_{ij}^{A}, a_{ij}^{B}\in\{0,1\}$ denote whether the probe is accepted against the reference template of parent $A$ or parent $B$, respectively.
We use the standard indicator notation $\mathbf{1}\{\cdot\}$, which equals 1 when the condition in braces is true and 0 otherwise.
The morph fools matcher $j$ only if both parent comparisons are accepted, i.e. the logical AND of $a_{ij}^{A}$ and $a_{ij}^{B}$ is true:
\begin{equation}
	f_{ij} =
	\mathbf{1}\left\{a_{ij}^{A}=1 \ \mathrm{and}\ a_{ij}^{B}=1\right\}.
\end{equation}
Mated Morph Presentation Match Rate (\mmpmr{}) follows standard morphing-attack evaluation practice~\cite{scherhag2019survey} and reports per-FRS vulnerability, i.e. the fraction of morph probes that fool matcher $j$:
\begin{equation}
	\mmpmr{}_j = \frac{1}{N}\sum_{i=1}^{N} f_{ij}.
\end{equation}
Cross-FRS attack transferability is reported as \map{} parameterised by criterion $c$, where $c$ is the minimum number of FRSs that must be fooled by the same morph probe:
\begin{equation}
	\map{}(c) =
	\frac{1}{N}\sum_{i=1}^{N}
	\mathbf{1}\left\{\sum_{j=1}^{J} f_{ij} \geq c\right\},
	\qquad c \in \{1,\ldots,J\}.
\end{equation}
The main tables report $c=1$, $c=3$, and $c=6$.
Thus, $c=1$ counts morphs accepted by at least one matcher, $c=3$ requires transfer to at least half of the six matchers, and $c=6$ requires all matchers to accept the morph as both parents.
This distinguishes weak attacks that fool one matcher from stronger attacks that remain successful across several or all matchers.
Per-FRS \mmpmr{} is reported to show which recognition systems are most vulnerable.
This analysis is useful because two methods can have similar aggregate \map{} while failing on different matchers.
We visualise the per-FRS matrix as a heatmap and also report the average \mmpmr{} across matchers.

\subsubsection{Stress Tests and Qualitative Inspection}
\label{sec:attack-supporting-analyses}

To complement aggregate averages, we evaluate identity preservation using parent-wise similarities computed in face-recognition feature space.
For a morph with similarity $s_A$ to parent $A$ and $s_B$ to parent $B$, we report:
\begin{equation}
	\begin{aligned}
	s_{\mathrm{mean}} &= \frac{s_A+s_B}{2},\\
	s_{\mathrm{min}} &= \min(s_A,s_B),\\
	\Delta_{\mathrm{id}} &= |s_A-s_B|.
	\end{aligned}
\end{equation}
The minimum similarity measures whether the weaker parent is still represented, while the imbalance term measures collapse toward one parent.
This is important because a morph can achieve high average similarity while still being practically dominated by one identity.

The attack evaluation includes two targeted stress tests.
FEI and FRLL are stratified by gender pairing and parent-pair similarity, producing male--male, female--female, and cross-gender subsets at the top and bottom 1\% of parent similarity.
CFD is used only for \morphunet{} robustness analysis under demographic pairing and domain shift, with categories such as same-ethnicity/same-gender, same-ethnicity/different-gender, same-gender/different-ethnicity, and similarity-extreme parent pairs.
We keep CFD separate from the main baseline comparison because its purpose is to test a different strength of \morphunet{}: generation on previously unseen identities under demographic and similarity stress conditions.
Several competing morphing pipelines require identity-specific optimisation, latent fitting, or dataset-specific preprocessing for the identities being morphed, making a direct CFD comparison less informative without re-running their full identity-specific pipelines.
For \morphunet{}, CFD is therefore used as an unseen-domain stress test after training on FRLL.
This analysis shows how the method behaves on new identities and under controlled category shifts rather than only on the primary FEI/FRLL benchmark pairs.
Qualitative parent--morph grids are used as visual checks that the generated samples remain plausible and that the reported metrics correspond to realistic images.

\subsubsection{Morphing Attack Detectability}
\label{sec:mad-setup}

MAD is evaluated as a detector-side attack analysis rather than as part of the \map{} score.
Genuine reference images are treated as bona fide samples, while generated morphs are treated as attack samples.
The MAD system follows the robust ensemble morph-detection framework of Kashiani \etal{}~\cite{kashiani2022robust}, trained to output an attack probability for each image.
The ensemble contains a ResNet-18 detector~\cite{he2016resnet} and two vision-transformer detectors, ViT-B/16 and ViT-L/32~\cite{dosovitskiy2021vit}.
The final detector score is the soft-vote mean of the three attack probabilities.

We report detector performance using the presentation-attack terminology of ISO/IEC 30107-3~\cite{iso30107}.
BPCER measures the fraction of bona fide images incorrectly rejected as attacks, and APCER measures the fraction of morph attacks incorrectly accepted as bona fide.
For a fixed operating point $\tau$ chosen to achieve a target BPCER, let
$N_{\mathrm{acc}}^{\mathrm{attack}}(\tau)$ be the number of attack scores accepted as bona fide and
$N_{\mathrm{attack}}$ be the number of attacks.
We report:
\begin{equation}
	\begin{aligned}
	\mathrm{APCER}(\tau)
	&=
	\frac{N_{\mathrm{acc}}^{\mathrm{attack}}(\tau)}
	{N_{\mathrm{attack}}},\\
	\mathrm{DetectionRate}(\tau)
	&=
	1-\mathrm{APCER}(\tau).
	\end{aligned}
\end{equation}
Higher APCER means the morphs are harder for the detector to identify.
We report APCER at 1\%, 5\%, 10\%, and 20\% BPCER, with 5\% BPCER used as the main operating point for the MAP--MAD comparison.

\subsection{Quality Evaluation}
\label{sec:image-quality-metrics}

Image quality is assessed using distributional and perceptual metrics.
FID~\cite{heusel2017gans}, KID~\cite{binkowski2018demystifying}, and CMMD~\cite{jayasumana2024rethinking} compare generated morph distributions against bona fide reference images; lower values indicate that the generated images are closer to the reference distribution.
FID and KID use Inception features, while CMMD uses CLIP embeddings with a maximum-mean-discrepancy distance.
LPIPS~\cite{zhang2018lpips} is used to measure perceptual distance between each morph and its two parents; we report LPIPS to parent $A$, LPIPS to parent $B$, their mean, and their absolute imbalance.

\section{Results}
\label{sec:results}

The results are organised around the two main attack axes defined in Section~\ref{sec:attack-evaluation-protocol}: \map{} for FRS attack success and MAD for detector-side attack detectability.
We begin with qualitative FEI/FRLL examples to make the morphing task visually concrete, then report the FEI/FRLL \map{} benchmark, including matcher-specific vulnerability and identity balance.
We then analyse MAD, image quality, controlled stress tests, and unseen-identity CFD robustness.
The supplementary material provides the extended robustness analysis supporting the main FEI/FRLL and CFD results, including stricter cross-FRS degradation, identity-imbalance distributions, full gender/similarity stress-test behaviour, and qualitative grids.
It also reports the conditioning ablations used to analyse which parts of the \morphunet{} identity-conditioning path are most responsible for the observed attack-strength and quality trade-off.
This structure separates three questions: whether the morph is accepted by recognition systems, whether both parents remain represented, and whether the image remains visually plausible and difficult to detect.

\subsection{Qualitative FEI and FRLL Examples}
\label{sec:qualitative-results}

\begin{figure}[htbp]
	\centering
	\includegraphics[width=0.49\textwidth]{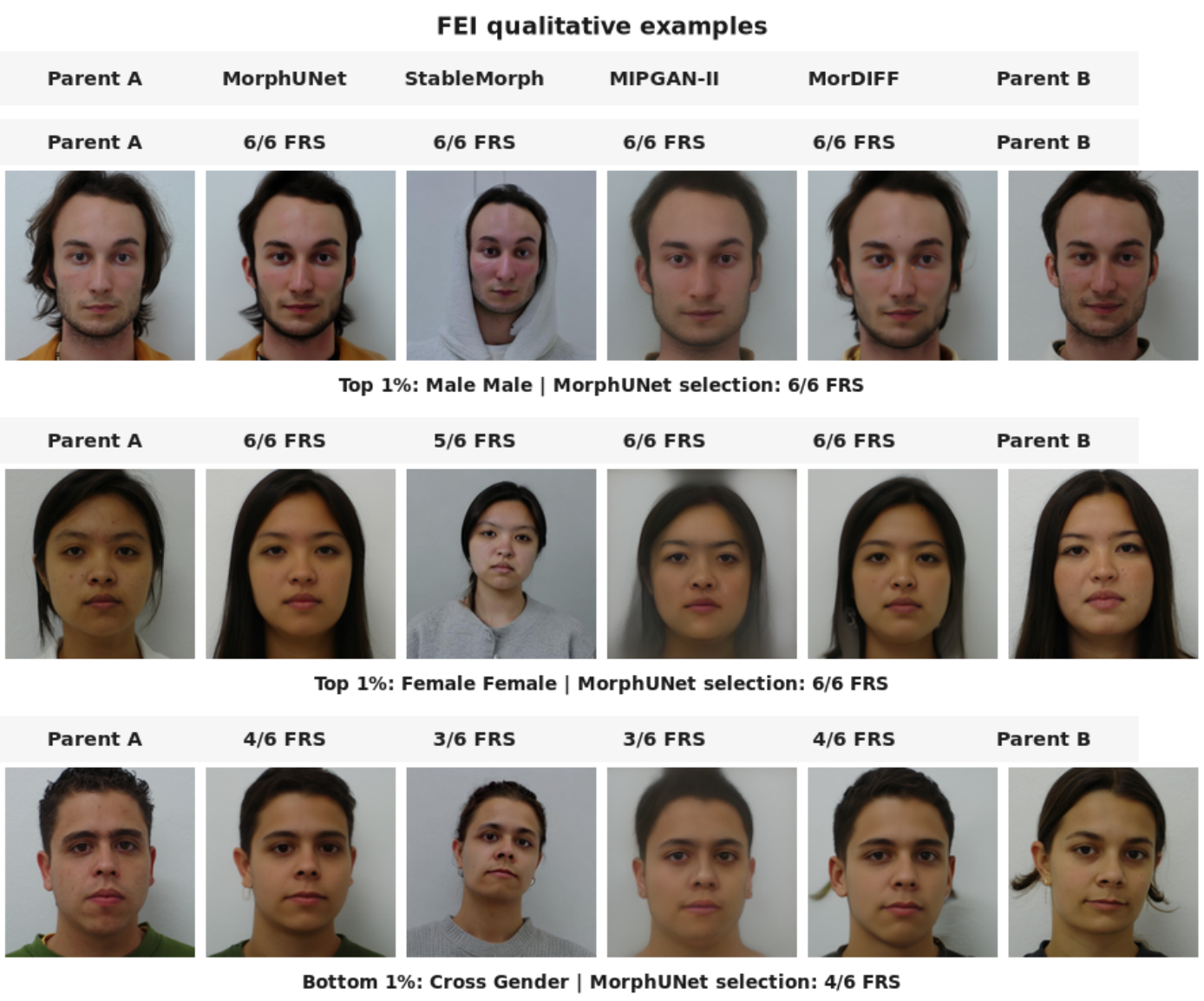}
	\hfill
	\includegraphics[width=0.49\textwidth]{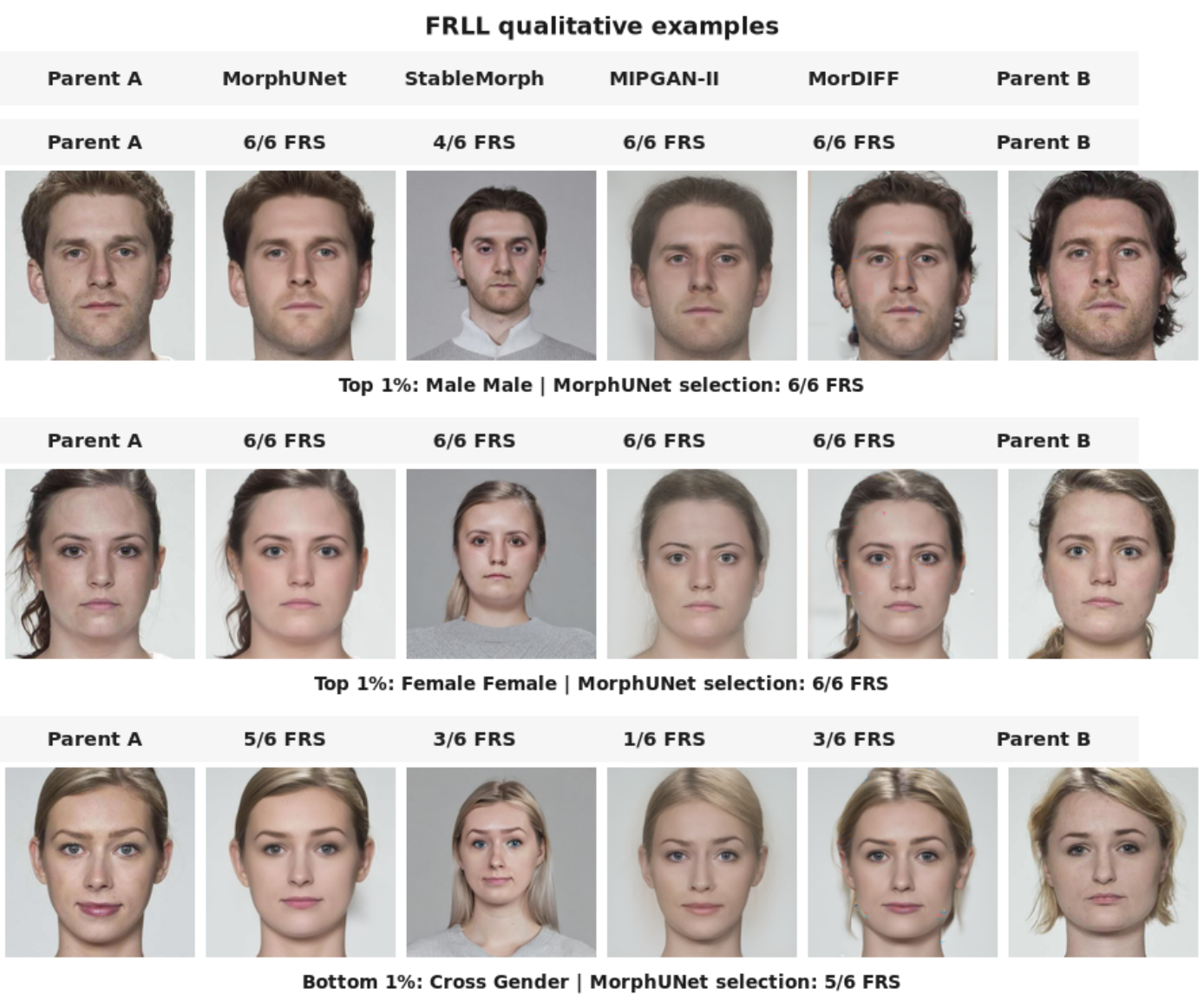}
	\caption{Qualitative FEI and FRLL examples. Each grid shows the two parents and the generated morphs for representative gender/similarity stress-test pairs.}
	\label{fig:qualitative-main}
\end{figure}

The qualitative grids in Figure~\ref{fig:qualitative-main} show parent images and generated morphs under the same pair selections used for the gender/similarity stress tests.
They are not used as evidence in place of the quantitative results, but they provide an important check on whether high \map{} is achieved through visually plausible images.
The \morphunet{} samples preserve central facial structure while avoiding the strong blending artifacts typical of purely image-space morphing.
The examples also show why identity balance is a difficult objective: even when the image is realistic, small shifts in eye shape, mouth region, or face outline can make one parent dominate a face-recognition embedding.
Full FEI and FRLL stress-test qualitative grids are included in the supplementary material.

\subsection{MAP Benchmarking}
\label{sec:main-map-results}

Table~\ref{tab:main-map-benchmark} reports the main FEI and FRLL six-FRS benchmark from the final MAP evaluation table.
At this setting, a morph is counted as successful only when at least three of the six independent face recognition systems accept it as both parents.
\morphunet{} obtains the strongest FRLL \map{} at $c=3$ with 0.886, ahead of MorDIFF at 0.818, MIPGAN-II at 0.501, and StableMorph at 0.441.
On FEI, \morphunet{} reaches 0.919 at $c=3$, narrowly exceeding MorDIFF at 0.913 and substantially exceeding MIPGAN-II and StableMorph.
The stricter $c=6$ setting further shows that FEI morphs from \morphunet{} and MorDIFF retain non-trivial all-matcher transferability, whereas FRLL is harder for all methods at this criterion.

\begin{figure}[htbp]
	\centering
	\includegraphics[width=0.78\linewidth]{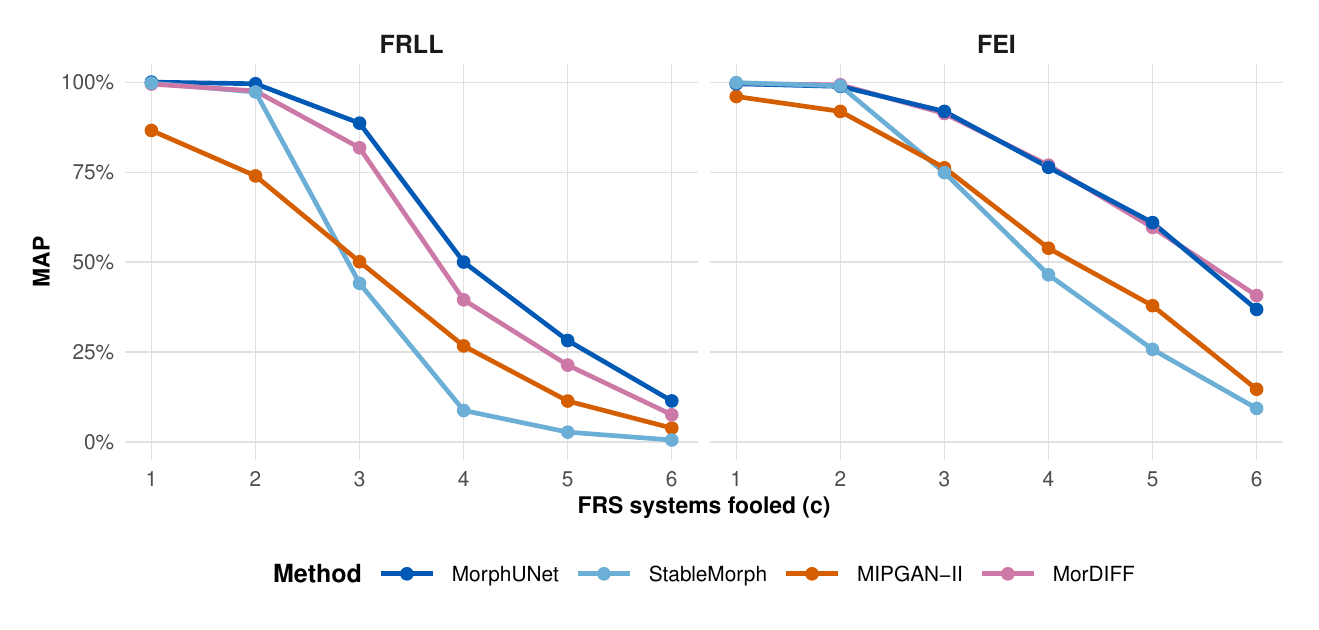}

	\caption{\map{} comparison: attack success as the cross-FRS criterion increases from $c=1$ to $c=6$.}
	\label{fig:map-c-curve}
\end{figure}

\begin{table}[htbp]
	\centering
	\scriptsize
	\caption{Six-FRS MAP benchmark. Higher \map{} and mean minimum similarity are better; lower identity imbalance is better. Bold indicates the best value per dataset and metric, with ties included.}
	\label{tab:main-map-benchmark}
	\resizebox{\linewidth}{!}{%
	\begin{tabular}{llrrrrrr}
		\toprule
		Dataset & Method & \#Morphs & \map{} $c=1$ ($\uparrow$) & \map{} $c=3$ ($\uparrow$) & \map{} $c=6$ ($\uparrow$) & Min Sim. ($\uparrow$) & Imbal. ($\downarrow$) \\
		\midrule
		FRLL & \morphunet{} & 1096 & \textbf{1.00} & \textbf{0.89} & \textbf{0.11} & \textbf{0.55} & \textbf{0.09} \\
		FRLL & StableMorph & 1096 & \textbf{1.00} & 0.44 & 0.01 & 0.45 & 0.12 \\
		FRLL & MIPGAN-II & 1266 & 0.87 & 0.50 & 0.04 & 0.46 & 0.11 \\
		FRLL & MorDIFF & 1096 & 0.99 & 0.82 & 0.08 & 0.52 & 0.10 \\
		\midrule
		FEI & \morphunet{} & 676 & \textbf{1.00} & \textbf{0.92} & 0.37 & \textbf{0.59} & \textbf{0.09} \\
		FEI & StableMorph & 676 & \textbf{1.00} & 0.75 & 0.09 & 0.53 & 0.11 \\
		FEI & MIPGAN-II & 676 & 0.96 & 0.76 & 0.15 & 0.50 & 0.10 \\
		FEI & MorDIFF & 676 & \textbf{1.00} & 0.91 & \textbf{0.41} & \textbf{0.59} & \textbf{0.09} \\
		\bottomrule
	\end{tabular}%
	}
\end{table}
Figure~\ref{fig:map-c-curve} visualises the same result across $c=1$, $c=3$, and $c=6$.
The important pattern is the slope of degradation as the acceptance criterion becomes stricter.
Many methods can fool at least one matcher, but a stronger morphing attack should remain effective when several recognition systems must be fooled simultaneously.
\begin{figure}[htbp]
	\centering
	\includegraphics[width=0.90\linewidth]{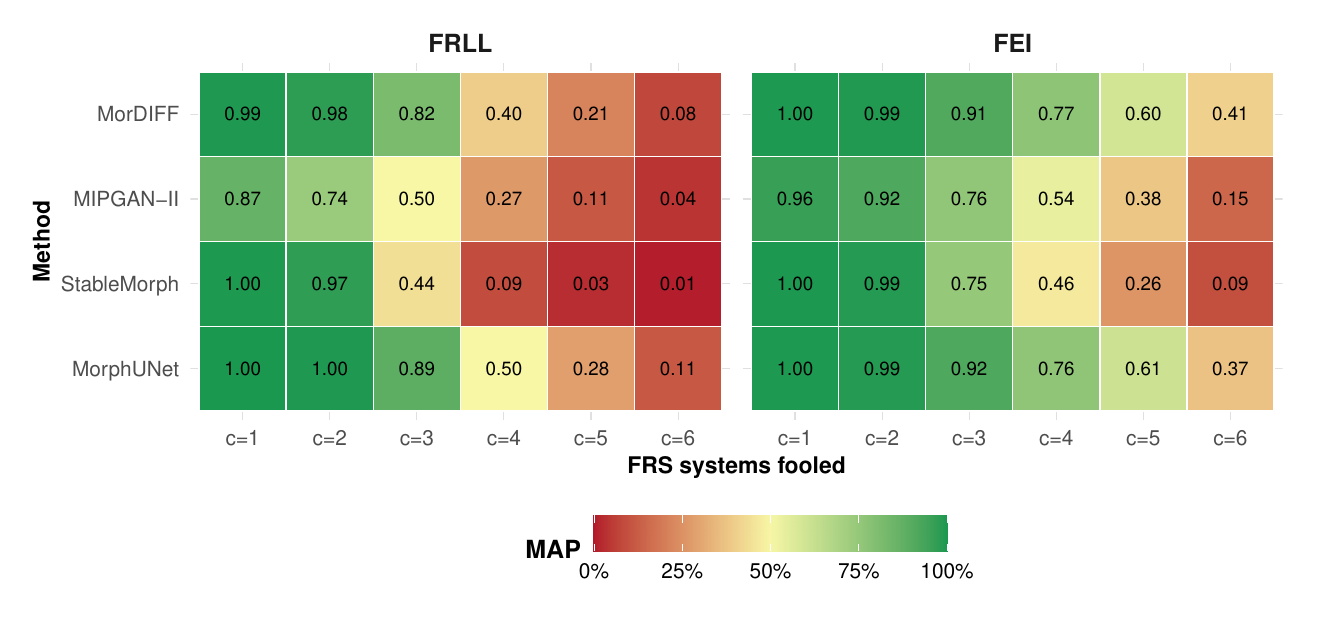}

	\caption{Method-by-dataset heatmap for the principal \map{} criterion.}
	\label{fig:map-heatmap}
\end{figure}
\morphunet{} has the best $c=3$ result on both datasets and the best FRLL result at all three criteria.
On FEI, MorDIFF has the highest $c=6$ score, but \morphunet{} achieves the strongest $c=3$ score while also providing the best image-distribution quality in Table~\ref{tab:image-quality-results}.
The FEI result is therefore best interpreted as a close high-strength regime in which \morphunet{} and MorDIFF are both strong attacks, while the FRLL result shows a clearer separation in favour of \morphunet{}.
Importantly, the strict $c=6$ criterion measures only all-matcher biometric acceptance.
It does not by itself determine the best morphing method, because a deployable morph must also remain visually plausible, avoid local synthesis artefacts, preserve both parents, and withstand detector-side analysis, as evaluated through qualitative inspection, identity balance, MAD, and image-quality metrics in Sections~\ref{sec:qualitative-results}, \ref{sec:identity-results}, \ref{sec:mad-results}, and~\ref{sec:quality-results}.

Across both datasets, \morphunet{} is the only method that combines top $c=3$ \map{}, low identity imbalance, and the best FID.

\subsubsection{Per-FRS Vulnerability}
\label{sec:per-frs-results}

\begin{figure}[htbp]
	\centering
	\includegraphics[width=\linewidth]{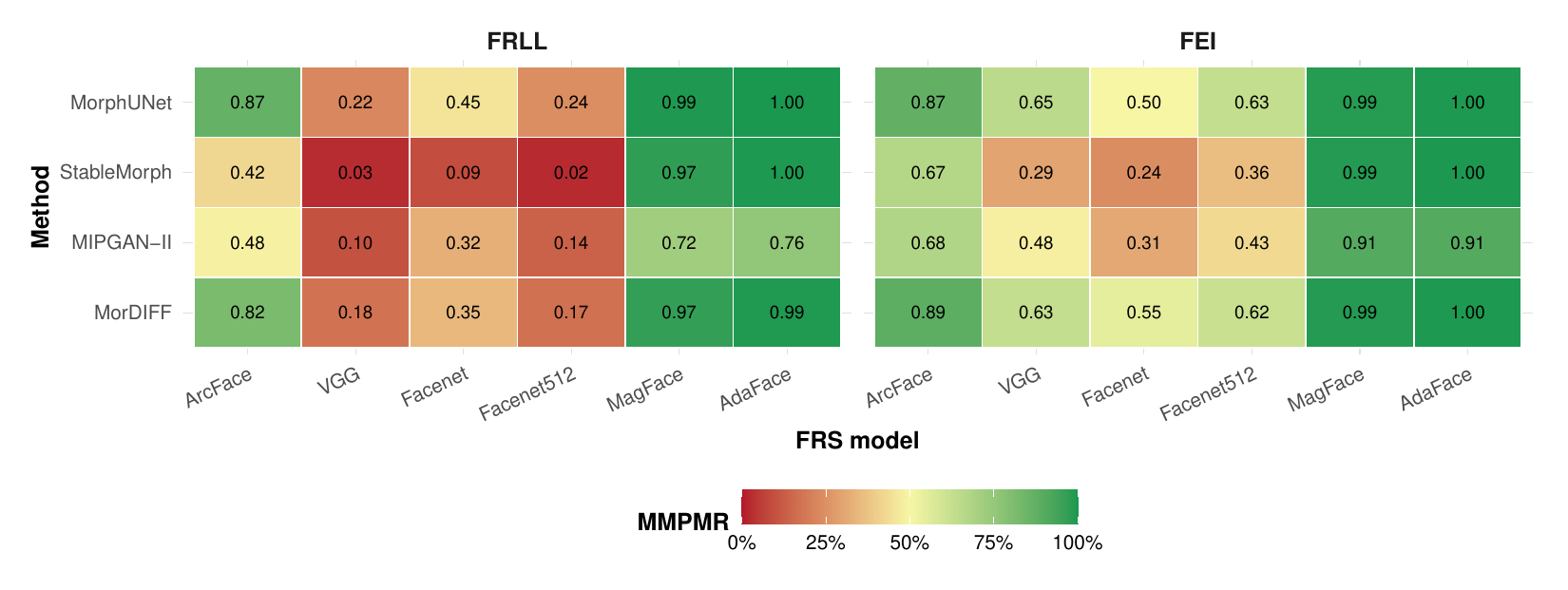}
	\caption{Per-FRS vulnerability. The heatmap reports \mmpmr{} for each method, dataset, and recognition system, showing where attack transferability is strongest or weakest.}
	\label{fig:per-frs-vulnerability}
\end{figure}

The aggregate \map{} scores are supported by per-matcher vulnerability in Figure~\ref{fig:per-frs-vulnerability}.
On FRLL, \morphunet{} achieves the highest average \mmpmr{} across the six matchers, with particularly strong transfer to ArcFace, MagFace, and AdaFace.
The lower VGG-Face and FaceNet scores show that the attack is not uniformly easy across all recognition architectures, which is precisely why the multi-FRS protocol is needed.
On FEI, \morphunet{} and MorDIFF are close in average \mmpmr{}, with MorDIFF slightly stronger on some matchers and \morphunet{} remaining highly competitive across the full set.
The per-FRS breakdown explains the behaviour of the stricter \map{} criteria.
High MagFace and AdaFace vulnerability alone is not sufficient for high $c=3$ or $c=6$; methods must also retain enough transfer to the more resistant matchers.
This is where the FRLL difference becomes important: \morphunet{} improves the weaker-matcher region relative to the other methods, increasing the number of matchers fooled by the same morph.

\subsubsection{Identity Preservation and Balance}
\label{sec:identity-results}
Strong morphing attacks should preserve both parents, not simply generate an image close to the easier identity.
\morphunet{} improves the minimum parent similarity on FRLL and remains close to the best FEI minimum similarity, while keeping identity imbalance low.
On FRLL, the method achieves a mean minimum similarity of 0.550 and an imbalance of 0.091, compared with 0.524 and 0.098 for MorDIFF.
On FEI, MorDIFF and \morphunet{} are nearly tied in identity balance, with both methods rounding to a minimum similarity of approximately 0.59 and both maintaining low imbalance.
\begin{figure}[htbp]
	\centering
	\includegraphics[width=0.49\linewidth]{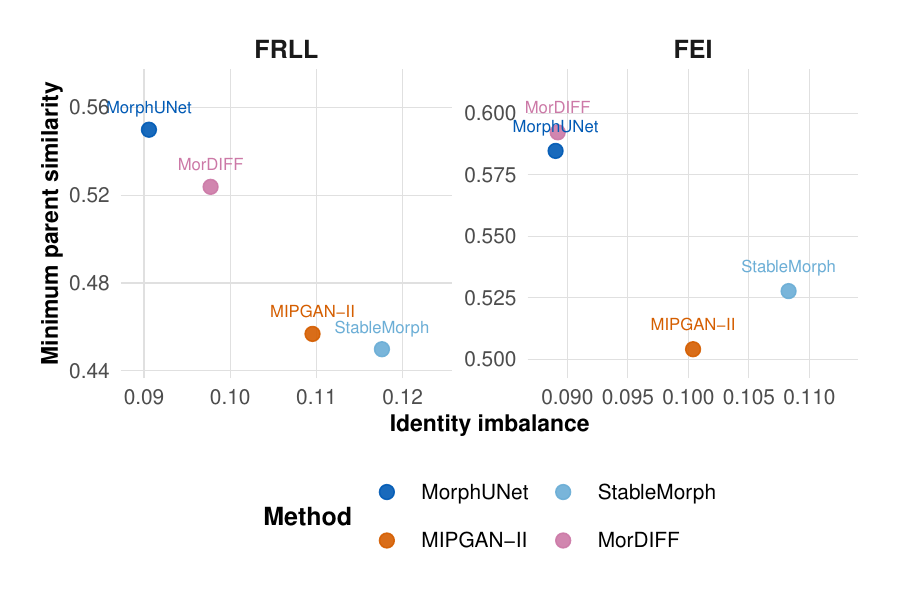}
	\includegraphics[width=0.49\linewidth]{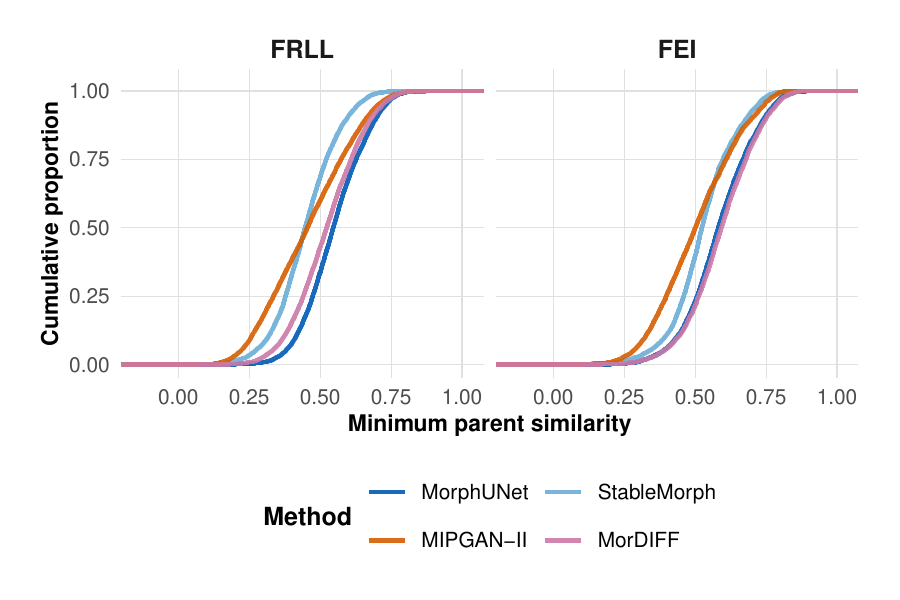}
	\caption{Identity preservation and balance. Top: scatter plot relating weaker-parent similarity to parent imbalance. Bottom: ECDF showing the distribution of minimum parent similarity across methods.}
	\label{fig:identity-balance}
\end{figure}

The ECDF and scatter plots in Figure~\ref{fig:identity-balance} show that this behaviour is distributional rather than a single aggregate effect: \morphunet{} tends to occupy the desirable region of high weaker-parent similarity and low imbalance.
This supports the central design motivation of Biometric Transport Layers: strong attack performance is obtained by preserving the weaker contributor, not by producing a visually plausible face dominated by one parent.

\subsection{Morphing Attack Detectability}
\label{sec:mad-results}

The MAD experiments evaluate a second primary attack axis: whether a trained morph detector identifies the generated images as attacks.
For this analysis, higher APCER at a fixed BPCER means the detector misses more attacks and the morphs are therefore harder to detect.
Table~\ref{tab:mad-summary} reports APCER at BPCER targets of 1\%, 5\%, 10\%, and 20\% for both same-dataset and cross-dataset detector settings.
The table specifies the detector training dataset and the evaluation dataset explicitly: FEI$\rightarrow$FEI and FRLL$\rightarrow$FRLL are same-dataset settings, while FRLL$\rightarrow$FEI and FEI$\rightarrow$FRLL test cross-dataset transfer.
\begin{table}[htbp]
	\caption{MAD APCER at fixed BPCER targets. Higher APCER means more attacks are accepted as bona fide by the detector and are therefore harder to detect. Training dataset indicates which auxiliary data was used to train the detector before evaluation on the final benchmark morphs. Bold indicates the best value within each training--evaluation block and metric, with ties included.}
	\label{tab:mad-summary}
	\centering
	\scriptsize
	\resizebox{\linewidth}{!}{%
	\begin{tabular}{lllrrrrr}
		\toprule
		Training dataset & Evaluation dataset & Method & \map{} $c=3$ ($\uparrow$) & APCER@1\% ($\uparrow$) & APCER@5\% ($\uparrow$) & APCER@10\% ($\uparrow$) & APCER@20\% ($\uparrow$) \\
		\midrule
		FEI & FEI & MIPGAN-II & 0.76 & 0.15 & 0.04 & 0.02 & 0.01 \\
		FEI & FEI & MorDIFF & 0.91 & 0.01 & 0.00 & 0.00 & 0.00 \\
		FEI & FEI & \morphunet{} & \textbf{0.92} & \textbf{0.55} & \textbf{0.29} & \textbf{0.18} & \textbf{0.13} \\
		FEI & FEI & StableMorph & 0.75 & 0.00 & 0.00 & 0.00 & 0.00 \\
		\midrule
		FRLL & FRLL & MIPGAN-II & 0.50 & 0.00 & 0.00 & 0.00 & \textbf{0.00} \\
		FRLL & FRLL & MorDIFF & 0.82 & 0.00 & 0.00 & 0.00 & \textbf{0.00} \\
		FRLL & FRLL & \morphunet{} & \textbf{0.89} & \textbf{0.92} & \textbf{0.04} & \textbf{0.01} & \textbf{0.00} \\
		FRLL & FRLL & StableMorph & 0.44 & 0.00 & 0.00 & 0.00 & \textbf{0.00} \\
		\midrule
		FRLL & FEI & MIPGAN-II & 0.76 & 0.01 & 0.00 & 0.00 & 0.00 \\
		FRLL & FEI & MorDIFF & 0.91 & 0.14 & 0.03 & 0.01 & 0.00 \\
		FRLL & FEI & \morphunet{} & \textbf{0.92} & \textbf{1.00} & \textbf{1.00} & \textbf{0.99} & \textbf{0.90} \\
		FRLL & FEI & StableMorph & 0.75 & 0.72 & 0.41 & 0.28 & 0.06 \\
		\midrule
		FEI & FRLL & MIPGAN-II & 0.50 & 0.96 & 0.84 & 0.59 & 0.46 \\
		FEI & FRLL & MorDIFF & 0.82 & 0.93 & 0.65 & 0.19 & 0.07 \\
		FEI & FRLL & \morphunet{} & \textbf{0.89} & \textbf{0.98} & \textbf{0.95} & \textbf{0.80} & \textbf{0.67} \\
		FEI & FRLL & StableMorph & 0.44 & 0.01 & 0.00 & 0.00 & 0.00 \\
		\bottomrule
	\end{tabular}%
	}
\end{table}

At 5\% BPCER in the same-dataset detector setting, \morphunet{} obtains APCER 0.293 on FEI and 0.041 on FRLL.
These values show that detector-side behaviour is not identical to cross-FRS attack success.
On FEI, \morphunet{} combines the best $c=3$ \map{} with the highest APCER among the high-\map{} methods, indicating a strong attack that is also comparatively harder to detect.
\begin{figure}[htbp]
	\centering
	\includegraphics[width=0.49\linewidth]{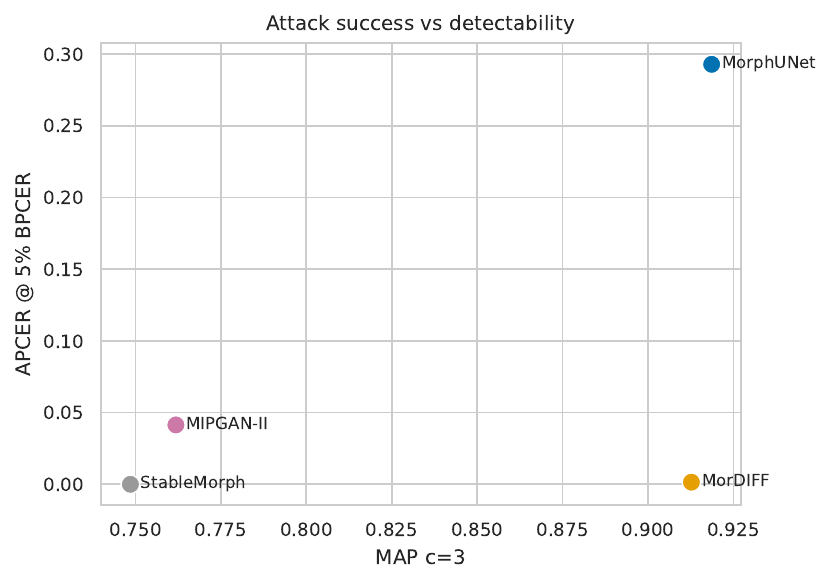}
	\includegraphics[width=0.49\linewidth]{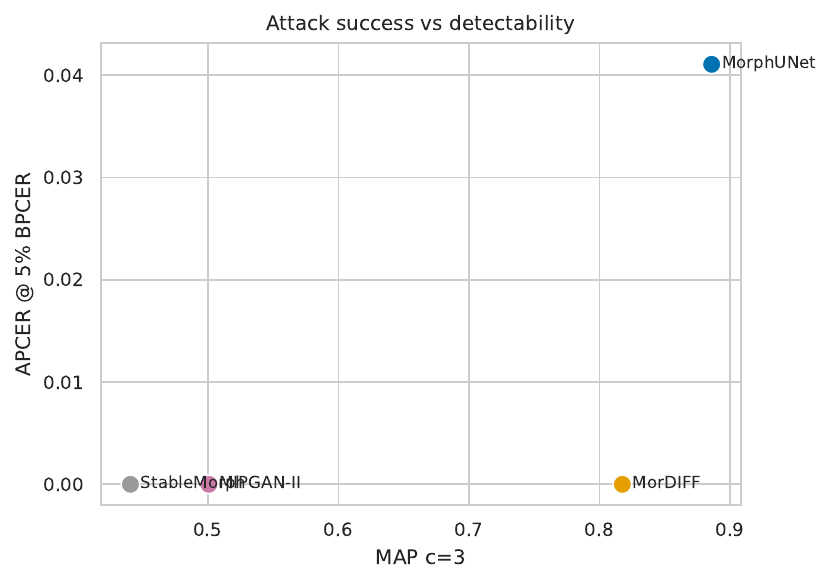}
	\caption{Attack success versus detectability. Left: FEI same-dataset detector setting. Right: FRLL same-dataset detector setting. The plots compare \map{} at $c=3$ with APCER at 5\% BPCER; higher APCER means harder to detect.}
	\label{fig:mad-results}
\end{figure}

On FRLL, the same-dataset detector is more sensitive to \morphunet{} at 5\% BPCER, even though \morphunet{} remains the strongest FRS attack.
The cross-dataset results are more severe: \morphunet{} remains highly difficult to detect when detector training and evaluation are drawn from different dataset families, with APCER 0.996 on FEI and 0.946 on FRLL at 5\% BPCER.
This contrast is important: MAP and MAD measure complementary failure modes, and a method can be strong against recognition systems while still varying substantially in detector visibility across operating points and domains.

Figure~\ref{fig:mad-results} plots attack success against detector-side detectability.
The separation between methods confirms that MAD should not be treated as a substitute for FRS attack success.
Instead, it is a second main evaluation axis: \map{} measures whether the morph is accepted by recognition systems, while MAD measures whether the morph is rejected by an attack detector.

\subsection{Image Quality and Realism}
\label{sec:quality-results}
\begin{table}[htbp]
	\centering
	\scriptsize
	\caption{Image quality metrics on FRLL and FEI. Lower values are better for all metrics. Bold indicates the best value per dataset and metric, with ties included.}
	\label{tab:image-quality-results}
	\resizebox{\linewidth}{!}{%
	\begin{tabular}{llrrrrrrr}
		\toprule
		Dataset & Method & FID ($\downarrow$) & KID ($\downarrow$) & CMMD ($\downarrow$) & LPIPS-A ($\downarrow$) & LPIPS-B ($\downarrow$) & Mean LPIPS ($\downarrow$) & LPIPS Imbal. ($\downarrow$) \\
		\midrule
		FRLL & \morphunet{} & \textbf{44.86} & \textbf{0.03} & 1.55 & \textbf{0.17} & \textbf{0.17} & \textbf{0.17} & 0.10 \\
		FRLL & StableMorph & 161.00 & 0.19 & \textbf{1.30} & 0.53 & 0.53 & 0.53 & 0.03 \\
		FRLL & MIPGAN-II & 80.69 & 0.08 & 3.67 & 0.27 & 0.27 & 0.27 & 0.06 \\
		FRLL & MorDIFF & 68.51 & 0.05 & 2.45 & 0.19 & 0.19 & 0.19 & \textbf{0.02} \\
		\midrule
		FEI & \morphunet{} & \textbf{35.19} & \textbf{0.02} & \textbf{0.24} & \textbf{0.19} & \textbf{0.19} & \textbf{0.19} & 0.07 \\
		FEI & StableMorph & 127.78 & 0.13 & 1.29 & 0.52 & 0.52 & 0.52 & 0.04 \\
		FEI & MIPGAN-II & 113.14 & 0.10 & 4.32 & 0.41 & 0.42 & 0.41 & 0.04 \\
		FEI & MorDIFF & 79.19 & 0.05 & 1.25 & \textbf{0.19} & \textbf{0.19} & \textbf{0.19} & \textbf{0.02} \\
		\bottomrule
	\end{tabular}%
	}
\end{table}
\morphunet{} obtains the best FID on both FRLL and FEI, with 44.86 on FRLL and 35.19 on FEI.
It also gives the best FRLL LPIPS mean and is essentially tied with MorDIFF on FEI LPIPS.
CMMD is strongest for StableMorph on FRLL, but StableMorph has substantially weaker \map{} and much worse FID and LPIPS, indicating that its CLIP distributional proximity does not translate into biometric attack strength.
Overall, \morphunet{} provides the clearest quality--attack compromise: high cross-FRS success, low perceptual distance to parents, and the strongest FID on both main datasets.
This matters because poor image quality can make a morph operationally implausible even if it fools an embedding model.
The same point is relevant to the FEI $c=6$ comparison: MorDIFF is slightly stronger under the all-matcher \map{} criterion, but \morphunet{} remains very competitive at $c=6$ while giving substantially better FID and CMMD on FEI, as reported in the image-quality analysis in Section~\ref{sec:quality-results}.
Qualitative inspection also matters in this regime, since local texture artefacts, isolated stray pixels, or boundary inconsistencies can reduce the practical acceptability of an otherwise high-\map{} morph.
Conversely, good visual quality alone is not sufficient, as shown by methods whose distributional scores improve without comparable cross-FRS attack transferability.

\begin{figure}[htbp]
	\centering
	\includegraphics[width=0.85\linewidth]{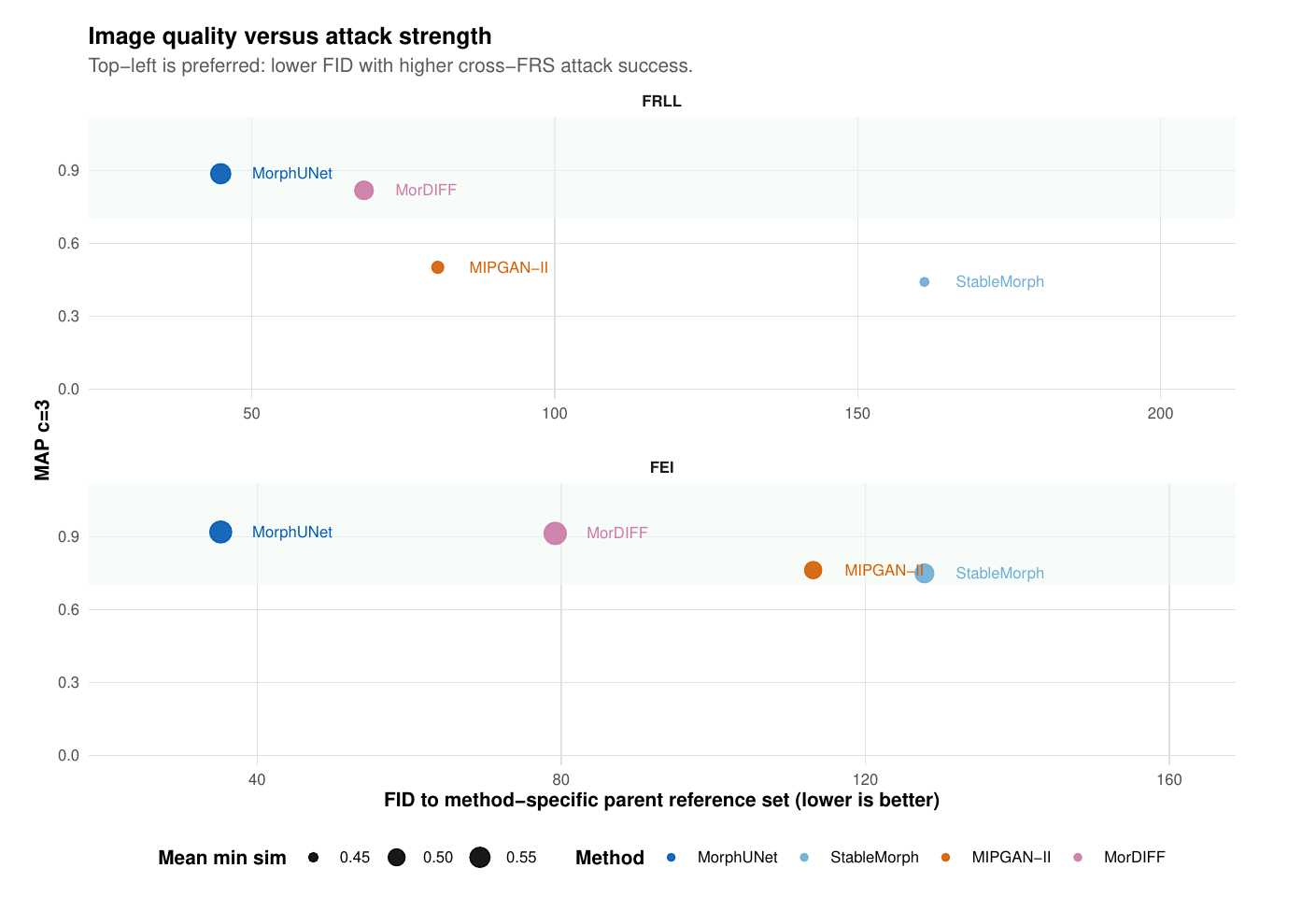}
	\caption{Image quality versus attack strength. The plot compares distributional realism against \map{} at $c=3$, highlighting whether a method improves visual realism without sacrificing biometric attack success.}
	\label{fig:quality-attack}
\end{figure}

\subsection{Gender and Similarity Stress Tests}
\label{sec:gender-similarity-results}
\begin{figure}[htbp]
	\centering
	\includegraphics[width=0.78\linewidth]{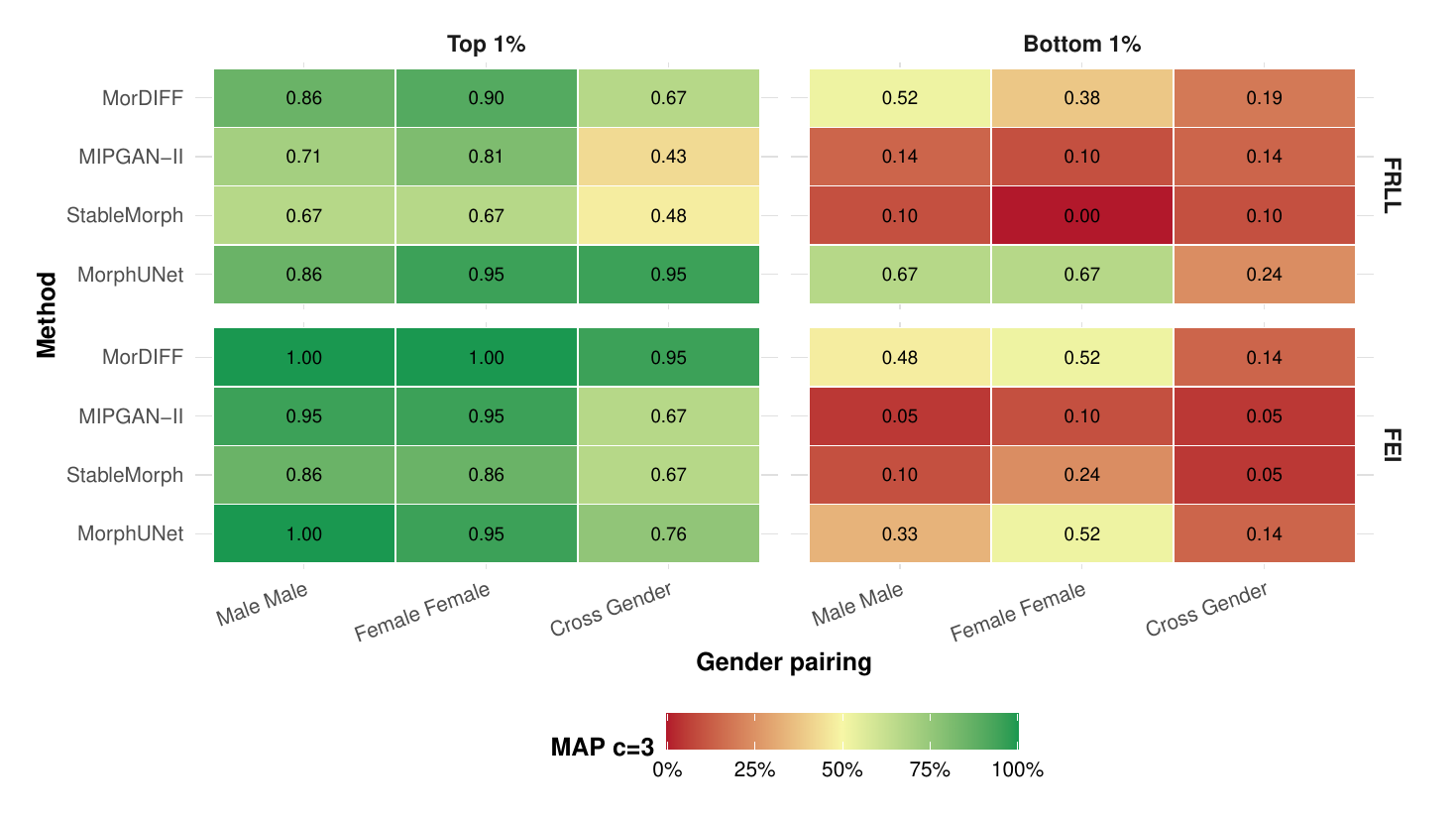}
	\caption{Gender and similarity stress-test heatmap on FEI and FRLL. The figure reports \map{} under gender-pair and parent-similarity buckets, showing how attack success changes for same-gender, cross-gender, top-similarity, and bottom-similarity parent pairs.}
	\label{fig:gender-stress-heatmap}
\end{figure}
The gender and similarity stress-test heatmap in Figure~\ref{fig:gender-stress-heatmap} evaluates whether the benchmark results are driven by easy parent pairs, while the compact top- and bottom-similarity bar summary is provided in the supplementary material.
The top-1\% parent-similarity subsets are expected to be easier because both parents are already close in identity space.
In these subsets, \morphunet{} is highly competitive across male--male, female--female, and cross-gender pairs.
The more informative case is the bottom-1\% subset, where the two parents are less similar and the morph must bridge a wider identity gap.
On FRLL, \morphunet{} remains strongest or tied strongest for same-gender bottom-similarity subsets and keeps the highest overall FRLL $c=3$ benchmark score.
On FEI, \morphunet{} and MorDIFF are closely matched in the most difficult bottom-similarity cases, while both remain ahead of the weaker baselines.
This close FEI stress-test behaviour should be interpreted together with the broader evaluation: an acceptable morphing attack is not determined by stress-test \map{} alone, but by the combined evidence from visual quality, parent-identity balance, detector-side behaviour, and cross-FRS transferability.

These stress tests show that \morphunet{}'s main benchmark performance is not only a result of favourable pair selection.
The method remains effective when parent similarity, gender pairing, and recognition model all vary, which supports the design choice of combining appearance trajectory information with explicit identity transport.
The bottom-1\% setting is particularly important because it tests the model where parent identity evidence is hardest to reconcile.
Maintaining competitive performance in this regime indicates that the alpha-parametrised transport path is not merely exploiting already-similar parent pairs.
The supplementary material expands this analysis with the compact top/bottom bar summary, the full stress-test table, additional robustness-drop visualisations, identity-imbalance distributions, and larger qualitative grids.
These supplementary results are used to check whether the main stress-test conclusions hold beyond the heatmap shown in the main text.

\subsection{CFD Unseen-Identity Robustness}
\label{sec:cfd-results}

\begin{figure}[htbp]
	\centering
	\includegraphics[width=0.78\linewidth]{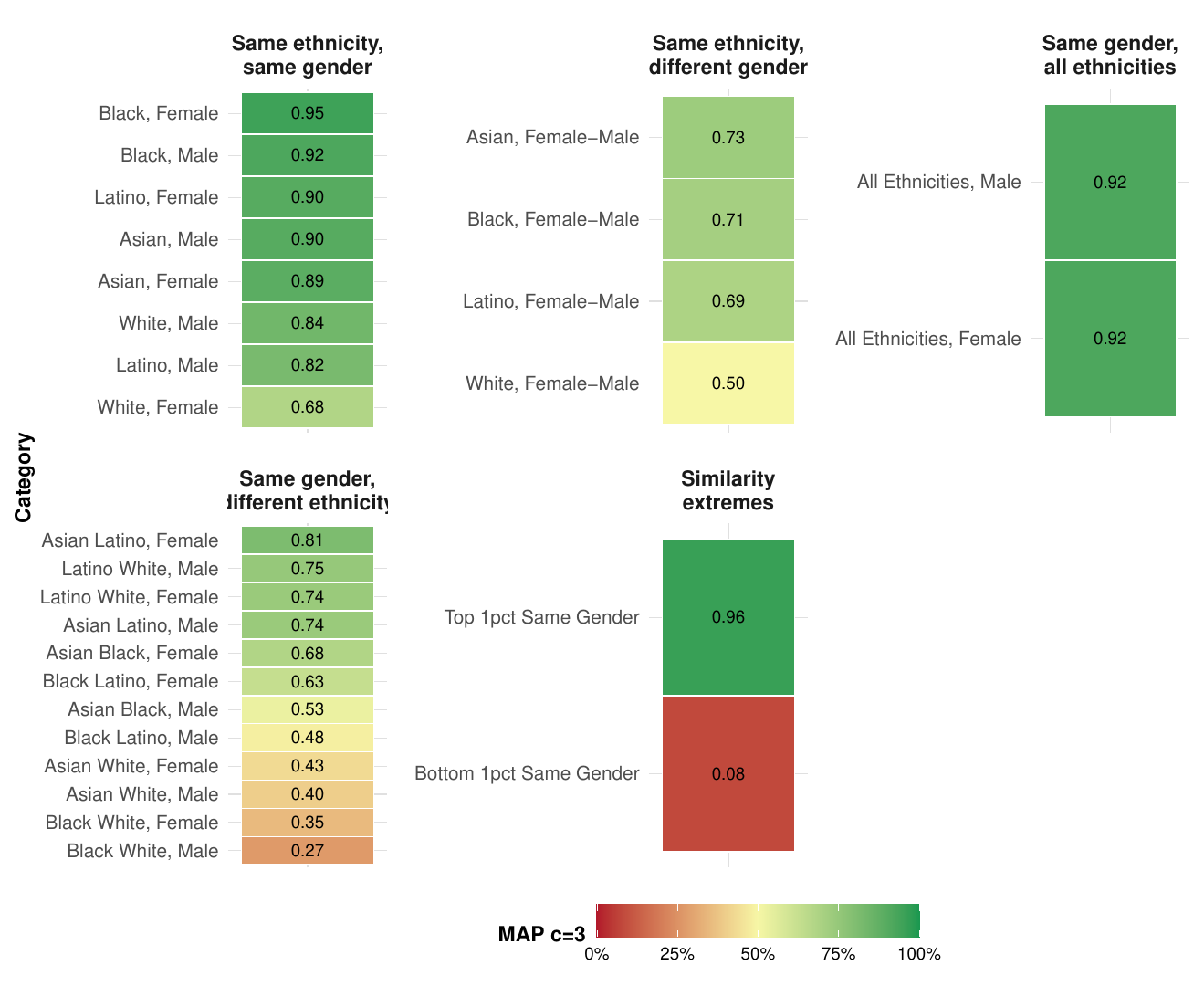}
	\caption{CFD unseen-identity robustness for \morphunet{} grouped by demographic relation and parent-pair difficulty. The heatmap shows how \map{} varies across broad category families and where demographic or similarity shifts reduce attack success.}
	\label{fig:cfd-category-family}
\end{figure}

CFD is used as a \morphunet{}-only robustness analysis rather than as a direct baseline comparison.
The goal is to test how the method behaves on identities and demographic pairings outside the primary FEI/FRLL benchmark.
Figure~\ref{fig:cfd-category-family} shows that \morphunet{} is strongest in the easier CFD category families and in high-similarity parent-pair regimes.
For example, the top-1\% same-gender similarity subset reaches high $c=3$ success, while the bottom-1\% subset is substantially harder.
Category shifts and low parent similarity reduce attack strength, indicating that unseen-identity morphing difficulty is not uniform across the CFD stress conditions.

The main CFD discussion focuses on the category-family heatmap and representative qualitative examples, while the supplementary material reports the stricter same-ethnicity/same-gender $c=4$ analysis and additional CFD qualitative variants.
This keeps the main text focused on the overall unseen-identity robustness trend, while the supplementary material records the detailed category-level evidence.
This result is useful because it separates two questions that are often conflated.
The FEI/FRLL experiments compare methods under the common final benchmark, while CFD asks whether \morphunet{} can generate plausible and attack-relevant morphs for unseen identities under controlled demographic shifts.
The observed degradation on low-similarity and cross-category pairs is expected, but the method remains successful across many CFD families, demonstrating robustness beyond the primary evaluation identities.
The CFD analysis therefore functions as an external stress test of the morphing mechanism rather than as a replacement for the main comparative benchmark.

\begin{figure}[htbp]
	\centering

	\begin{minipage}[c]{0.48\textwidth}
		\centering
		\includegraphics[width=\linewidth]{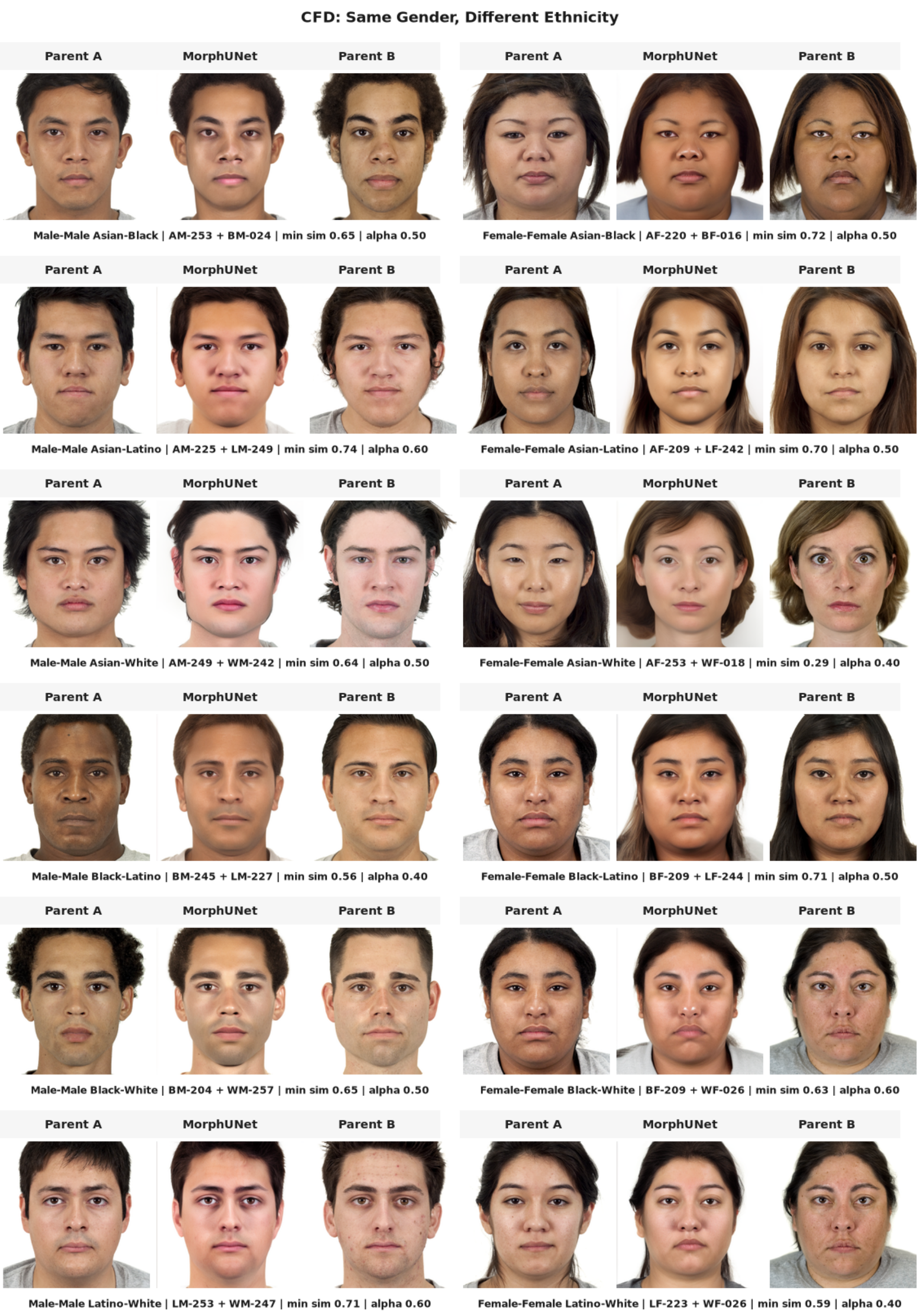}
	\end{minipage}
	\hfill
	\begin{minipage}[c]{0.48\textwidth}
		\centering
		\includegraphics[width=\linewidth]{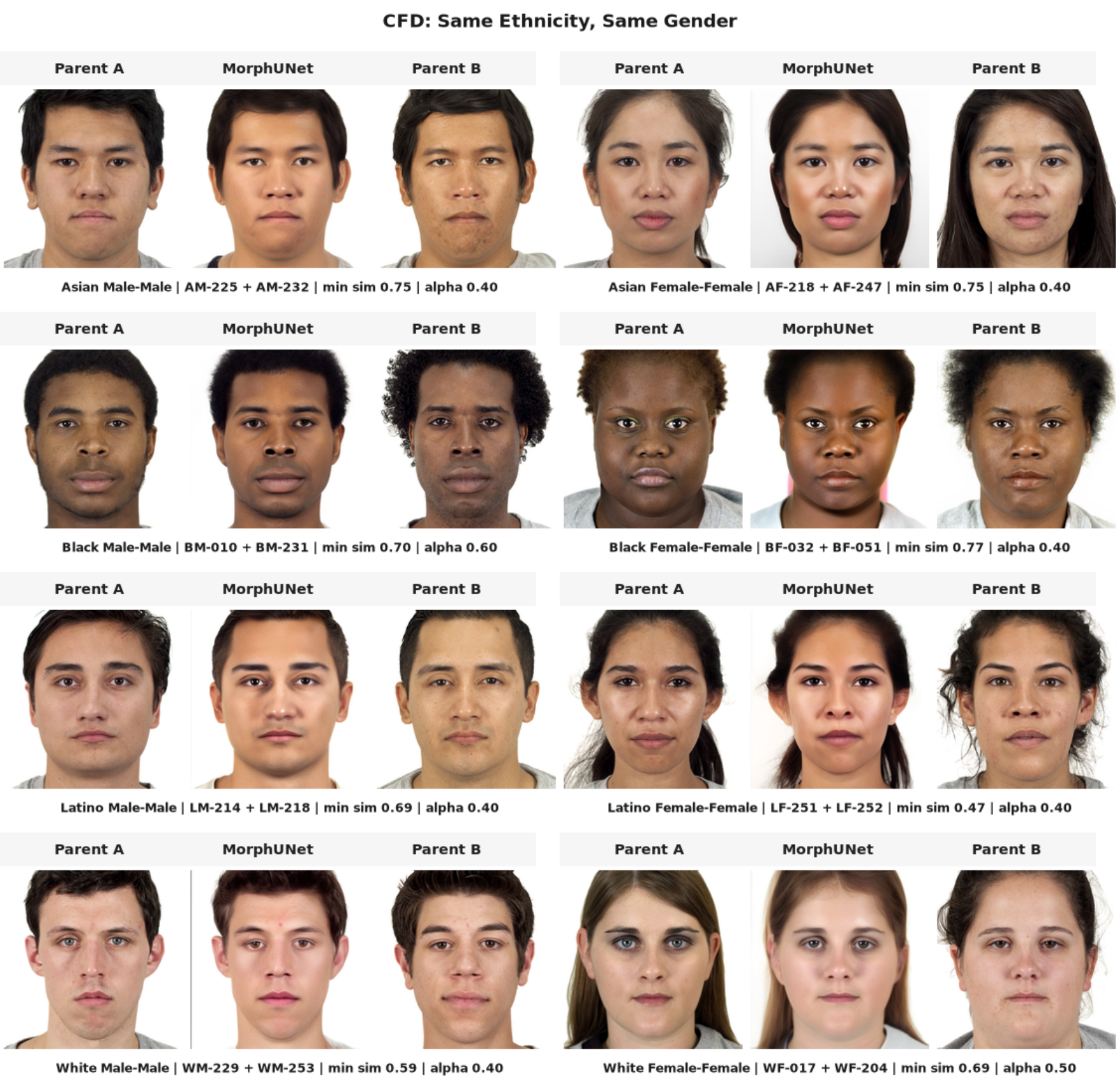}

		\vspace{0.4em}

		\includegraphics[width=\linewidth]{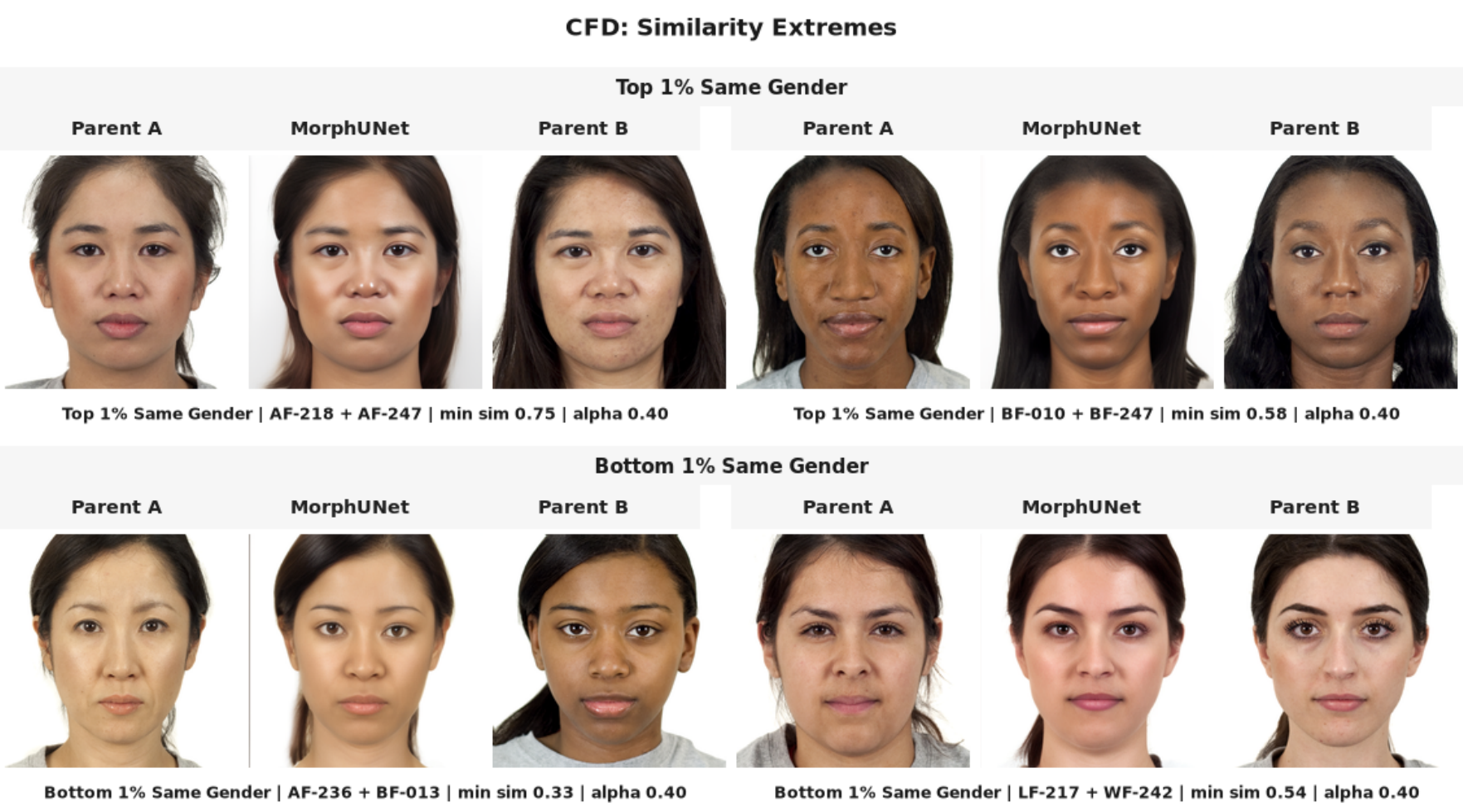}
	\end{minipage}

	\caption{Qualitative CFD stress examples for \morphunet{}. The grids show unseen-identity morphs under same-gender different-ethnicity, same-ethnicity same-gender, and parent-similarity extreme settings.}
	\label{fig:qualitative-cfd}
\end{figure}

Figure~\ref{fig:qualitative-cfd} gives qualitative CFD examples for unseen-identity stress conditions.
The selected cases include same-gender different-ethnicity, same-ethnicity same-gender, and similarity-extreme pairs.
These examples illustrate the qualitative counterpart to Figure~\ref{fig:cfd-category-family}: morphs become more challenging when the parents are farther apart in identity, gender, or demographic appearance, but \morphunet{} continues to produce coherent face images across the evaluated categories.

\section{Discussion}
\label{sec:discussion}

The experimental evidence supports three main observations.
First, morphing risk is poorly described by a single recognition model.
All evaluated methods can produce attacks that succeed under at least one matcher, but their behaviour separates when the acceptance criterion requires fooling multiple independent systems.
The gap between $c=1$, $c=3$, and $c=6$ therefore exposes whether a morph is merely exploiting one vulnerable embedding space or whether it carries identity evidence that transfers across recognition architectures.
Under this stricter view, \morphunet{} is consistently strong: it obtains the highest \map{} at $c=3$ on both FEI and FRLL, while also remaining competitive at the strict $c=6$ criterion.
At the same time, \map{} should be interpreted together with identity balance, qualitative inspection, MAD, and image quality, as reported in Sections~\ref{sec:identity-results}, \ref{sec:qualitative-results}, \ref{sec:mad-results}, and~\ref{sec:quality-results}, because biometric acceptance alone can overstate a method whose generated samples contain visible artefacts.

Second, the results show that attack strength and identity balance are tightly linked.
A morph can be visually realistic and still be weak if it preserves one parent much more strongly than the other.
The minimum parent similarity and imbalance analyses show that \morphunet{} does not obtain high \map{} by collapsing toward the easier identity.
Instead, the method improves or closely matches the best weaker-parent similarity while maintaining low imbalance.
This supports the design choice of keeping the two parent token banks separate inside the U-Net and mixing their transported residuals only after parent-specific attention has taken place.
In this sense, the Biometric Transport Layer is not simply another conditioning block; it is the mechanism that delays identity fusion until both parents have been individually represented in the denoising trajectory.

Third, the MAD experiments demonstrate that morph attack success and detector visibility are complementary, not interchangeable.
The same generated images can be strong under \map{} but differ substantially in APCER at fixed BPCER.
This is particularly clear in the cross-dataset detector setting, where \morphunet{} remains difficult to detect even when the detector is trained on the other dataset family.
The result suggests that the proposed morphs do not only transfer across recognition systems, but can also move outside the artefact distribution learned by a detector trained on a different source domain.
At the same time, the same-dataset MAD results show that detector sensitivity depends strongly on the operating point and dataset, so attack evaluation should report both FRS acceptance and MAD detectability rather than reducing morphing risk to one score.

The quality results complete this picture.
\morphunet{} achieves the best FID on both main datasets and low LPIPS to the parent images, indicating that its biometric strength is not obtained through obvious visual degradation.
This matters because a morphing attack must satisfy two constraints simultaneously: it must be acceptable as an image and it must preserve enough biometric evidence from both contributors.
Methods that optimise one side of this trade-off but not the other can look strong in a narrow metric, but are less convincing as complete morphing pipelines.
\morphunet{}'s strongest contribution is therefore the joint behaviour across \map{}, MAD, identity balance, and image quality.

\subsection{Limitations and Future Work}
\label{sec:limitations}

Several limitations remain.
First, the method depends on the quality of parent preprocessing and DDIM inversion.
Large pose, illumination, expression, or crop mismatches can make the parent trajectory harder to model and may reduce both identity balance and image quality.
Second, the current evaluation uses controlled benchmark imagery.
Although the FEI, FRLL, and CFD experiments cover different identities and stress conditions, further testing on operational document images, larger demographic coverage, lower-quality acquisition, and commercial recognition systems would be necessary before making deployment-level claims.
Third, the final morph choice is guided by recognition scores, so the selected output reflects the available scoring systems.
This is appropriate for measuring attack potential, but future work should study whether selection can be made less dependent on a fixed evaluator set.
Fourth, MAD behaviour is not monotonic with \map{}.
This is scientifically useful, but it also means that a complete threat model should include both recognition-system acceptance and detector rejection across several BPCER operating points.

More generally, future work on diffusion-based morphing should address three directions.
The first is broader domain evaluation, including unconstrained document-style imagery, additional unseen-identity datasets, and a wider range of acquisition conditions.
The second is detector-aware robustness analysis, where the goal is not simply to optimise against a specific MAD model, but to understand which visual or biometric cues make diffusion morphs detectable.
The third is improved controllability of the morph trajectory, particularly for difficult low-similarity parent pairs where identity evidence is harder to reconcile.

\section{Conclusion}
\label{sec:conclusion}

This paper presented \morphunet{}, an alpha-controlled biometric transport framework for diffusion-based face morphing attacks.
The method uses Biometric Token Alignment to map ArcFace identity evidence into a CLIP-compatible token space, preserves the two parents as separate identity-aware token banks, and injects them through Biometric Transport Layers during denoising.
DDIM-inverted latent blending provides structural continuity between the parents, while alpha-coupled identity transport controls how biometric evidence from each contributor is carried through the generation process.

Across FEI and FRLL, \morphunet{} achieves the strongest \map{} at $c=3$, reaching 0.919 on FEI and 0.886 on FRLL.
It also achieves the best FID on both datasets and maintains strong weaker-parent similarity with low identity imbalance.
The MAD analysis further shows that the method remains difficult to detect in cross-dataset detector settings, with APCER at 5\% BPCER of 0.996 on FEI and 0.946 on FRLL.
Together, these results show that \morphunet{} improves not only recognition-system attack success, but also the broader trade-off among identity balance, realism, and detector-side visibility.
The proposed biometric transport formulation therefore provides a strong and extensible basis for studying diffusion-based morphing attacks under realistic multi-system evaluation.

\bibliographystyle{unsrt}
\bibliography{references}

@article{thomaz2010fei,
  title={A new ranking method for principal components analysis and its application to face image analysis},
  author={Thomaz, Carlos E. and Giraldi, Gilson A.},
  journal={Image and Vision Computing},
  volume={28},
  number={6},
  pages={902--913},
  year={2010},
  doi={10.1016/j.imavis.2009.11.005}
}

@misc{debruine2017frll,
  title={Face Research Lab London Set},
  author={DeBruine, Lisa and Jones, Benedict},
  year={2017},
  publisher={figshare},
  url={https://figshare.com/articles/dataset/Face_Research_Lab_London_Set/5047666/5},
  doi={10.6084/m9.figshare.5047666.v5}
}

@article{ma2015chicago,
  title={The Chicago Face Database: A free stimulus set of faces and norming data},
  author={Ma, Debbie S. and Correll, Joshua and Wittenbrink, Bernd},
  journal={Behavior Research Methods},
  volume={47},
  number={4},
  pages={1122--1135},
  year={2015},
  doi={10.3758/s13428-014-0532-5}
}

@inproceedings{radford2021clip,
  title={Learning transferable visual models from natural language supervision},
  author={Radford, Alec and Kim, Jong Wook and Hallacy, Chris and Ramesh, Aditya and Goh, Gabriel and Agarwal, Sandhini and Sastry, Girish and Askell, Amanda and Mishkin, Pamela and Clark, Jack and Krueger, Gretchen and Sutskever, Ilya},
  booktitle={Proceedings of the 38th International Conference on Machine Learning},
  pages={8748--8763},
  year={2021},
  organization={PMLR}
}

@inproceedings{deng2019arcface,
  title={ArcFace: Additive angular margin loss for deep face recognition},
  author={Deng, Jiankang and Guo, Jia and Xue, Niannan and Zafeiriou, Stefanos},
  booktitle={Proceedings of the IEEE/CVF Conference on Computer Vision and Pattern Recognition},
  pages={4690--4699},
  year={2019},
  doi={10.1109/CVPR.2019.00482}
}

@inproceedings{ho2020ddpm,
  title={Denoising diffusion probabilistic models},
  author={Ho, Jonathan and Jain, Ajay and Abbeel, Pieter},
  booktitle={Advances in Neural Information Processing Systems},
  volume={33},
  pages={6840--6851},
  year={2020}
}

@inproceedings{song2020ddim,
  title={Denoising diffusion implicit models},
  author={Song, Jiaming and Meng, Chenlin and Ermon, Stefano},
  booktitle={International Conference on Learning Representations},
  year={2021}
}

@inproceedings{rombach2022ldm,
  title={High-resolution image synthesis with latent diffusion models},
  author={Rombach, Robin and Blattmann, Andreas and Lorenz, Dominik and Esser, Patrick and Ommer, Bj{\"o}rn},
  booktitle={Proceedings of the IEEE/CVF Conference on Computer Vision and Pattern Recognition},
  pages={10684--10695},
  year={2022}
}

@inproceedings{liu2015celeba,
  title={Deep learning face attributes in the wild},
  author={Liu, Ziwei and Luo, Ping and Wang, Xiaogang and Tang, Xiaoou},
  booktitle={Proceedings of the IEEE International Conference on Computer Vision},
  pages={3730--3738},
  year={2015}
}

@inproceedings{parkhi2015vggface,
  title={Deep face recognition},
  author={Parkhi, Omkar M. and Vedaldi, Andrea and Zisserman, Andrew},
  booktitle={British Machine Vision Conference},
  year={2015}
}

@inproceedings{schroff2015facenet,
  title={FaceNet: A unified embedding for face recognition and clustering},
  author={Schroff, Florian and Kalenichenko, Dmitry and Philbin, James},
  booktitle={Proceedings of the IEEE Conference on Computer Vision and Pattern Recognition},
  pages={815--823},
  year={2015}
}

@inproceedings{meng2021magface,
  title={MagFace: A universal representation for face recognition and quality assessment},
  author={Meng, Qiang and Zhao, Shichao and Huang, Zhida and Zhou, Feng},
  booktitle={Proceedings of the IEEE/CVF Conference on Computer Vision and Pattern Recognition},
  pages={14225--14234},
  year={2021}
}

@inproceedings{kim2022adaface,
  title={AdaFace: Quality adaptive margin for face recognition},
  author={Kim, Minchul and Jain, Anil K. and Liu, Xiaoming},
  booktitle={Proceedings of the IEEE/CVF Conference on Computer Vision and Pattern Recognition},
  pages={18750--18759},
  year={2022}
}

@inproceedings{heusel2017gans,
  title={GANs trained by a two time-scale update rule converge to a local Nash equilibrium},
  author={Heusel, Martin and Ramsauer, Hubert and Unterthiner, Thomas and Nessler, Bernhard and Hochreiter, Sepp},
  booktitle={Advances in Neural Information Processing Systems},
  volume={30},
  year={2017}
}

@inproceedings{binkowski2018demystifying,
  title={Demystifying MMD GANs},
  author={Bi{\'n}kowski, Miko{\l}aj and Sutherland, Danica J. and Arbel, Michael and Gretton, Arthur},
  booktitle={International Conference on Learning Representations},
  year={2018}
}

@inproceedings{zhang2018lpips,
  title={The unreasonable effectiveness of deep features as a perceptual metric},
  author={Zhang, Richard and Isola, Phillip and Efros, Alexei A. and Shechtman, Eli and Wang, Oliver},
  booktitle={Proceedings of the IEEE Conference on Computer Vision and Pattern Recognition},
  pages={586--595},
  year={2018}
}

@article{jayasumana2024rethinking,
  title={Rethinking FID: Towards a better evaluation metric for image generation},
  author={Jayasumana, Sadeep and Ramalingam, Srikumar and Veit, Andreas and Glasner, Daniel and Chakrabarti, Ayan and Kumar, Sanjiv},
  journal={arXiv preprint arXiv:2401.09603},
  year={2024}
}

@misc{iso30107,
  title={Information technology -- Biometric presentation attack detection -- Part 3: Testing and reporting},
  author={{International Organization for Standardization}},
  howpublished={ISO/IEC 30107-3:2017},
  year={2017}
}

@inproceedings{he2016resnet,
  title={Deep residual learning for image recognition},
  author={He, Kaiming and Zhang, Xiangyu and Ren, Shaoqing and Sun, Jian},
  booktitle={Proceedings of the IEEE Conference on Computer Vision and Pattern Recognition},
  pages={770--778},
  year={2016}
}

@inproceedings{dosovitskiy2021vit,
  title={An image is worth 16x16 words: Transformers for image recognition at scale},
  author={Dosovitskiy, Alexey and Beyer, Lucas and Kolesnikov, Alexander and Weissenborn, Dirk and Zhai, Xiaohua and Unterthiner, Thomas and Dehghani, Mostafa and Minderer, Matthias and Heigold, Georg and Gelly, Sylvain and Uszkoreit, Jakob and Houlsby, Neil},
  booktitle={International Conference on Learning Representations},
  year={2021}
}

@inproceedings{kashiani2022robust,
  title={Robust ensemble morph detection with domain generalization},
  author={Kashiani, Hossein and Sami, Shoaib Meraj and Soleymani, Sobhan and Nasrabadi, Nasser M.},
  booktitle={2022 IEEE International Joint Conference on Biometrics},
  year={2022},
  doi={10.48550/arXiv.2209.08130}
}

@inproceedings{kingma2014vae,
  title={Auto-Encoding Variational Bayes},
  author={Kingma, Diederik P. and Welling, Max},
  booktitle={International Conference on Learning Representations},
  year={2014}
}

@inproceedings{preechakul2022diffae,
  title={Diffusion autoencoders: Toward a meaningful and decodable representation},
  author={Preechakul, Konpat and Chatthee, Nattanat and Wizadwongsa, Suttisak and Suwajanakorn, Supasorn},
  booktitle={Proceedings of the IEEE/CVF Conference on Computer Vision and Pattern Recognition},
  pages={10619--10629},
  year={2022}
}

@inproceedings{ferrara2014magicpassport,
  title={The magic passport},
  author={Ferrara, Matteo and Franco, Annalisa and Maltoni, Davide},
  booktitle={2014 IEEE International Joint Conference on Biometrics},
  pages={1--7},
  year={2014},
  organization={IEEE},
  doi={10.1109/BTAS.2014.6996240}
}

@inproceedings{makrushin2017visapp,
  title={Automatic generation and detection of visually faultless facial morphs},
  author={Makrushin, Andrey and Neubert, Tom and Dittmann, Jana},
  booktitle={Proceedings of the 12th International Joint Conference on Computer Vision, Imaging and Computer Graphics Theory and Applications},
  pages={39--50},
  year={2017},
  doi={10.5220/0006131100390050}
}

@article{robertson2019facemorphing,
  title={Face morphing attacks: Investigating detection with humans and computers},
  author={Robertson, David J. and Kramer, Robin S. S. and Burton, A. Mike},
  journal={Cognitive Research: Principles and Implications},
  volume={4},
  number={1},
  pages={1--12},
  year={2019},
  doi={10.1186/s41235-019-0181-4}
}

@article{scherhag2019survey,
  title={Face recognition systems under morphing attacks: A survey},
  author={Scherhag, Ulrich and Rathgeb, Christian and Merkle, Johannes and Breithaupt, Ralph and Busch, Christoph},
  journal={IEEE Access},
  volume={7},
  pages={23012--23026},
  year={2019},
  doi={10.1109/ACCESS.2019.2899367}
}

@inproceedings{karras2019stylegan,
  title={A style-based generator architecture for generative adversarial networks},
  author={Karras, Tero and Laine, Samuli and Aila, Timo},
  booktitle={Proceedings of the IEEE/CVF Conference on Computer Vision and Pattern Recognition},
  pages={4401--4410},
  year={2019}
}

@inproceedings{karras2020stylegan2,
  title={Analyzing and improving the image quality of StyleGAN},
  author={Karras, Tero and Laine, Samuli and Aittala, Miika and Hellsten, Janne and Lehtinen, Jaakko and Aila, Timo},
  booktitle={Proceedings of the IEEE/CVF Conference on Computer Vision and Pattern Recognition},
  pages={8110--8119},
  year={2020}
}

@article{zhang2021mipgan,
  title={MIPGAN--Generating strong and high quality morphing attacks using identity prior driven GAN},
  author={Zhang, Haoyu and Venkatesh, Sushma and Ramachandra, Raghavendra and Raja, Kiran and Damer, Naser and Busch, Christoph},
  journal={IEEE Transactions on Biometrics, Behavior, and Identity Science},
  volume={3},
  number={3},
  pages={365--383},
  year={2021},
  doi={10.1109/TBIOM.2021.3072349}
}

@inproceedings{kabbani2025stablemorph,
  title={StableMorph: High-Quality Face Morph Generation with Stable Diffusion},
  author={Kabbani, Wassim and Raja, Kiran and Ramachandra, Raghavendra and Busch, Christoph},
  booktitle={2025 IEEE International Joint Conference on Biometrics (IJCB)},
  year={2025},
  doi={10.1109/IJCB65343.2025.11411174}
}

@article{damer2021regenmorph,
  title={ReGenMorph: Visibly realistic GAN generated face morphing attacks by attack re-generation},
  author={Damer, Naser and Raja, Kiran and S{\"u}{\ss}milch, Marius and Venkatesh, Sushma and Boutros, Fadi and Fang, Meiling and Kirchbuchner, Florian and Ramachandra, Raghavendra and Kuijper, Arjan},
  journal={arXiv preprint arXiv:2108.09130},
  year={2021}
}

@article{venkatesh2020ganmorphs,
  title={Vulnerability analysis of face morphing attacks from landmarks and generative adversarial networks},
  author={Venkatesh, Sushma and Zhang, Haoyu and Raja, Kiran and Ramachandra, Raghavendra and Busch, Christoph},
  journal={arXiv preprint arXiv:2012.05344},
  year={2020}
}

@inproceedings{damer2023mordiff,
  title={MorDIFF: Recognition vulnerability and attack detectability of face morphing attacks created by diffusion autoencoders},
  author={Damer, Naser and Fang, Meiling and Siebke, Patrick and Kolf, Jan Niklas and Huber, Marco and Boutros, Fadi},
  booktitle={International Workshop on Biometrics and Forensics},
  year={2023},
  doi={10.48550/arXiv.2302.01843}
}

@article{grimmer2024ladimo,
  title={LADIMO: Face morph generation through biometric template inversion with latent diffusion},
  author={Grimmer, Matthias and Boutros, Fadi and Fang, Meiling and Damer, Naser},
  journal={arXiv preprint arXiv:2410.07988},
  year={2024}
}

@article{ye2023ipadapter,
  title={IP-Adapter: Text compatible image prompt adapter for text-to-image diffusion models},
  author={Ye, Hu and Zhang, Jun and Liu, Sibo and Han, Xiao and Yang, Wei},
  journal={arXiv preprint arXiv:2308.06721},
  year={2023}
}

@article{wang2024instantid,
  title={InstantID: Zero-shot identity-preserving generation in seconds},
  author={Wang, Qixun and Bai, Xu and Wang, Haofan and Qin, Zekui and Chen, Anthony},
  journal={arXiv preprint arXiv:2401.07519},
  year={2024}
}

@inproceedings{gal2023textualinversion,
  title={An Image is Worth One Word: Personalizing Text-to-Image Generation using Textual Inversion},
  author={Gal, Rinon and Alaluf, Yuval and Atzmon, Yuval and Patashnik, Or and Bermano, Amit H. and Chechik, Gal and Cohen-Or, Daniel},
  booktitle={International Conference on Learning Representations},
  year={2023}
}

@inproceedings{ruiz2023dreambooth,
  title={DreamBooth: Fine Tuning Text-to-Image Diffusion Models for Subject-Driven Generation},
  author={Ruiz, Nataniel and Li, Yuanzhen and Jampani, Varun and Pritch, Yael and Rubinstein, Michael and Aberman, Kfir},
  booktitle={Proceedings of the IEEE/CVF Conference on Computer Vision and Pattern Recognition},
  pages={22500--22510},
  year={2023}
}

@article{hu2021lora,
  title={LoRA: Low-Rank Adaptation of Large Language Models},
  author={Hu, Edward J. and Shen, Yelong and Wallis, Phillip and Allen-Zhu, Zeyuan and Li, Yuanzhi and Wang, Shean and Wang, Lu and Chen, Weizhu},
  journal={arXiv preprint arXiv:2106.09685},
  year={2021}
}

@article{zhang2023diffmorpher,
  title={DiffMorpher: Unleashing the Capability of Diffusion Models for Image Morphing},
  author={Zhang, Kaiwen and Zhou, Yifan and Xu, Xudong and Pan, Xingang and Dai, Bo},
  journal={arXiv preprint arXiv:2312.07409},
  year={2023}
}

\clearpage
\begin{center}
  {\LARGE\bfseries Supplementary Material}
\end{center}
\vspace{1em}

\appendix

\section{Additional Robustness and Qualitative Results}
\label{sec:appendix-additional-results}

This supplemental material extends the main evaluation with stricter robustness views, full stress-test values, identity-balance distributions, and additional qualitative examples.
The results show how the reported attack strength behaves when the evaluation criterion becomes harder, when parent pairs are less similar, and when \morphunet{} is tested on unseen CFD identities.
The supplement is organised as an evidence audit rather than as a gallery of extra figures.
Each group of plots is interpreted with respect to one question: whether the main-paper conclusions remain stable under stricter cross-FRS criteria, difficult parent-pair subsets, unseen-identity CFD stress conditions, and simplified conditioning variants.
The qualitative examples are therefore used to support the metric interpretation by showing whether the reported attack strength corresponds to visually plausible morphs, rather than to samples with obvious local artefacts or one-parent collapse.

\FloatBarrier
\subsection{Cross-FRS Robustness}
\label{sec:appendix-cross-frs-robustness}

\begin{figure}[htbp]
	\centering
	\includegraphics[width=0.78\linewidth]{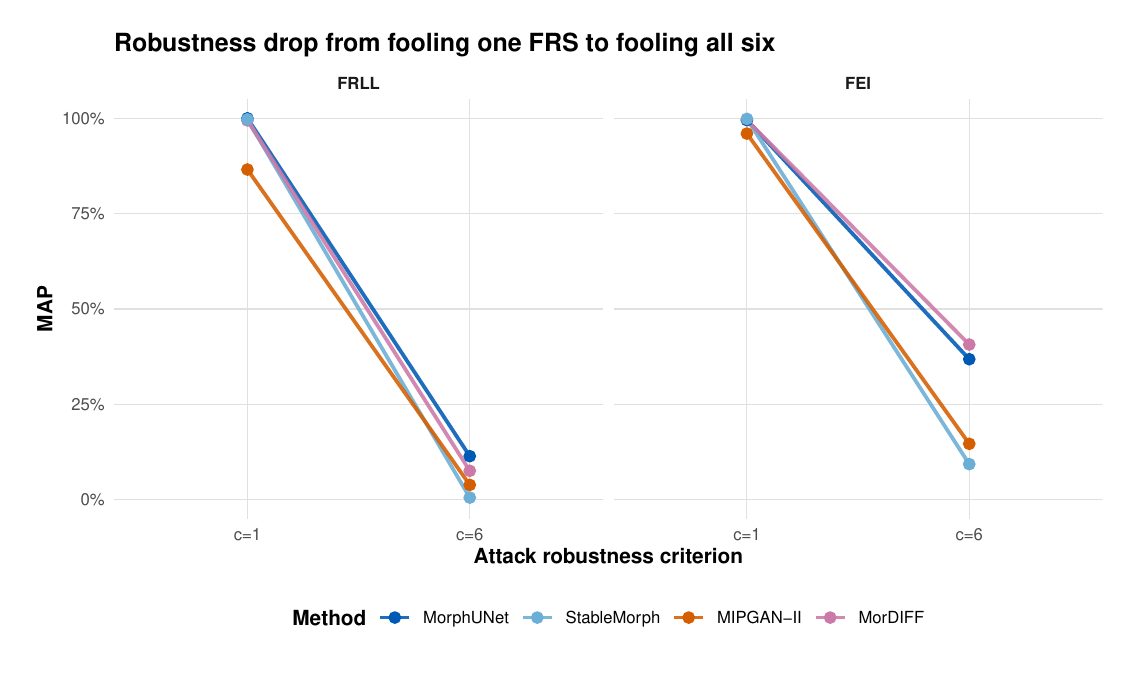}
	\caption{Robustness drop from fooling one FRS to fooling all six FRS systems. The plot summarises how much each method degrades when the attack criterion is tightened from $c=1$ to $c=6$.}
	\label{fig:appendix-robustness-drop}
\end{figure}

Figure~\ref{fig:appendix-robustness-drop} gives a stricter view of the main \map{} results.
A high value at $c=1$ shows that a morph can fool at least one matcher, but this is a weak condition because it may reflect vulnerability in only one recognition space.
The drop toward $c=6$ shows how much identity evidence survives when the same morph must transfer across all six systems.
This supports the use of $c=3$ as the main criterion, since it captures cross-FRS transfer without relying on the extreme all-matcher case.
The main interpretation is that robustness should be read from the shape of the degradation curve, not only from a single endpoint.
Methods that remain high at $c=1$ but fall rapidly at stricter criteria are useful for showing single-matcher vulnerability, but they provide weaker evidence of transferable two-parent identity preservation.
\morphunet{} is designed for the intermediate-to-strict region: it keeps the strongest main $c=3$ behaviour while remaining competitive at $c=6$, which is consistent with the delayed parent-specific transport mechanism used in the U-Net.
This is also why the main paper interprets \map{} together with identity balance, MAD, and image quality rather than treating the all-matcher endpoint as the only indicator of morph quality.

\FloatBarrier
\subsection{Gender and Parent-Similarity Stress Tests}
\label{sec:appendix-gender-similarity-stress}

\begin{figure}[htbp]
	\centering
	\includegraphics[width=0.48\linewidth]{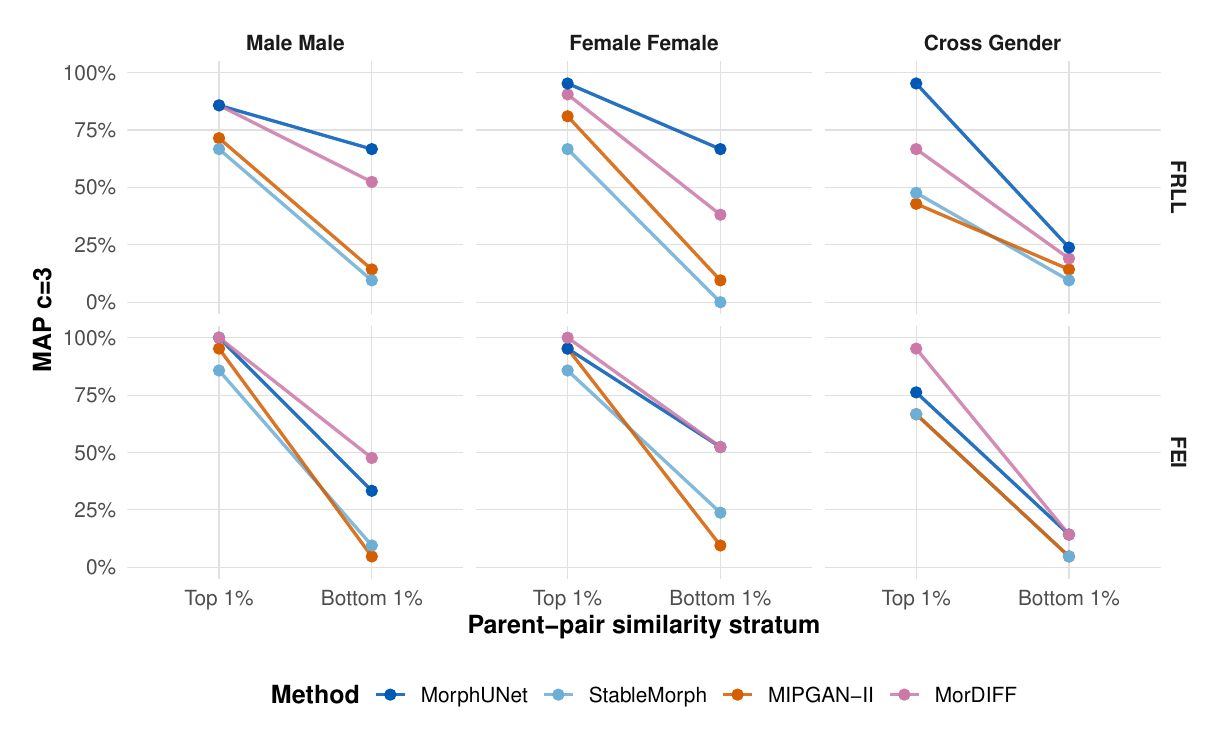}
	\hfill
	\includegraphics[width=0.48\linewidth]{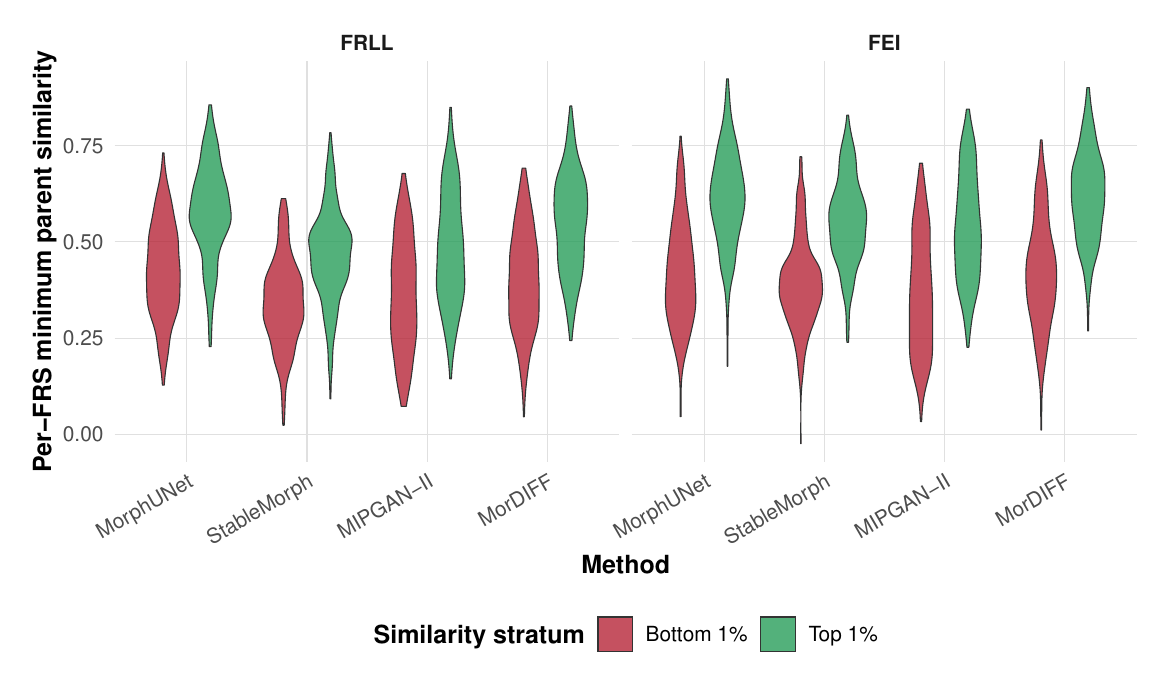}
	\caption{Additional gender and parent-similarity stress-test views. Left: method behaviour across the top- and bottom-similarity stress buckets. Right: distribution of parent-pair similarity values used for the stress-test splits.}
	\label{fig:appendix-similarity-stress-extra}
\end{figure}

\begin{figure}[htbp]
	\centering
	\includegraphics[width=0.55\linewidth]{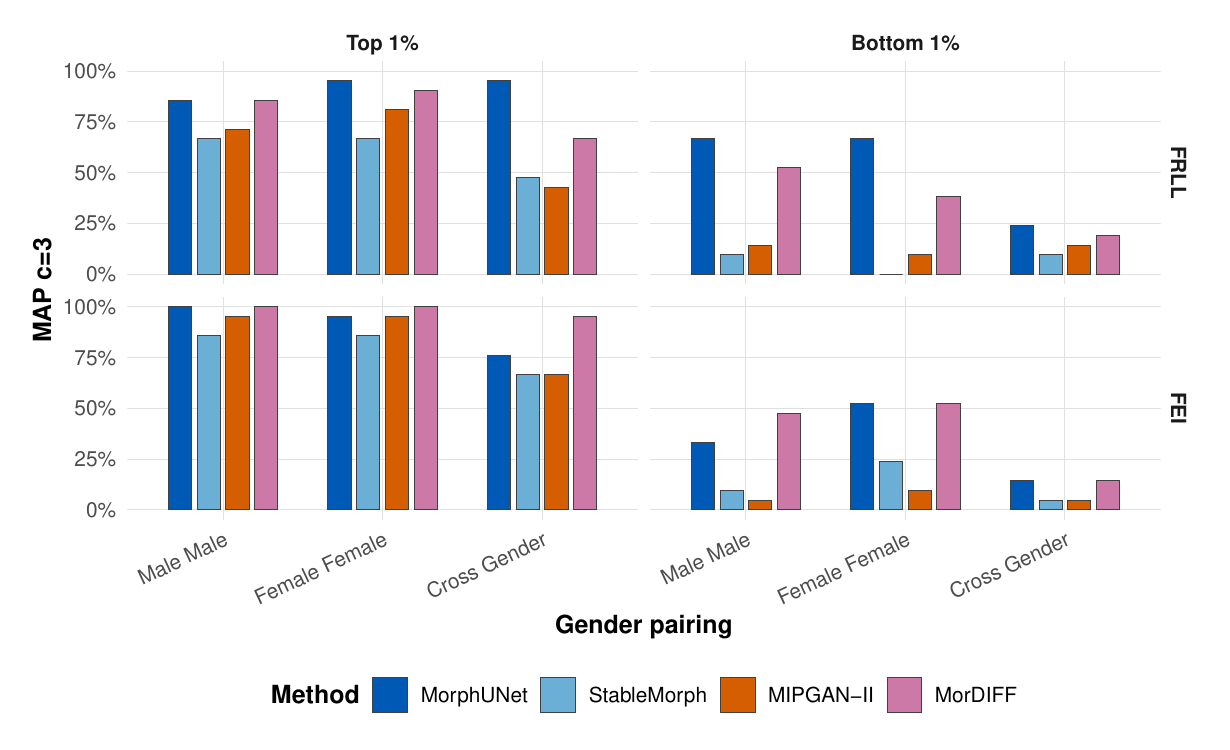}
	\caption{Top- and bottom-similarity stress-test summary on FEI and FRLL. The bars compare \map{} across gender-pair categories for the easiest parent pairs from the top 1\% similarity bucket and the hardest parent pairs from the bottom 1\% similarity bucket.}
	\label{fig:gender-stress-bars}
\end{figure}

The stress-test plots show that parent-pair difficulty has a clear effect on morphing success.
Top-similarity pairs are easier because the contributors are already close in the recognition embedding space.
Bottom-similarity pairs are more demanding because the generated image must preserve two more distant identities.
The stronger performance of \morphunet{} in these harder settings supports the conclusion that the method is not only benefiting from favourable parent pairs.
The right-hand similarity distribution is important for interpreting the bars: the stress subsets are not arbitrary small samples, but the extremes of the parent-pair similarity distribution.
Consequently, a drop from top to bottom buckets should be expected for all methods.
The useful question is whether a method degrades gracefully and whether the weaker parent remains represented when the identity gap is large.

\begin{figure}[htbp]
	\centering
	\includegraphics[width=0.74\linewidth]{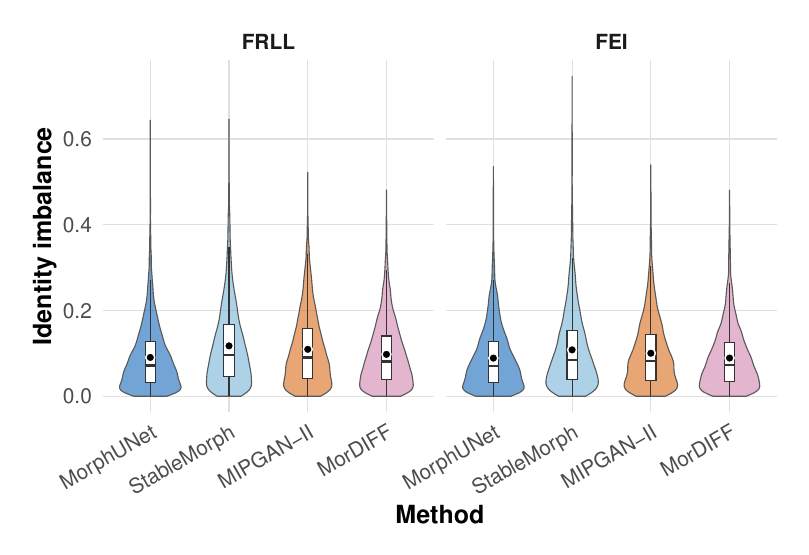}
	\caption{Identity-imbalance distribution on FEI and FRLL. Lower values indicate that a generated morph represents both parents more evenly rather than collapsing toward one contributor.}
	\label{fig:appendix-identity-imbalance-distribution}
\end{figure}

Figure~\ref{fig:appendix-identity-imbalance-distribution} gives a distributional view of parent balance.
A morph can obtain strong average similarity while still being dominated by one parent.
Lower imbalance indicates that the weaker contributor is preserved more consistently, supporting the role of the Biometric Transport Layers in maintaining separate parent evidence before alpha-controlled mixing.

\begin{table}[htbp]
\centering
\scriptsize
\caption{Full updated gender and parent-similarity stress-test table. The \map{} $c=3$ values show that \morphunet{} is the strongest method in most FRLL stress buckets and remains one of the two strongest methods on FEI.}
\label{tab:gender_similarity_stress}
\resizebox{\linewidth}{!}{%
\begin{tabular}{lrrrrrrrr}
\toprule
Dataset & Bucket & Method & \#Morphs & MAP c=1 ($\uparrow$) & MAP c=3 ($\uparrow$) & MAP c=6 ($\uparrow$) & Mean Min Sim ($\uparrow$) & Imbalance ($\downarrow$) \\
\midrule
FRLL & Bottom 1\%: Cross Gender & MorphUNet & 21.00 & 0.86 & \textbf{0.24} & \textbf{0.00} & \textbf{0.38} & \textbf{0.13} \\
FRLL & Bottom 1\%: Cross Gender & StableMorph & 21.00 & \textbf{0.90} & 0.10 & 0.00 & 0.28 & 0.26 \\
FRLL & Bottom 1\%: Cross Gender & MIPGAN-II & 21.00 & 0.24 & 0.14 & 0.00 & 0.33 & 0.13 \\
FRLL & Bottom 1\%: Cross Gender & MorDIFF & 21.00 & 0.86 & 0.19 & 0.00 & 0.34 & 0.17 \\
FRLL & Bottom 1\%: Female Female & MorphUNet & 21.00 & \textbf{1.00} & \textbf{0.67} & \textbf{0.00} & \textbf{0.46} & \textbf{0.11} \\
FRLL & Bottom 1\%: Female Female & StableMorph & 21.00 & 0.95 & 0.00 & 0.00 & 0.39 & 0.12 \\
FRLL & Bottom 1\%: Female Female & MIPGAN-II & 21.00 & 0.57 & 0.10 & 0.00 & 0.38 & 0.14 \\
FRLL & Bottom 1\%: Female Female & MorDIFF & 21.00 & 0.90 & 0.38 & 0.00 & 0.43 & 0.12 \\
FRLL & Bottom 1\%: Male Male & MorphUNet & 21.00 & \textbf{0.90} & \textbf{0.67} & \textbf{0.00} & \textbf{0.45} & \textbf{0.09} \\
FRLL & Bottom 1\%: Male Male & StableMorph & 21.00 & 0.90 & 0.10 & 0.00 & 0.36 & 0.13 \\
FRLL & Bottom 1\%: Male Male & MIPGAN-II & 21.00 & 0.62 & 0.14 & 0.00 & 0.39 & 0.13 \\
FRLL & Bottom 1\%: Male Male & MorDIFF & 21.00 & 0.90 & 0.52 & 0.00 & 0.43 & 0.09 \\
FRLL & Top 1\%: Cross Gender & MorphUNet & 21.00 & \textbf{1.00} & \textbf{0.95} & 0.05 & \textbf{0.55} & \textbf{0.10} \\
FRLL & Top 1\%: Cross Gender & StableMorph & 21.00 & 1.00 & 0.48 & 0.05 & 0.42 & 0.22 \\
FRLL & Top 1\%: Cross Gender & MIPGAN-II & 21.00 & 0.95 & 0.43 & 0.05 & 0.45 & 0.14 \\
FRLL & Top 1\%: Cross Gender & MorDIFF & 21.00 & 1.00 & 0.67 & \textbf{0.10} & 0.50 & 0.12 \\
FRLL & Top 1\%: Female Female & MorphUNet & 21.00 & \textbf{1.00} & \textbf{0.95} & \textbf{0.19} & \textbf{0.61} & \textbf{0.08} \\
FRLL & Top 1\%: Female Female & StableMorph & 21.00 & 1.00 & 0.67 & 0.05 & 0.51 & 0.10 \\
FRLL & Top 1\%: Female Female & MIPGAN-II & 21.00 & 0.95 & 0.81 & 0.14 & 0.49 & 0.09 \\
FRLL & Top 1\%: Female Female & MorDIFF & 21.00 & 1.00 & 0.90 & 0.19 & 0.59 & 0.10 \\
FRLL & Top 1\%: Male Male & MorphUNet & 21.00 & \textbf{1.00} & \textbf{0.86} & \textbf{0.38} & \textbf{0.57} & 0.10 \\
FRLL & Top 1\%: Male Male & StableMorph & 21.00 & 1.00 & 0.67 & 0.00 & 0.46 & 0.12 \\
FRLL & Top 1\%: Male Male & MIPGAN-II & 21.00 & 0.95 & 0.71 & 0.24 & 0.51 & \textbf{0.09} \\
FRLL & Top 1\%: Male Male & MorDIFF & 21.00 & 1.00 & 0.86 & 0.14 & 0.55 & 0.11 \\
\addlinespace[0.35em]
\midrule
FEI & Bottom 1\%: Cross Gender & MorphUNet & 21.00 & 0.57 & \textbf{0.14} & \textbf{0.00} & \textbf{0.36} & 0.14 \\
FEI & Bottom 1\%: Cross Gender & StableMorph & 21.00 & \textbf{0.95} & 0.05 & 0.00 & 0.33 & 0.18 \\
FEI & Bottom 1\%: Cross Gender & MIPGAN-II & 21.00 & 0.24 & 0.05 & 0.00 & 0.32 & 0.14 \\
FEI & Bottom 1\%: Cross Gender & MorDIFF & 21.00 & 0.57 & 0.14 & 0.00 & 0.35 & \textbf{0.12} \\
FEI & Bottom 1\%: Female Female & MorphUNet & 21.00 & 0.95 & \textbf{0.52} & \textbf{0.10} & \textbf{0.47} & \textbf{0.10} \\
FEI & Bottom 1\%: Female Female & StableMorph & 21.00 & \textbf{1.00} & 0.24 & 0.00 & 0.43 & 0.10 \\
FEI & Bottom 1\%: Female Female & MIPGAN-II & 21.00 & 0.43 & 0.10 & 0.00 & 0.39 & 0.12 \\
FEI & Bottom 1\%: Female Female & MorDIFF & 21.00 & 0.95 & 0.52 & 0.10 & 0.47 & 0.13 \\
FEI & Bottom 1\%: Male Male & MorphUNet & 21.00 & 0.90 & 0.33 & \textbf{0.10} & 0.43 & \textbf{0.11} \\
FEI & Bottom 1\%: Male Male & StableMorph & 21.00 & \textbf{1.00} & 0.10 & 0.00 & 0.41 & 0.12 \\
FEI & Bottom 1\%: Male Male & MIPGAN-II & 21.00 & 0.57 & 0.05 & 0.00 & 0.40 & 0.12 \\
FEI & Bottom 1\%: Male Male & MorDIFF & 21.00 & 0.95 & \textbf{0.48} & 0.00 & \textbf{0.45} & 0.13 \\
FEI & Top 1\%: Cross Gender & MorphUNet & 21.00 & \textbf{1.00} & 0.76 & 0.29 & 0.55 & 0.13 \\
FEI & Top 1\%: Cross Gender & StableMorph & 21.00 & 1.00 & 0.67 & 0.10 & 0.50 & 0.14 \\
FEI & Top 1\%: Cross Gender & MIPGAN-II & 21.00 & 0.95 & 0.67 & 0.10 & 0.49 & \textbf{0.10} \\
FEI & Top 1\%: Cross Gender & MorDIFF & 21.00 & 1.00 & \textbf{0.95} & \textbf{0.38} & \textbf{0.56} & \textbf{0.10} \\
FEI & Top 1\%: Female Female & MorphUNet & 21.00 & \textbf{1.00} & 0.95 & 0.52 & \textbf{0.65} & \textbf{0.08} \\
FEI & Top 1\%: Female Female & StableMorph & 21.00 & 1.00 & 0.86 & 0.24 & 0.59 & 0.10 \\
FEI & Top 1\%: Female Female & MIPGAN-II & 21.00 & 1.00 & 0.95 & 0.29 & 0.54 & 0.09 \\
FEI & Top 1\%: Female Female & MorDIFF & 21.00 & 1.00 & \textbf{1.00} & \textbf{0.57} & 0.64 & 0.09 \\
FEI & Top 1\%: Male Male & MorphUNet & 21.00 & \textbf{1.00} & \textbf{1.00} & \textbf{0.52} & 0.64 & 0.08 \\
FEI & Top 1\%: Male Male & StableMorph & 21.00 & 1.00 & 0.86 & 0.24 & 0.57 & 0.08 \\
FEI & Top 1\%: Male Male & MIPGAN-II & 21.00 & 1.00 & 0.95 & 0.43 & 0.58 & \textbf{0.07} \\
FEI & Top 1\%: Male Male & MorDIFF & 21.00 & 1.00 & 1.00 & 0.52 & \textbf{0.66} & 0.08 \\
\bottomrule
\end{tabular}%
}
\end{table}

Table~\ref{tab:gender_similarity_stress} reports the full stress-test values behind the compact plots.
It separates the effect of gender pairing from the effect of parent-similarity difficulty.
This makes it easier to identify where performance remains stable and where morph generation becomes harder because the two parent identities are less compatible.
On FRLL, \morphunet{} is strongest at $c=3$ in all bottom-similarity categories, including the difficult female--female and male--male subsets where it reaches 0.67.
It also keeps the best weaker-parent similarity in the same FRLL bottom-similarity categories, showing that the gain is not merely caused by shifting toward one easier identity.
On FEI, the hardest bottom-similarity cases are closer: \morphunet{} ties MorDIFF in bottom cross-gender and female--female $c=3$, while MorDIFF is stronger in bottom male--male $c=3$.
This mixed FEI behaviour is consistent with the main-paper interpretation: \morphunet{} and MorDIFF are both strong in some FEI high-strength regimes, but \morphunet{} provides a stronger overall quality--attack compromise when the full evaluation is considered.
The stress table is therefore useful because it exposes where the method is clearly separated, where it is tied, and where another baseline remains competitive.

\FloatBarrier
\subsection{Gender and Parent-Similarity Qualitative Examples}
\label{sec:appendix-gender-similarity-qualitative}

\begin{figure}[htbp]
	\centering

	\begin{minipage}[t]{0.49\textwidth}
		\centering
		\includegraphics[
			width=\linewidth,
			height=0.72\textheight,
			keepaspectratio
		]{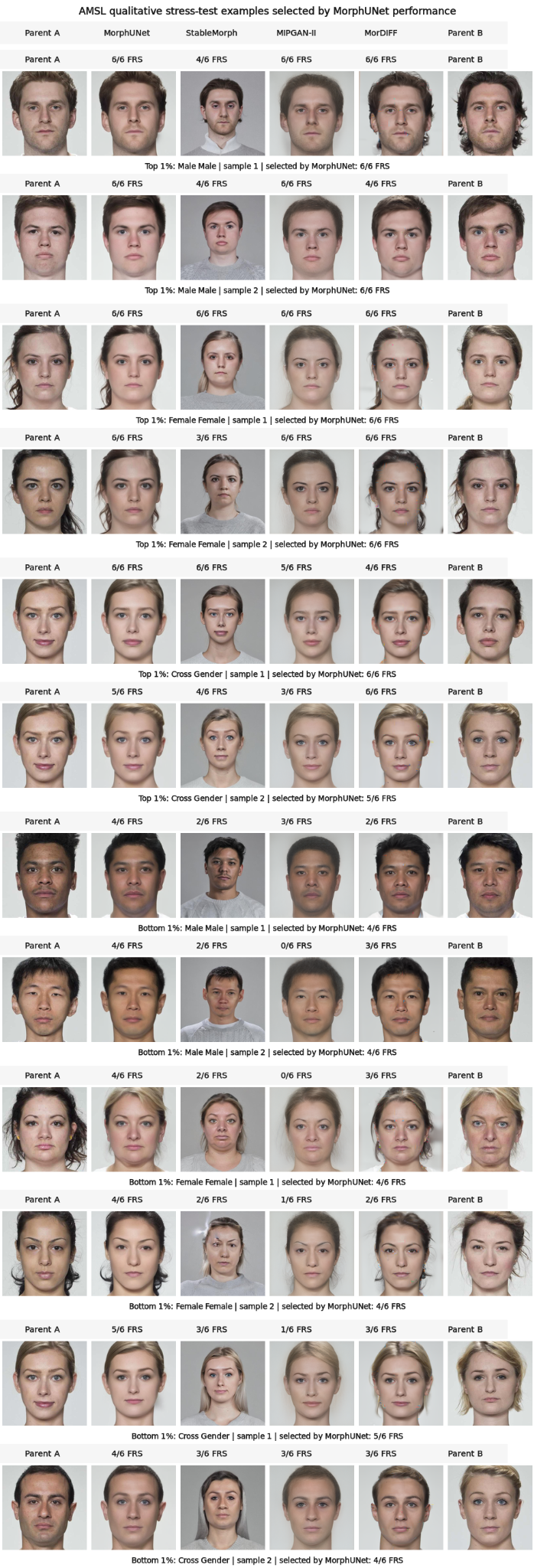}

		\vspace{0.2em}
		{\scriptsize\textbf{(a)} FRLL examples.}
	\end{minipage}
	\hfill
	\begin{minipage}[t]{0.49\textwidth}
		\centering
		\includegraphics[
			width=\linewidth,
			height=0.72\textheight,
			keepaspectratio
		]{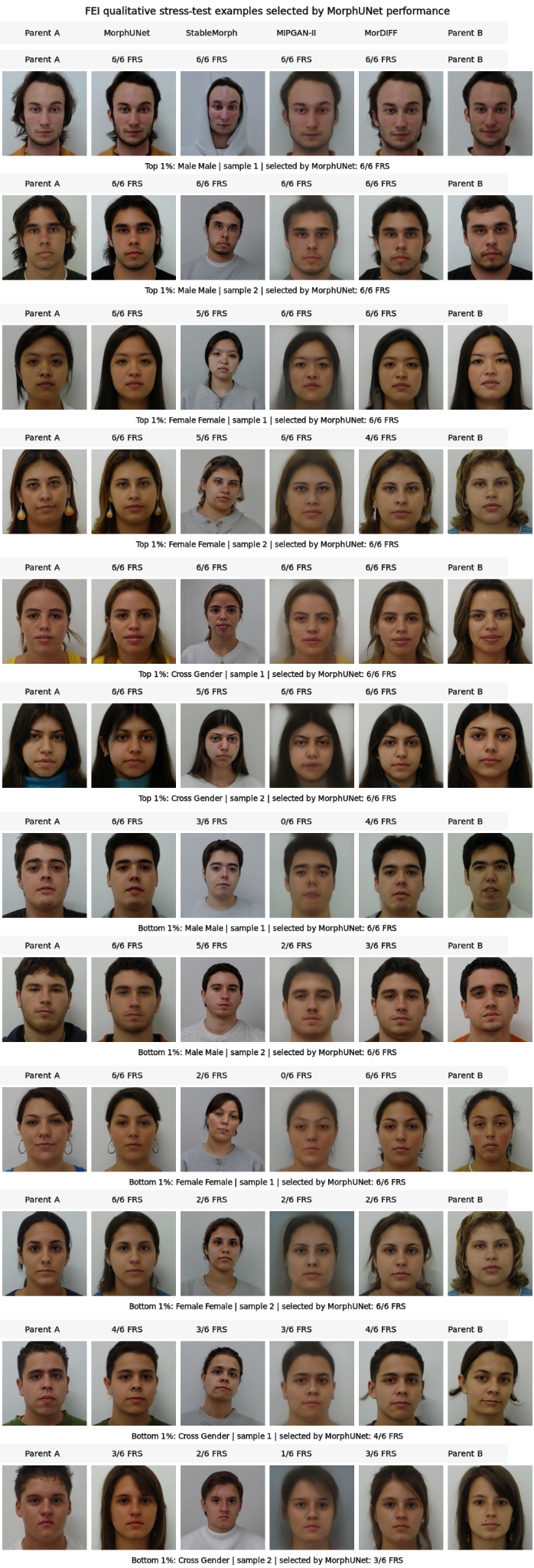}

		\vspace{0.2em}
		{\scriptsize\textbf{(b)} FEI examples.}
	\end{minipage}

	\caption{Qualitative gender and similarity stress-test examples for \morphunet{}. The FRLL and FEI examples show visually plausible morphs under easier top-similarity cases and more difficult bottom-similarity cases.}
	\label{fig:appendix-qualitative-amsl-fei}
\end{figure}

The qualitative stress-test examples support the metric-level analysis.
The generated morphs remain visually plausible in many difficult settings, but harder bottom-similarity pairs show stronger pressure on weaker-parent preservation.
This is visible in subtle changes around the eyes, mouth, and face outline, where the generated image can become closer to one parent even when the overall image remains realistic.
These grids should be read together with the stress metrics rather than as standalone visual examples.
The top-similarity cases mainly check that the model can preserve image realism when the identity trajectory is easier.
The bottom-similarity cases are more diagnostic: they reveal whether the model preserves coherent facial structure while still carrying evidence from both parents.
When a sample looks realistic but one parent's periocular or mouth-region traits become weak, the visual evidence explains why the minimum-similarity and imbalance metrics are necessary.

\FloatBarrier
\subsection{CFD Robustness Details}
\label{sec:appendix-cfd-details}

The CFD evaluation tests unseen identities under controlled demographic and similarity shifts.
The results show that \morphunet{} generalises beyond the FEI and FRLL identities, but the difficulty is not uniform across all CFD pairings.
Attack strength is highest when parent pairs are closer in demographic and identity structure, and lower when ethnicity, gender, or identity distance introduces a larger gap between the contributors.

\begin{table}[htbp]
	\centering
	\scriptsize
	\caption{CFD same-ethnicity same-gender robustness using \map{} $c=4$. This stricter criterion gives a more demanding category-level view than the main $c=3$ setting.}
	\label{tab:appendix-cfd-same-ethnicity-c4}
	\resizebox{\linewidth}{!}{%
	\begin{tabular}{lrrrr}
		\toprule
		CFD category & \#Morphs & \map{} $c=3$ ($\uparrow$) & \map{} $c=4$ ($\uparrow$) & \map{} $c=6$ ($\uparrow$) \\
		\midrule
		Asian Female & 100 & 0.89 & 0.81 & 0.42 \\
		Asian Male & 100 & 0.90 & 0.80 & 0.40 \\
		Black Female & 100 & 0.95 & 0.84 & 0.46 \\
		Black Male & 100 & 0.92 & 0.77 & 0.41 \\
		Latino Female & 100 & 0.90 & 0.66 & 0.16 \\
		Latino Male & 100 & 0.82 & 0.61 & 0.17 \\
		White Female & 100 & 0.68 & 0.41 & 0.07 \\
		White Male & 100 & 0.84 & 0.46 & 0.10 \\
		\bottomrule
	\end{tabular}
	}
\end{table}

The stricter $c=4$ CFD results show that several same-ethnicity same-gender categories retain strong multi-FRS transfer.
Asian, Black, and Latino subsets remain particularly strong under this criterion, while the drop toward $c=6$ confirms that all-system transfer is still much harder.
The category-level variation also shows why CFD robustness should be interpreted from the full category pattern rather than from a single aggregate score.
The table should not be read as a demographic ranking of face-recognition vulnerability.
Its purpose is narrower: it stress-tests whether the learned morphing mechanism transfers to unseen identities under controlled pairing categories.
The stronger $c=4$ values for Asian and Black same-gender subsets, together with the lower White subsets, indicate that CFD difficulty depends on the particular parent-pair distribution, identity spacing, and visual compatibility present in each category.
The important conclusion is therefore not that one demographic group is intrinsically easier, but that unseen-domain morphing has structured failure modes that are hidden by a single aggregate CFD score.

\begin{figure}[htbp]
	\centering
	\begin{minipage}[t]{0.48\linewidth}
		\centering
		\includegraphics[width=\linewidth]{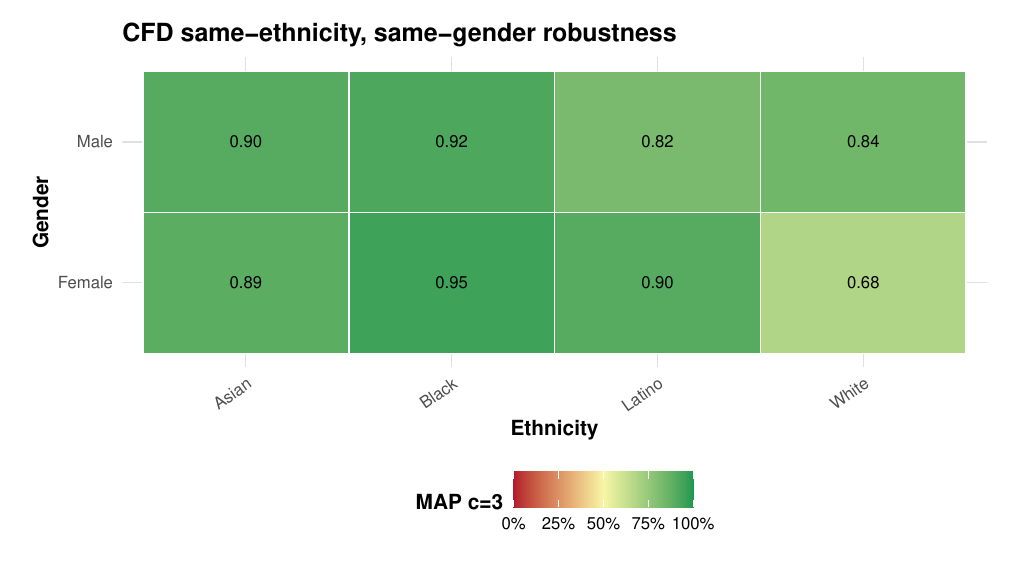}
		\vspace{0.35em}
		{\scriptsize\textbf{(a)} Same-ethnicity same-gender.}
	\end{minipage}
	\hfill
	\begin{minipage}[t]{0.48\linewidth}
		\centering
		\includegraphics[width=\linewidth]{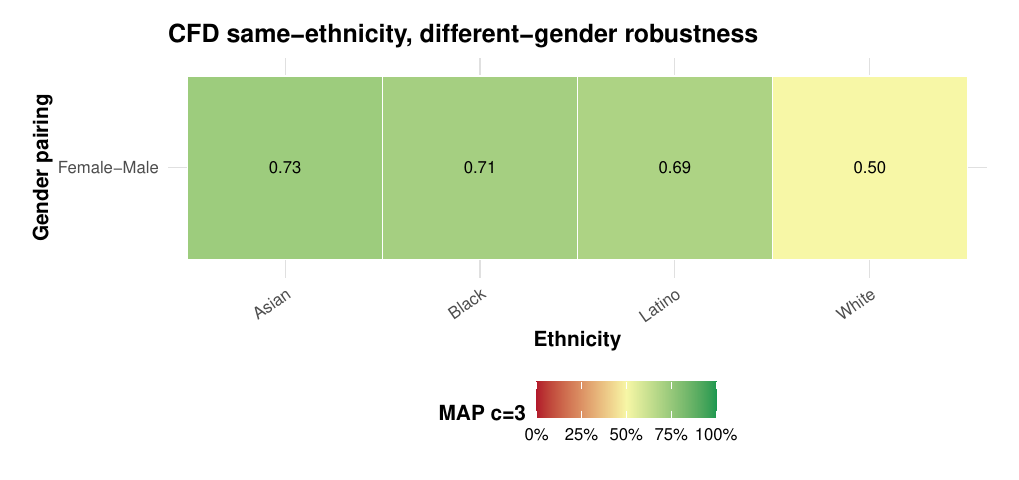}
		\vspace{0.35em}
		{\scriptsize\textbf{(b)} Same-ethnicity different-gender.}
	\end{minipage}

	\vspace{0.65em}
	\includegraphics[width=0.62\linewidth]{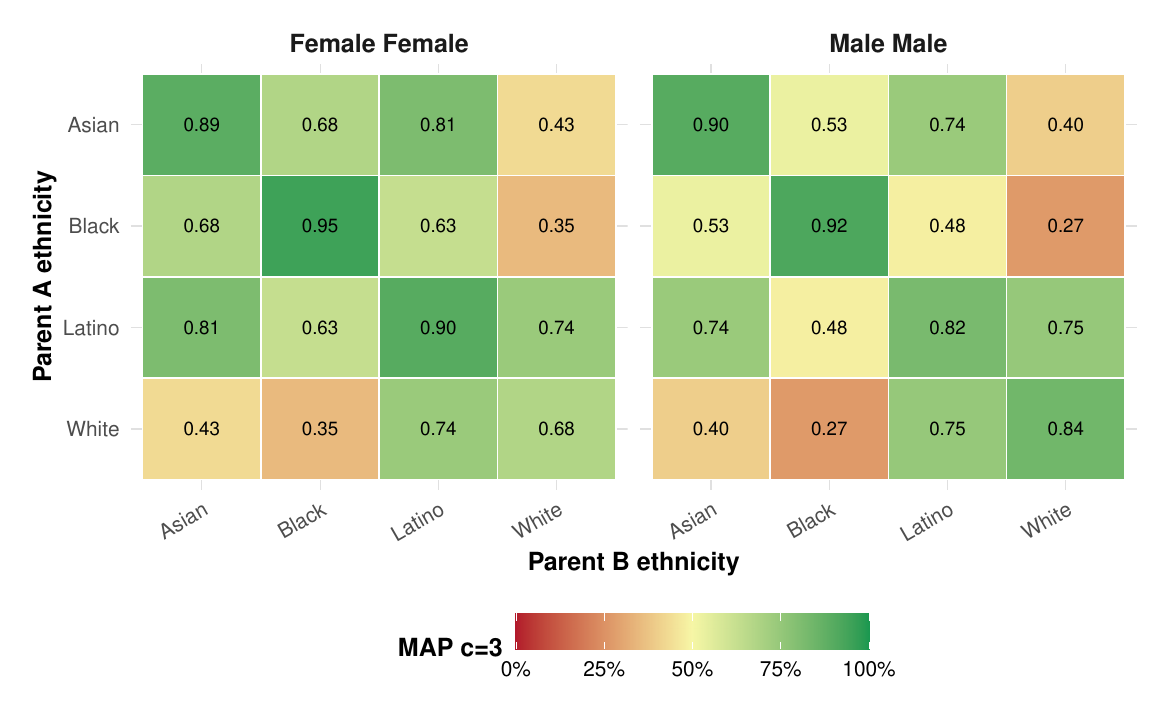}

	\vspace{0.35em}
	{\scriptsize\textbf{(c)} Ethnicity-pair stress categories.}
	\caption{Additional CFD variant robustness grids. The panels show how \map{} changes across same-ethnicity same-gender, same-ethnicity different-gender, and ethnicity-pair CFD stress categories.}
	\label{fig:appendix-cfd-variant-tiles}
\end{figure}

Figure~\ref{fig:appendix-cfd-variant-tiles} shows that CFD robustness follows a structured pattern.
Same-ethnicity and same-gender settings are generally more favourable because the parent images are closer in visual and demographic structure.
Cross-category settings reduce transfer because the model must reconcile larger identity and appearance gaps.
This agrees with the main CFD heatmap and supports the interpretation that unseen-identity robustness depends strongly on parent-pair compatibility.
The three panels also separate two sources of difficulty.
Changing gender while keeping ethnicity fixed primarily tests facial-shape and appearance variation within a broad visual domain.
Changing ethnicity while keeping gender fixed adds a stronger domain shift in texture, facial proportions, and population-level appearance cues.
The fact that performance changes systematically across these panels supports the use of CFD as a stress test of morph-trajectory robustness rather than as another ordinary benchmark split.

\FloatBarrier
\subsection{CFD Qualitative Robustness Examples}
\label{sec:appendix-cfd-qualitative}

\begin{figure}[htbp]
	\centering
	\includegraphics[
		width=\linewidth,
		height=0.43\textheight,
		keepaspectratio
	]{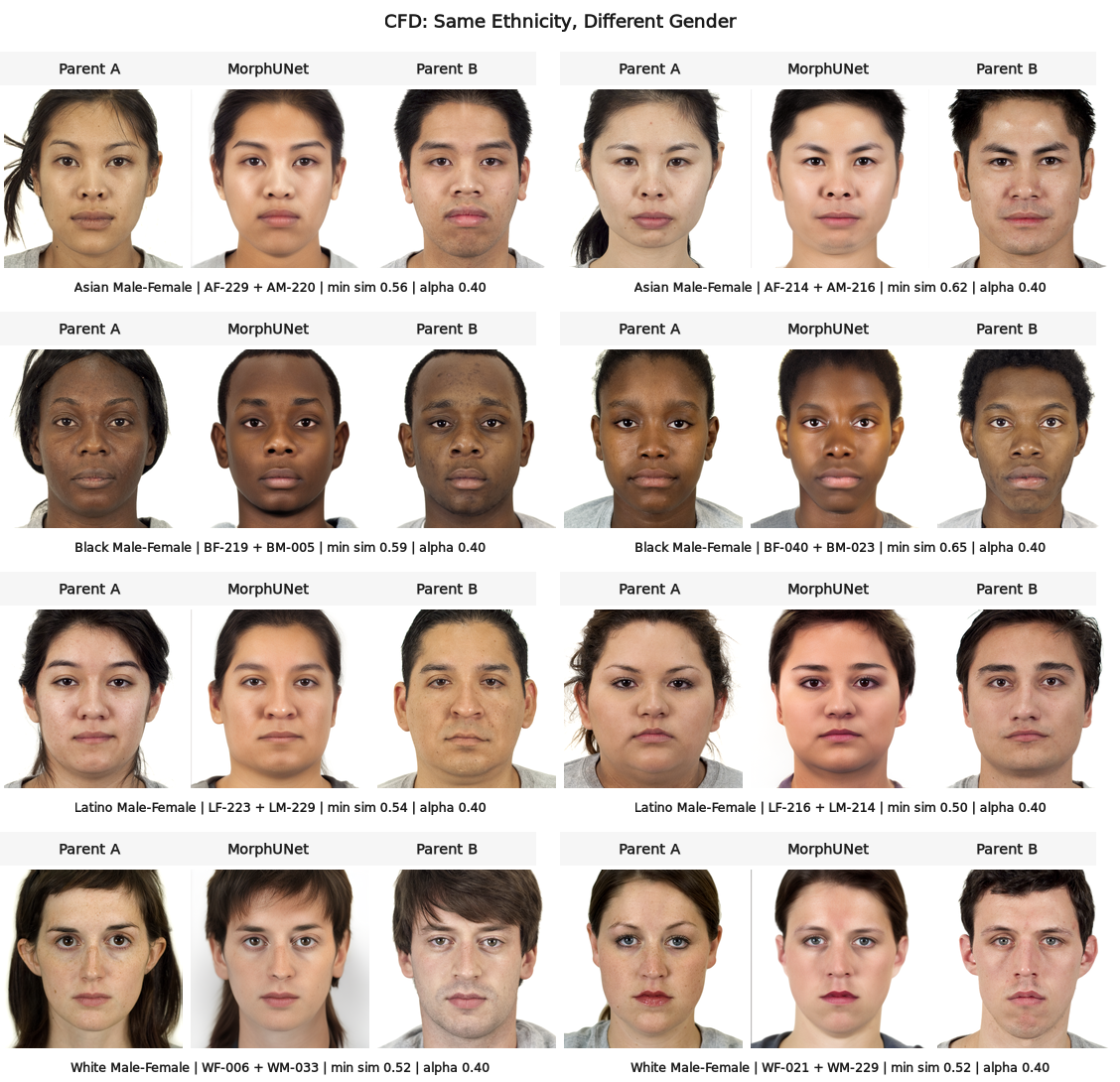}\\[0.4em]

	\includegraphics[
		width=\linewidth,
		height=0.23\textheight,
		keepaspectratio
	]{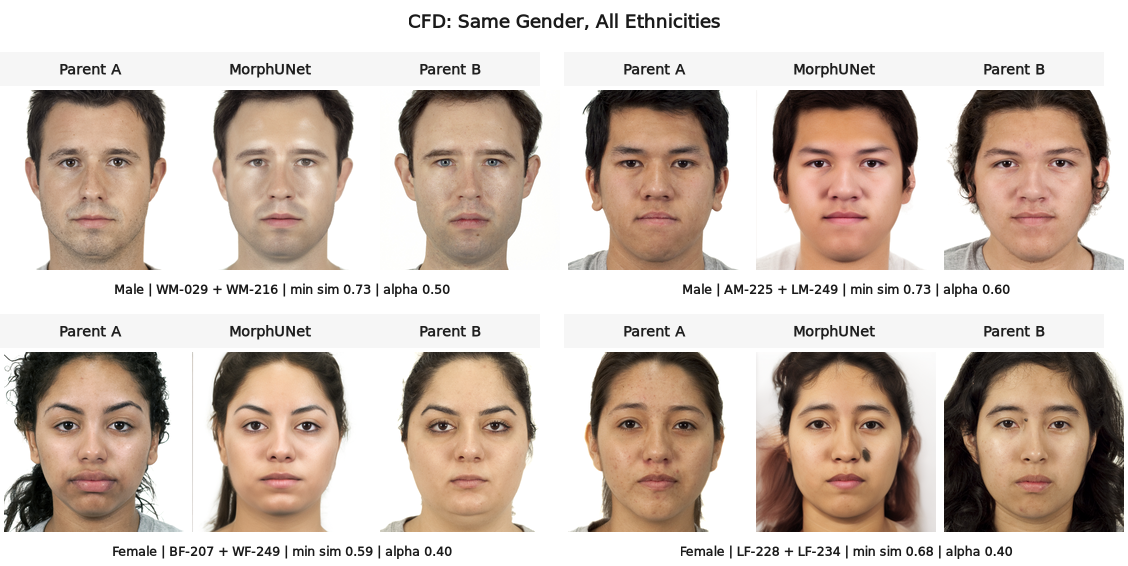}

	\caption{CFD qualitative examples for same-ethnicity and same-gender parent-pair settings. The first grid shows same-ethnicity different-gender pairings, while the second grid shows same-gender examples across ethnicities. These examples test whether \morphunet{} can preserve plausible passport-style appearance under gender and ethnicity variation.}
	\label{fig:appendix-cfd-same-ethnicity}
\end{figure}

\begin{figure}[htbp]
	\centering
	\includegraphics[
		width=\linewidth,
		height=0.66\textheight,
		keepaspectratio
	]{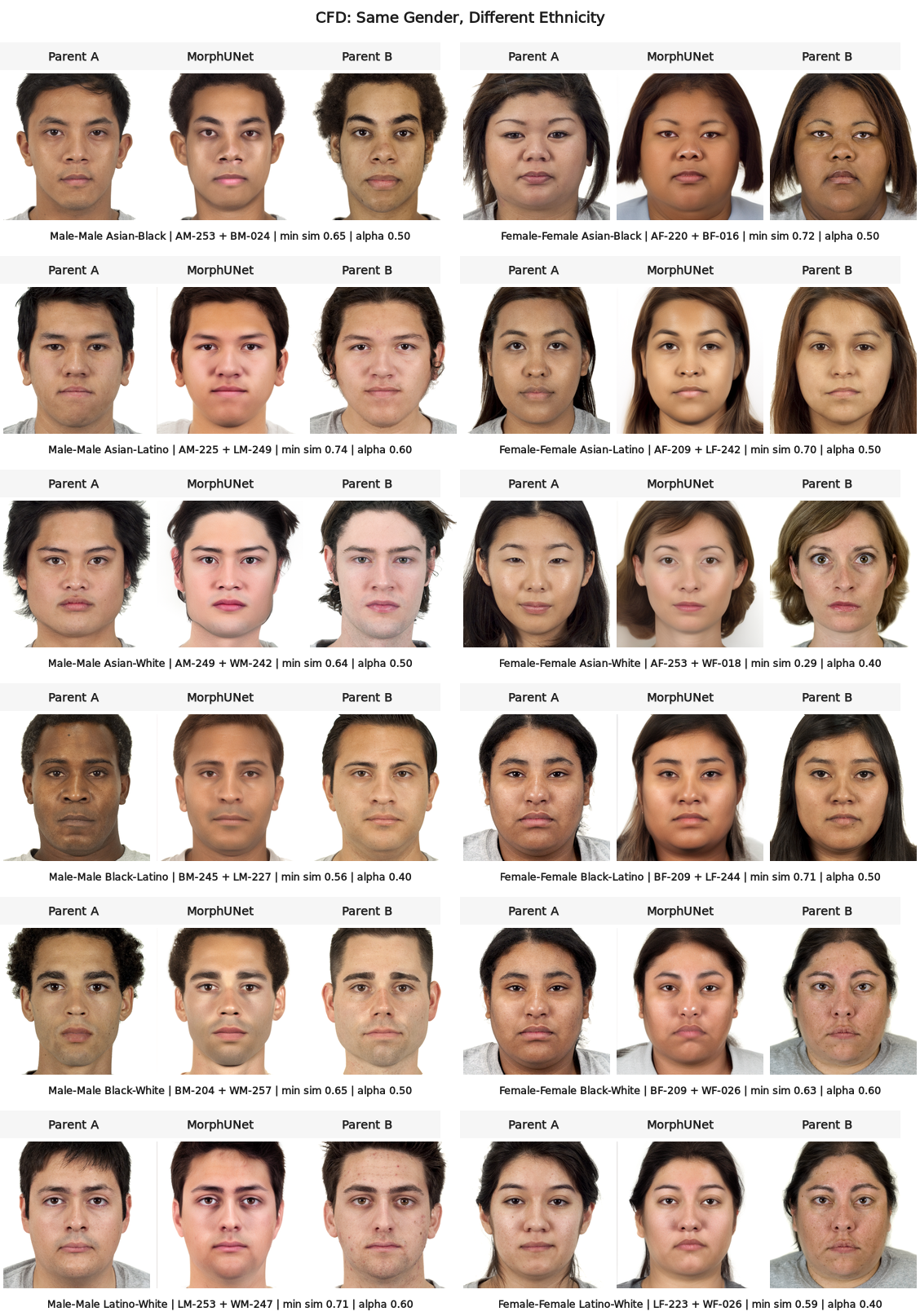}

\caption{CFD qualitative examples for same-gender different-ethnicity parent pairs. These examples show how \morphunet{} behaves when gender is fixed but ethnicity and identity-domain variation change.}
	\label{fig:appendix-cfd-broader-stress}
\end{figure}

The CFD qualitative grids provide visual evidence for the unseen-identity behaviour shown in the quantitative CFD results.
They show how \morphunet{} behaves when parent pairs vary by gender and ethnicity while still requiring a coherent passport-style output. Figures~\ref{fig:appendix-cfd-same-ethnicity} and~\ref{fig:appendix-cfd-broader-stress} show that the generated CFD samples usually remain visually coherent under unseen pairings.
Same-gender and same-ethnicity pairings are generally more stable, while different-ethnicity pairings create a larger visual gap between the parents.
This matches the quantitative CFD results, where cross-category pairs tend to reduce multi-FRS attack transfer.
The examples also show why qualitative inspection remains necessary even when \map{} is high.
Some difficult pairings can produce plausible global faces while still showing local tension in the eye region, face boundary, or mouth shape.
Those local effects are precisely the kinds of artefacts that may affect operational acceptability or detector response, even if the generated image still fools several recognition systems.

Figure~\ref{fig:appendix-cfd-broader-stress} is the most visually demanding CFD setting because the model must bridge identity and ethnicity-domain changes while keeping gender fixed.
The results show that \morphunet{} usually maintains coherent full-face structure, but the morph trajectory can become less balanced when the parents differ strongly in local facial geometry or skin/texture statistics.
This visual pattern is consistent with the category-level \map{} reductions and supports the conclusion that unseen-identity robustness should be evaluated under explicit demographic and similarity stress conditions.

\FloatBarrier
\section{Conditioning Ablation Details}
\label{sec:appendix-conditioning-ablation}

The conditioning ablations analyse which parts of the \morphunet{} identity-conditioning path contribute most to attack strength and image quality.
Each variant simplifies one part of the full design, separating the role of CLIP appearance evidence, ArcFace biometric evidence, parent-specific attention, and alpha-controlled residual transport.

The ablations support the architecture in the architecture section of the main paper.
If the full model improves over CLIP-only or ArcFace-only variants, this indicates that neither appearance nor biometric identity evidence is sufficient on its own.
If the full model improves over raw interpolation or simple self-attention fusion, this supports the claim that the two parents should remain separately accessible before alpha-controlled mixing.
The ablation section is therefore not intended to introduce new competing morphing systems.
Instead, each simplified variant removes or weakens one design decision so that the supplement can test whether the final performance comes from the full two-parent transport path or from a simpler conditioning shortcut.
This is especially important because a diffusion backbone can produce realistic images even when the biometric conditioning mechanism is weak; realism alone would not validate the proposed architecture.

\FloatBarrier
\subsection{Attack-Strength Ablation}
\label{sec:appendix-attack-strength-ablation}

\begin{table}[htbp]
\centering
\scriptsize
\caption{Adapter-ablation MAP summary on the merged benchmark package.}
\label{tab:adapter_ablation_map}
\resizebox{\linewidth}{!}{%
\begin{tabular}{lrrrrrrr}
\toprule
Dataset & Method & \# Morphs & MAP@1 ($\uparrow$) & MAP@3 ($\uparrow$) & MAP@4 ($\uparrow$) & MAP@8 ($\uparrow$) & Mean MAP ($\uparrow$) \\
\midrule
FRLL & Arc Queries + CLIP & 1096.00 & 99.36 & 91.79 & 79.20 & 8.12 & 62.42 \\
FRLL & ArcFace Only & 1096.00 & 99.09 & 91.88 & 79.74 & 7.94 & 62.06 \\
FRLL & CLIP Only & 1096.00 & 99.27 & 89.96 & 76.64 & 6.84 & 60.40 \\
FRLL & MorphUNet & 1096.00 & \textbf{100.00} & \textbf{94.98} & \textbf{82.57} & 8.58 & \textbf{64.02} \\
FRLL & Raw Interp. Injection & 1096.00 & 99.64 & 92.88 & 79.47 & 6.93 & 61.43 \\
FRLL & Self-Attn & 1096.00 & 99.18 & 93.80 & 80.75 & \textbf{8.76} & 62.94 \\
\bottomrule
\end{tabular}%
}
\end{table}

The attack-strength ablation shows how each conditioning design behaves as the required number of fooled recognition systems increases.
In this ablation table, MAP@8 denotes the strictest criterion in the ablation benchmark package, whereas the main paper reports the final six-FRS protocol at $c=1$, $c=3$, and $c=6$.
Simplified variants can remain close under weaker criteria, but the separation becomes clearer once the same morph must fool multiple matchers.
This is where the full transport design is most useful, because it tests whether both identities are preserved in a way that transfers across different recognition models.
In Table~\ref{tab:adapter_ablation_map}, the full \morphunet{} configuration obtains the best MAP@1, MAP@3, MAP@4, and mean MAP.
The self-attention variant is marginally higher only at the strict MAP@8 endpoint, but it does not provide the same overall mean attack strength.
This pattern is important because the best design should not be chosen from one extreme operating point alone; it should maintain strong transfer across the full criterion range.

\begin{figure}[htbp]
	\centering
	\includegraphics[width=0.78\linewidth]{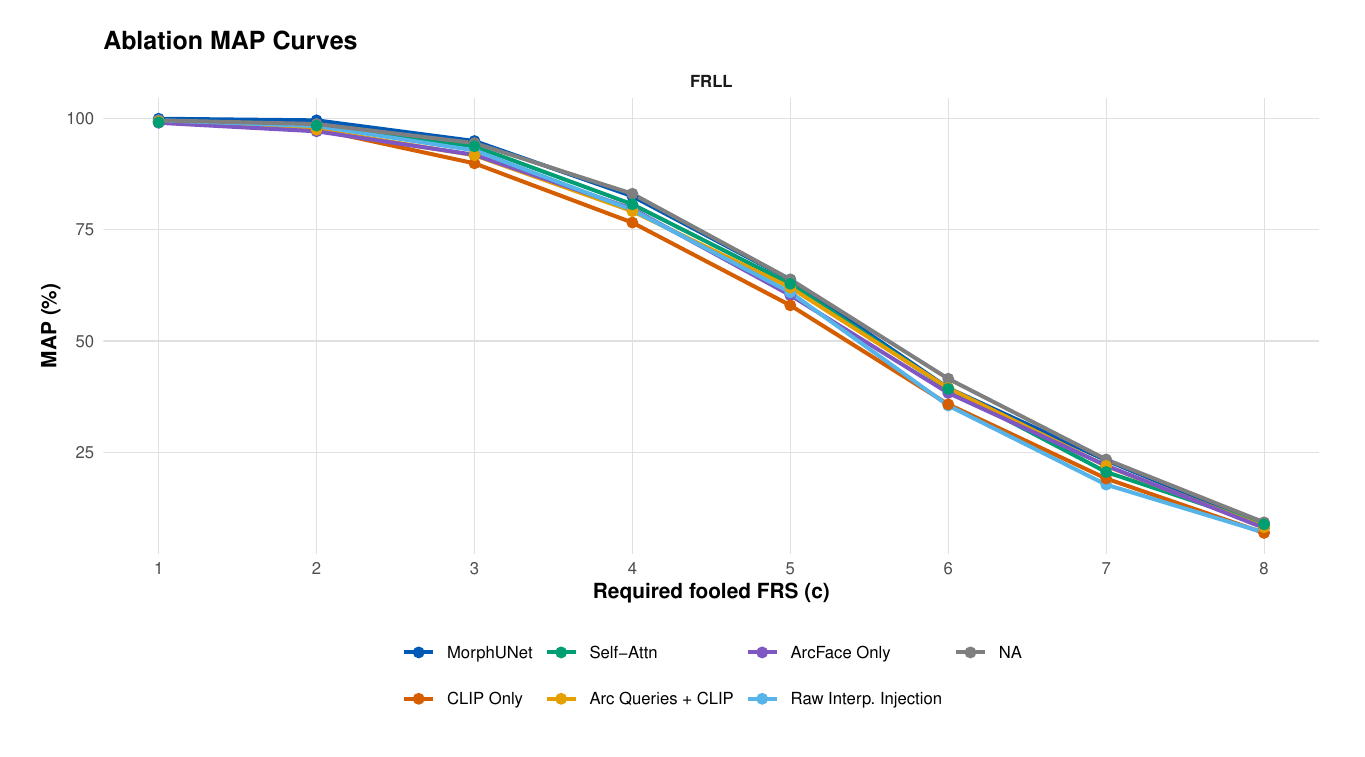}
	\caption{MAP-versus-$c$ curves for the FRLL conditioning ablations. The full \morphunet{} configuration remains among the strongest variants across progressively stricter multi-FRS acceptance criteria.}
	\label{fig:adapter-ablation-map}
\end{figure}

Figure~\ref{fig:adapter-ablation-map} gives a stricter view than the summary table alone.
At low $c$, several conditioning variants appear close because fooling one or two matchers is less demanding.
As $c$ increases, the variants separate because the morph must carry identity evidence that transfers across several recognition systems.
This behaviour supports the use of separate parent attention before alpha-controlled residual mixing.
The curve view also helps distinguish robustness from peak performance.
A variant that performs well only at one criterion may be exploiting a narrow set of matcher sensitivities, whereas a flatter high curve indicates more stable biometric transfer.
\morphunet{} shows the strongest overall curve profile, which supports the claim that parent-specific token banks and delayed residual fusion preserve both identities more consistently than early-fusion alternatives.

\FloatBarrier
\subsection{Quality and Attack Trade-Off}
\label{sec:appendix-quality-attack-ablation}

\begin{table}[htbp]
\centering
\scriptsize
\caption{Adapter-ablation image-quality summary on the merged benchmark package.}
\label{tab:adapter_ablation_quality}
\resizebox{0.55\linewidth}{!}{%
\begin{tabular}{llrr}
\toprule
Dataset & Method & FID ($\downarrow$) & CMMD ($\downarrow$) \\
\midrule
FRLL & Arc Queries + CLIP & 48.54 & 1.59 \\
FRLL & ArcFace Only & 47.23 & 1.58 \\
FRLL & CLIP Only & 48.01 & 1.65 \\
FRLL & MorphUNet & \textbf{44.96} & 1.57 \\
FRLL & Raw Interp. Injection & 46.64 & \textbf{1.47} \\
FRLL & Self-Attn & 46.76 & 1.63 \\
\bottomrule
\end{tabular}
}
\end{table}

The quality ablation shows that visual realism and attack strength do not always move together.
A variant can improve one quality metric without giving the strongest biometric transfer.
The full \morphunet{} configuration gives the strongest overall compromise, combining high \map{} with the best FID among the ablated conditioning choices.
Table~\ref{tab:adapter_ablation_quality} shows the trade-off explicitly.
Raw interpolated injection obtains the best CMMD, but it does not have the strongest mean \map{} in Table~\ref{tab:adapter_ablation_map}.
\morphunet{} obtains the best FID and the best mean \map{}, indicating that the full transport path improves distributional realism while also preserving biometric attack strength.
This supports the main-paper argument that morph quality must be evaluated jointly across recognition transfer and image realism.

\begin{figure}[htbp]
	\centering
	\includegraphics[width=0.78\linewidth]{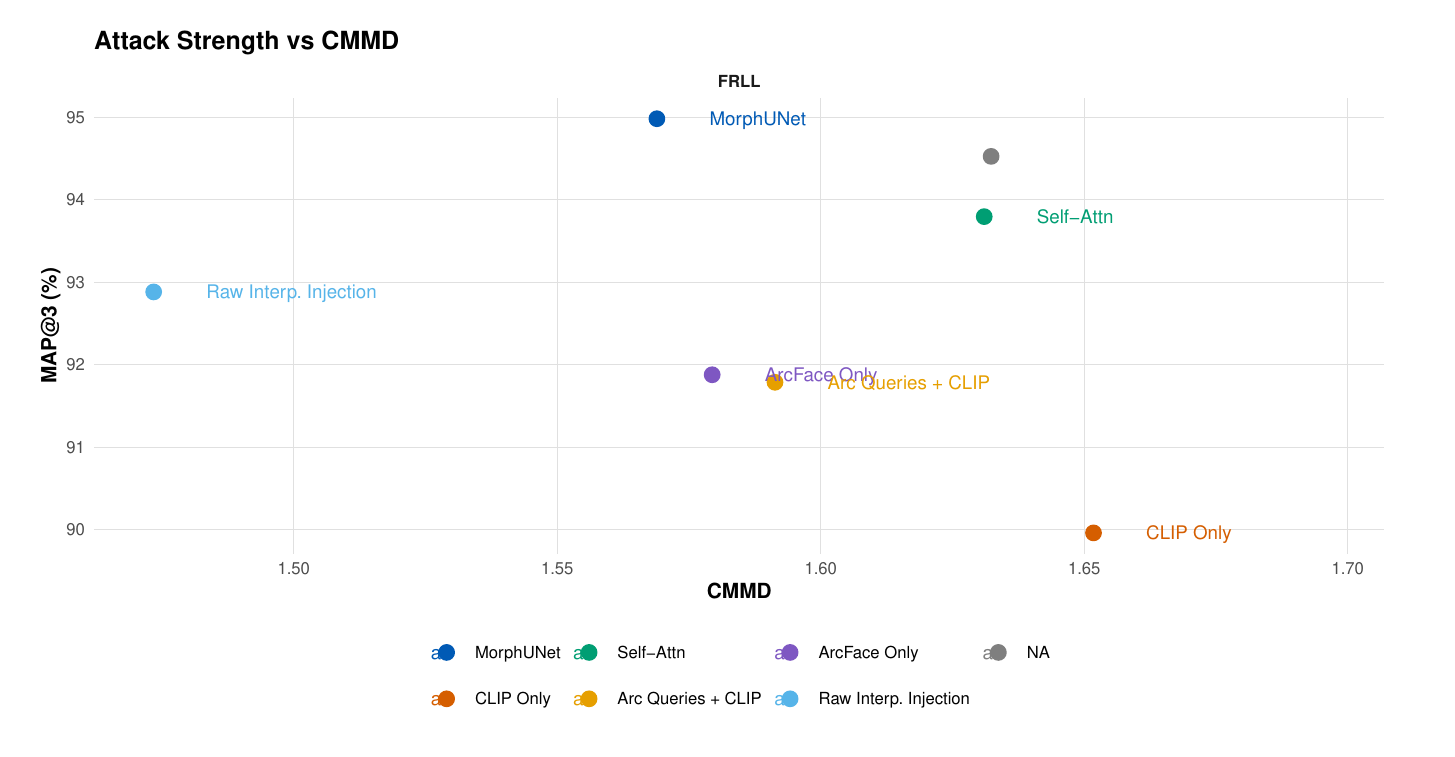}
	\caption{Attack-strength/CMMD trade-off for the FRLL conditioning ablations. The plot shows how simplified conditioning alternatives move along the quality and attack-success axes.}
	\label{fig:adapter-ablation-quality-cmmd}
\end{figure}

\begin{figure}[htbp]
	\centering
	\includegraphics[width=0.78\linewidth]{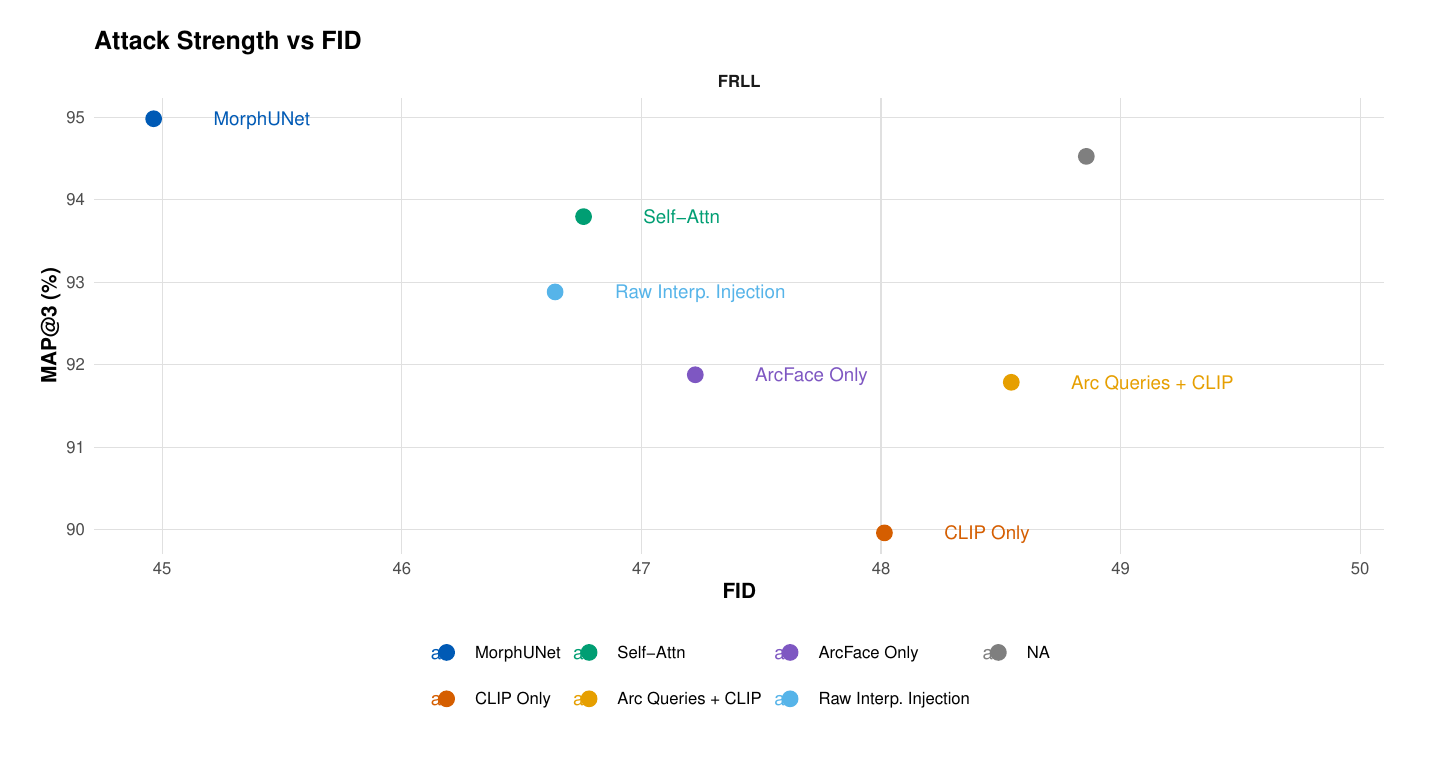}
	\caption{Attack-strength/FID trade-off for the FRLL conditioning ablations. The final \morphunet{} model combines strong attack performance with the best FID among the compared variants.}
	\label{fig:adapter-ablation-quality-fid}
\end{figure}

Figures~\ref{fig:adapter-ablation-quality-cmmd} and~\ref{fig:adapter-ablation-quality-fid} show that the strongest morphing configuration is not simply the one with the best individual quality score.
A useful morphing attack must remain visually plausible while preserving enough biometric evidence from both parents.
The full \morphunet{} variant occupies the strongest compromise region because it combines high \map{} with strong image realism.
The CMMD and FID plots make slightly different points.
CMMD is sensitive to CLIP-distribution proximity, which can favour variants that remain visually close to the reference distribution without necessarily giving the best biometric transfer.
FID gives a complementary distributional view, where \morphunet{} is strongest among the ablated variants.
Reading the two plots together prevents over-interpreting any single quality metric.

\FloatBarrier
\subsection{Alpha-Trajectory Behaviour}
\label{sec:appendix-alpha-trajectory-ablation}
\begin{figure}[htbp]
	\centering
	\includegraphics[width=0.98\textwidth]{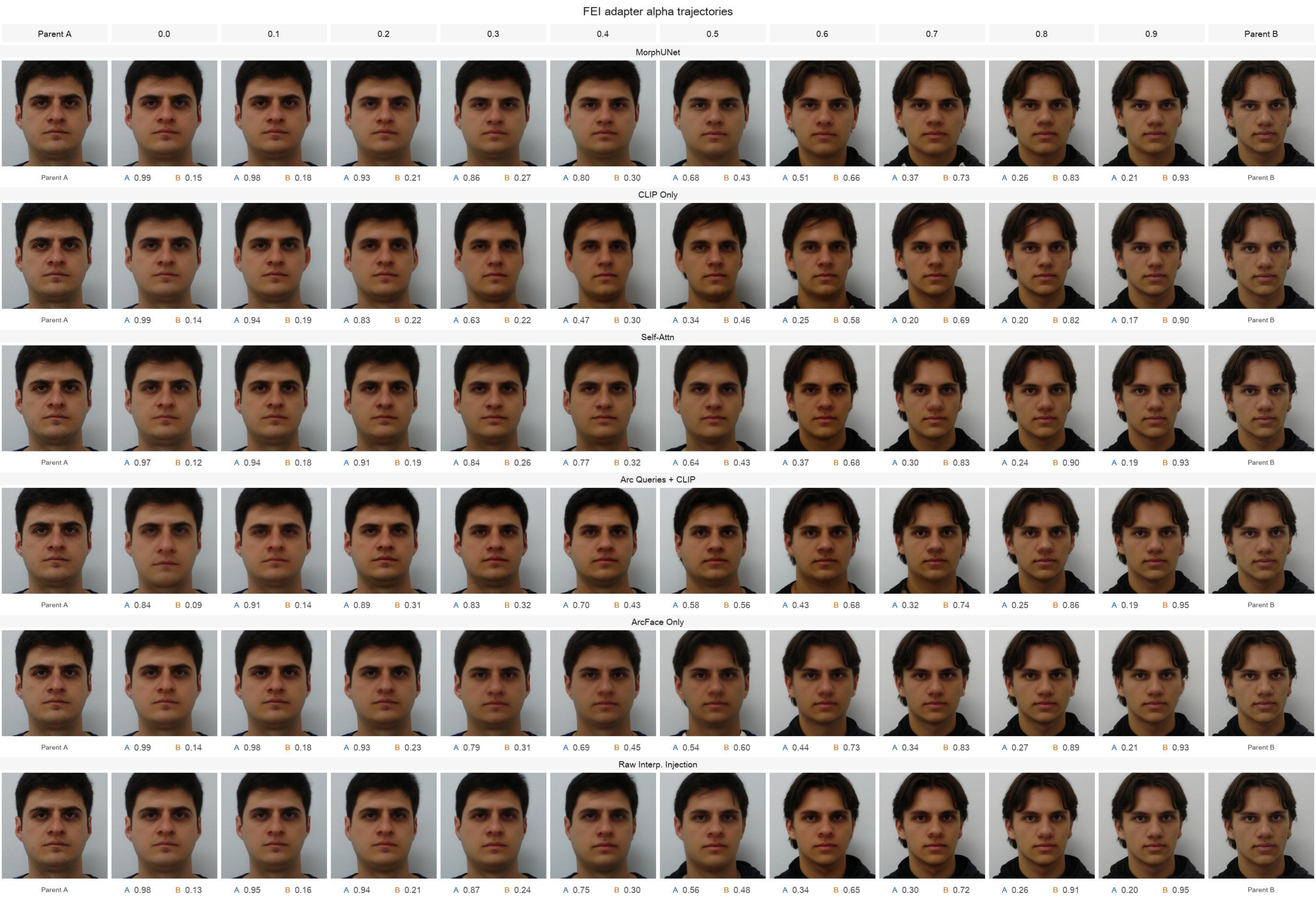}
	\vspace{0.4em}
	\includegraphics[width=0.98\textwidth]{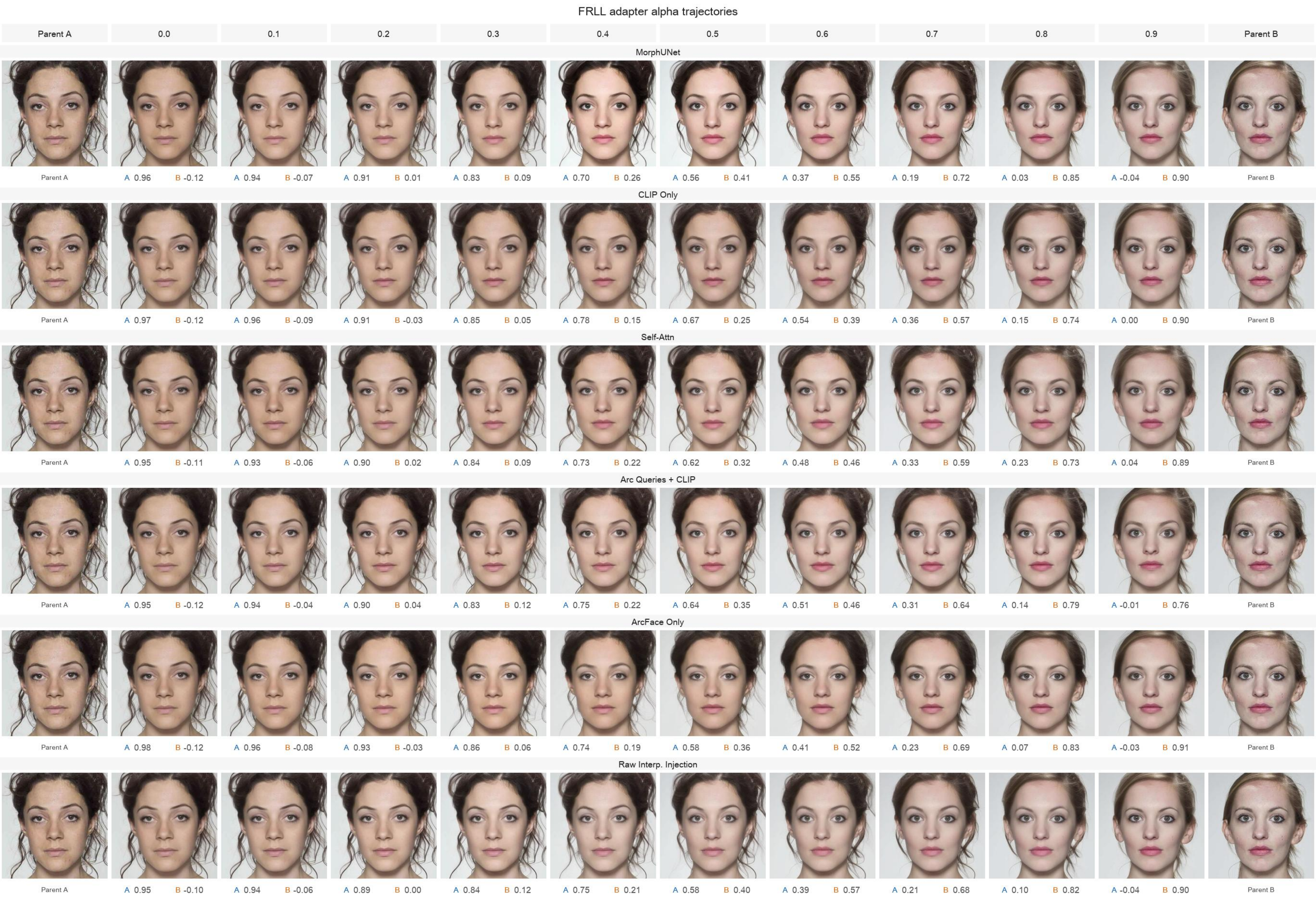}
	\caption{Qualitative alpha-trajectory ablations for FEI and FRLL. Each row shows how the generated morph evolves as the interpolation coefficient increases, while the columns compare the final \morphunet{} configuration against simplified conditioning alternatives.}
	\label{fig:adapter-ablation-qualitative}
\end{figure}

Figure~\ref{fig:adapter-ablation-qualitative} shows how the ablation differences appear visually across the alpha trajectory.
A strong conditioning design should not only produce one successful midpoint morph; it should also change smoothly as $\alpha$ moves from one parent toward the other.
The full \morphunet{} configuration gives a clearer identity transition across the trajectory, while simplified variants are more likely to show identity collapse, weaker-parent loss, or unstable appearance continuity.
This visual trajectory is the qualitative counterpart of alpha-parametrised supervision.
If alpha were only a post-hoc interpolation knob, the trajectory could change appearance without preserving a meaningful biometric path between the parents.
The full model instead maintains a smoother parent-to-parent progression, which supports the claim that the transport layers learn how identity evidence should move during denoising.

Overall, the ablation evidence supports the central architectural claim of the paper.
Strong diffusion morphing requires both appearance-compatible conditioning and biometric identity evidence, but these signals should not be collapsed too early.
The full \morphunet{} transport path gives the most balanced behaviour because each parent can influence the U-Net hidden state separately before alpha-controlled residual mixing.

\end{document}